\PassOptionsToPackage{unicode}{hyperref}
\PassOptionsToPackage{hyphens}{url}
\PassOptionsToPackage{dvipsnames,svgnames,x11names}{xcolor}
\documentclass[
]{article}
\usepackage{xcolor}
\usepackage{amsmath,amssymb}
\usepackage{iftex}
\ifPDFTeX
  \usepackage[T1]{fontenc}
  \usepackage[utf8]{inputenc}
  \usepackage{textcomp} % provide euro and other symbols
\else % if luatex or xetex
  \usepackage{unicode-math} % this also loads fontspec
  \defaultfontfeatures{Scale=MatchLowercase}
  \defaultfontfeatures[\rmfamily]{Ligatures=TeX,Scale=1}
\fi
\usepackage{lmodern}
\ifPDFTeX\else
\fi
\IfFileExists{upquote.sty}{\usepackage{upquote}}{}
\IfFileExists{microtype.sty}{% use microtype if available
  \usepackage[]{microtype}
  \UseMicrotypeSet[protrusion]{basicmath} % disable protrusion for tt fonts
}{}
\makeatletter
\@ifundefined{KOMAClassName}{% if non-KOMA class
  \IfFileExists{parskip.sty}{%
    \usepackage{parskip}
  }{% else
    \setlength{\parindent}{0pt}
    \setlength{\parskip}{6pt plus 2pt minus 1pt}}
}{% if KOMA class
  \KOMAoptions{parskip=half}}
\makeatother
\makeatletter
\ifx\paragraph\undefined\else
  \let\oldparagraph\paragraph
  \renewcommand{\paragraph}{
    \@ifstar
      \xxxParagraphStar
      \xxxParagraphNoStar
  }
  \newcommand{\xxxParagraphStar}[1]{\oldparagraph*{#1}\mbox{}}
  \newcommand{\xxxParagraphNoStar}[1]{\oldparagraph{#1}\mbox{}}
\fi
\ifx\subparagraph\undefined\else
  \let\oldsubparagraph\subparagraph
  \renewcommand{\subparagraph}{
    \@ifstar
      \xxxSubParagraphStar
      \xxxSubParagraphNoStar
  }
  \newcommand{\xxxSubParagraphStar}[1]{\oldsubparagraph*{#1}\mbox{}}
  \newcommand{\xxxSubParagraphNoStar}[1]{\oldsubparagraph{#1}\mbox{}}
\fi
\makeatother

\usepackage{longtable,booktabs,array}
\usepackage{calc} % for calculating minipage widths
\usepackage{etoolbox}
\makeatletter
\patchcmd\longtable{\par}{\if@noskipsec\mbox{}\fi\par}{}{}
\makeatother
\IfFileExists{footnotehyper.sty}{\usepackage{footnotehyper}}{\usepackage{footnote}}
\makesavenoteenv{longtable}
\usepackage{graphicx}
\makeatletter
\newsavebox\pandoc@box
\newcommand*\pandocbounded[1]{% scales image to fit in text height/width
  \sbox\pandoc@box{#1}%
  \Gscale@div\@tempa{\textheight}{\dimexpr\ht\pandoc@box+\dp\pandoc@box\relax}%
  \Gscale@div\@tempb{\linewidth}{\wd\pandoc@box}%
  \ifdim\@tempb\p@<\@tempa\p@\let\@tempa\@tempb\fi% select the smaller of both
  \ifdim\@tempa\p@<\p@\scalebox{\@tempa}{\usebox\pandoc@box}%
  \else\usebox{\pandoc@box}%
  \fi%
}
\def\fps@figure{htbp}
\makeatother

\NewDocumentCommand\citeproctext{}{}
\NewDocumentCommand\citeproc{mm}{%
  \begingroup\def\citeproctext{#2}\cite{#1}\endgroup}
\makeatletter
 \let\@cite@ofmt\@firstofone
 \def\@biblabel#1{}
 \def\@cite#1#2{{#1\if@tempswa , #2\fi}}
\makeatother
\newlength{\cslhangindent}
\newlength{\csllabelwidth}
\newenvironment{CSLReferences}[2] % #1 hanging-indent, #2 entry-spacing
 {\begin{list}{}{%
  \setlength{\itemindent}{0pt}
  \setlength{\leftmargin}{0pt}
  \setlength{\parsep}{0pt}
  \ifodd #1
   \setlength{\leftmargin}{\cslhangindent}
   \setlength{\itemindent}{-1\cslhangindent}
  \fi
  \setlength{\itemsep}{#2\baselineskip}}}
 {\end{list}}
\usepackage{calc}

\providecommand{\tightlist}{%
  \setlength{\itemsep}{0pt}\setlength{\parskip}{0pt}}

\usepackage{placeins}
\usepackage{siunitx}
\usepackage[title]{appendix}
\usepackage{arxiv}
\usepackage{orcidlink}
\usepackage{amsmath}
\usepackage[T1]{fontenc}
\usepackage{placeins}
\usepackage{siunitx}
\makeatletter
\@ifpackageloaded{tcolorbox}{}{\usepackage[skins,breakable]{tcolorbox}}
\@ifpackageloaded{fontawesome5}{}{\usepackage{fontawesome5}}
\definecolor{quarto-callout-color}{HTML}{909090}
\definecolor{quarto-callout-note-color}{HTML}{0758E5}
\definecolor{quarto-callout-important-color}{HTML}{CC1914}
\definecolor{quarto-callout-warning-color}{HTML}{EB9113}
\definecolor{quarto-callout-tip-color}{HTML}{00A047}
\definecolor{quarto-callout-caution-color}{HTML}{FC5300}
\definecolor{quarto-callout-color-frame}{HTML}{acacac}
\definecolor{quarto-callout-note-color-frame}{HTML}{4582ec}
\definecolor{quarto-callout-important-color-frame}{HTML}{d9534f}
\definecolor{quarto-callout-warning-color-frame}{HTML}{f0ad4e}
\definecolor{quarto-callout-tip-color-frame}{HTML}{02b875}
\definecolor{quarto-callout-caution-color-frame}{HTML}{fd7e14}
\makeatother
\makeatletter
\@ifpackageloaded{caption}{}{\usepackage{caption}}
\AtBeginDocument{%
\ifdefined\contentsname
  \renewcommand*\contentsname{Table of contents}
\else
  \newcommand\contentsname{Table of contents}
\fi
\ifdefined\listfigurename
  \renewcommand*\listfigurename{List of Figures}
\else
  \newcommand\listfigurename{List of Figures}
\fi
\ifdefined\listtablename
  \renewcommand*\listtablename{List of Tables}
\else
  \newcommand\listtablename{List of Tables}
\fi
\ifdefined\figurename
  \renewcommand*\figurename{Figure}
\else
  \newcommand\figurename{Figure}
\fi
\ifdefined\tablename
  \renewcommand*\tablename{Table}
\else
  \newcommand\tablename{Table}
\fi
}
\@ifpackageloaded{float}{}{\usepackage{float}}
\floatstyle{ruled}
\@ifundefined{c@chapter}{\newfloat{codelisting}{h}{lop}}{\newfloat{codelisting}{h}{lop}[chapter]}
\floatname{codelisting}{Listing}

\usepackage{amsthm}
\theoremstyle{definition}
\newtheorem{definition}{Definition}[section]
\theoremstyle{definition}
\newtheorem{exercise}{RQ}[section]
\theoremstyle{plain}
\newtheorem{proposition}{Proposition}[section]
\theoremstyle{remark}
\AtBeginDocument{}

\makeatother
\makeatletter
\@ifpackageloaded{caption}{}{\usepackage{caption}}
\@ifpackageloaded{subcaption}{}{\usepackage{subcaption}}
\makeatother
\makeatletter
\@ifpackageloaded{algorithm}{}{\usepackage{algorithm}}
\makeatother
\makeatletter
\@ifpackageloaded{algpseudocode}{}{\usepackage{algpseudocode}}
\makeatother
\makeatletter
\@ifpackageloaded{caption}{}{\usepackage{caption}}
\makeatother
\newcounter{quartocalloutnteno}
\newcommand{\quartocalloutnte}[1]{\refstepcounter{quartocalloutnteno}\label{#1}}
\usepackage{bookmark}
\IfFileExists{xurl.sty}{\usepackage{xurl}}{} % add URL line breaks if available
\hypersetup{
  pdftitle={Counterfactual Training: Teaching Models Plausible and Actionable Explanations},
  pdfauthor={Patrick Altmeyer; Aleksander Buszydlik; Arie van Deursen; Cynthia C. S. Liem},
  pdfkeywords={Counterfactual Training, Counterfactual
Explanations, Algorithmic Recourse, Explainable AI, Representation
Learning},
  colorlinks=true,
  linkcolor={blue},
  filecolor={Maroon},
  citecolor={Blue},
  urlcolor={Blue},
  pdfcreator={LaTeX via pandoc}}

\newcommand{\runninghead}{}
\renewcommand{\runninghead}{Counterfactual Training }
\title{Counterfactual Training: Teaching Models Plausible and Actionable
Explanations\thanks{This work has been accepted for publication at the
IEEE Conference on Secure and Trustworthy Machine Learning (SaTML). The
final version will be available on IEEE Xplore.}}
\def\asep{\\\\\\ } % default: all authors on same column
\def\asep{\And }
\author{\textbf{Patrick
Altmeyer}~\orcidlink{0000-0003-4726-8613}\\\\Delft University of
Technology\\\\\href{mailto:P.Altmeyer@tudelft.nl}{P.Altmeyer@tudelft.nl}\asep\textbf{Aleksander
Buszydlik}~\orcidlink{0009-0004-1219-855X}\\\\Delft University of
Technology\\\\\href{mailto:A.J.Buszydlik@tudelft.nl}{A.J.Buszydlik@tudelft.nl}\asep\textbf{Arie
van Deursen}~\orcidlink{0000-0003-4850-3312}\\\\Delft University of
Technology\\\\\href{mailto:Arie.vanDeursen@tudelft.nl}{Arie.vanDeursen@tudelft.nl}\asep\textbf{Cynthia
C. S. Liem}~\orcidlink{0000-0002-5385-7695}\\\\Delft University of
Technology\\\\\href{mailto:C.C.S.Liem@tudelft.nl}{C.C.S.Liem@tudelft.nl}}
\date{}
\begin{document}
\maketitle
\begin{abstract}
We propose a novel training regime termed counterfactual training that
leverages counterfactual explanations to increase the explanatory
capacity of models. Counterfactual explanations have emerged as a
popular post-hoc explanation method for opaque machine learning models:
they inform how factual inputs would need to change in order for a model
to produce some desired output. To be useful in real-world
decision-making systems, counterfactuals should be plausible with
respect to the underlying data and actionable with respect to the
feature mutability constraints. Much existing research has therefore
focused on developing post-hoc methods to generate counterfactuals that
meet these desiderata. In this work, we instead hold models directly
accountable for the desired end goal: counterfactual training employs
counterfactuals during the training phase to minimize the divergence
between learned representations and plausible, actionable explanations.
We demonstrate empirically and theoretically that our proposed method
facilitates training models that deliver inherently desirable
counterfactual explanations and additionally exhibit improved
adversarial robustness.
\end{abstract}
{\bfseries \emph Keywords}
\def\sep{\textbullet\ }
Counterfactual Training \sep Counterfactual
Explanations \sep Algorithmic Recourse \sep Explainable AI \sep 
Representation Learning

\pagebreak
\floatname{algorithm}{Algorithm}

\section{Introduction}\label{introduction}

Today's prominence of artificial intelligence (AI) has largely been
driven by the success of representation learning with high degrees of
freedom: instead of relying on features and rules hand-crafted by
humans, modern machine learning (ML) models are tasked with learning
highly complex representations directly from the data, guided by narrow
objectives such as predictive accuracy
(\citeproc{ref-goodfellow2016deep}{Goodfellow, Bengio, and Courville
2016}). These models tend to be so complex that humans cannot easily
interpret their decision logic.

\begin{figure}

\centering{

\pandocbounded{\includegraphics[keepaspectratio]{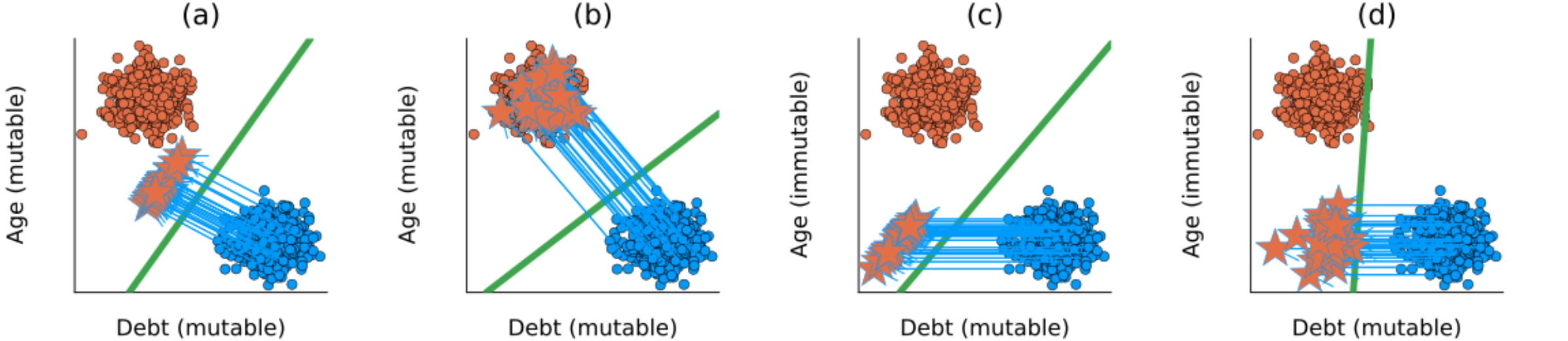}}

}

\caption{\label{fig-poc}Counterfactual explanations (stars) for linear
classifiers trained under different regimes on synthetic data: (a)
conventional training, all mutable; (b) CT, all mutable; (c)
conventional, \emph{age} immutable; (d) CT, \emph{age} immutable. The
linear decision boundary is shown in green along with training data
colored according to ground-truth labels: \(y^-=\text{"loan withheld"}\)
(blue) and \(y^+=\text{"loan provided"}\) (orange). Class and feature
annotations (\emph{debt} and \emph{age}) are for illustrative purposes.}

\end{figure}

Counterfactual explanations (CE) have become a key part of the broader
explainable AI (XAI) toolkit
(\citeproc{ref-molnar2022interpretable}{Molnar 2022}) that can be
applied to make sense of this complexity. They prescribe minimal changes
for factual inputs that, if implemented, would prompt some fitted model
to produce an alternative, more desirable output
(\citeproc{ref-wachter2017counterfactual}{Wachter, Mittelstadt, and
Russell 2017}). This is useful and necessary to not only understand how
opaque models make their predictions, but also to provide algorithmic
recourse to individuals subjected to them: a retail bank, for example,
could use CE to provide meaningful feedback to unsuccessful loan
applicants that were rejected based on an opaque automated
decision-making (ADM) system (Figure~\ref{fig-poc}).

For such feedback to be meaningful, counterfactual explanations need to
fulfill certain desiderata (\citeproc{ref-verma2020counterfactual}{Verma
et al. 2022}; \citeproc{ref-karimi2020survey}{Karimi et al.
2021})---they should be faithful to the model
(\citeproc{ref-altmeyer2024faithful}{Altmeyer et al. 2024}), plausible
(\citeproc{ref-joshi2019realistic}{Joshi et al. 2019}), and actionable
(\citeproc{ref-ustun2019actionable}{Ustun, Spangher, and Liu 2019}).
Plausibility is typically understood as counterfactuals being
\emph{in-domain}: unsuccessful loan applicants that implement the
provided recourse should end up with credit profiles that are genuinely
similar to that of individuals who have successfully repaid their loans
in the past. Actionable explanations further comply with practical
constraints: a young, unsuccessful loan applicant cannot increase their
age in an instant.

Existing state-of-the-art (SOTA) approaches in the field have largely
focused on designing model-agnostic CE methods that identify subsets of
counterfactuals, which comply with specific desiderata. This is
problematic because the narrow focus on any specific desideratum can
adversely affect others: it is possible, for example, to generate
plausible counterfactuals for models that are also highly vulnerable to
implausible, possibly adversarial counterfactuals
(\citeproc{ref-altmeyer2024faithful}{Altmeyer et al. 2024}). Indeed,
existing approaches generally fail to guarantee that the representations
learned by a model are compatible with truly meaningful explanations.

In this work, we propose an approach to bridge this gap, embracing the
paradigm that models---as opposed to explanation methods---should be
held accountable for explanations that are plausible and actionable.
While previous work has shown that at least plausibility can be
indirectly achieved through existing techniques aimed at models'
generative capacity, generalization and robustness
(\citeproc{ref-altmeyer2024faithful}{Altmeyer et al. 2024};
\citeproc{ref-augustin2020adversarial}{Augustin, Meinke, and Hein 2020};
\citeproc{ref-schut2021generating}{Schut et al. 2021}), we directly
incorporate both plausibility and actionability in the training
objective of models to improve their overall explanatory capacity.

Specifically, we introduce \textbf{counterfactual training (CT)}: a
novel training regime that leverages counterfactual explanations
on-the-fly to ensure that differentiable models learn plausible and
actionable explanations for the underlying data, while at the same time
being more robust to adversarial examples (AE). Figure~\ref{fig-poc}
illustrates the outcomes of CT compared to a conventionally trained
model. First, in panel (a), faithful and valid counterfactuals end up
near the decision boundary forming a clearly distinguishable cluster in
the target class (orange). In panel (b), CT is applied to the same
underlying linear classifier architecture resulting in much more
plausible counterfactuals. In panel (c), the classifier is again trained
conventionally and we have introduced a mutability constraint on the
\emph{age} feature at test time---counterfactuals are valid but the
classifier is roughly equally sensitive to both features. By contrast,
the decision boundary in panel (d) has tilted, making the model trained
with CT relatively less sensitive to the immutable \emph{age} feature.
To achieve these outcomes, CT draws inspiration from the literature on
contrastive and robust learning: we contrast faithful CEs with
ground-truth data while protecting immutable features, and capitalize on
methodological links between CE and AE by penalizing the model's
adversarial loss on interim (\emph{nascent}) counterfactuals. To the
best of our knowledge, CT represents the first venture in this direction
with promising empirical and theoretical results.

The remainder of this manuscript is structured as follows.
Section~\ref{sec-lit} presents related work, focusing on the links to
contrastive and robust learning. Then follow our two principal
contributions. In Section~\ref{sec-method}, we introduce our
methodological framework and show theoretically that it can be employed
to respect global actionability constraints. In our experiments
(Section~\ref{sec-experiments}), we find that thanks to counterfactual
training, (1) the implausibility of CEs decreases by up to 90\%; (2) the
cost of reaching valid counterfactuals with protected features decreases
by 19\% on average; and (3) models' adversarial robustness improves
across the board. Finally, we discuss open challenges in
Section~\ref{sec-discussion} and conclude in
Section~\ref{sec-conclusion}.

\section{Related Literature}\label{sec-lit}

To make the desiderata for CT more concrete, we follow previous work,
tying the explanatory capacity of models to the quality of CEs that can
be generated for them (\citeproc{ref-altmeyer2024faithful}{Altmeyer et
al. 2024}; \citeproc{ref-augustin2020adversarial}{Augustin, Meinke, and
Hein 2020}).

\subsection{Explanatory Capacity and Contrastive
Learning}\label{sec-explainability-and-contrastive-learning}

A closely related work shows that model averaging and, in particular,
contrastive model objectives can produce models that have a higher
explanatory capacity, and hence ones that are more trustworthy
(\citeproc{ref-altmeyer2024faithful}{Altmeyer et al. 2024}). The authors
propose a way to generate counterfactuals that are maximally faithful in
that they are consistent with what models have learned about the
underlying data. Formally, they rely on tools from energy-based
modelling (\citeproc{ref-teh2003energy}{Teh et al. 2003}) to minimize
the contrastive divergence between the distribution of counterfactuals
and the conditional posterior over inputs learned by a model. Their
algorithm, \emph{ECCCo}, yields plausible counterfactual explanations if
and only if the underlying model has learned representations that align
with them. The authors find that both deep ensembles
(\citeproc{ref-lakshminarayanan2016simple}{Lakshminarayanan, Pritzel,
and Blundell 2017}) and joint energy-based models (JEMs)
(\citeproc{ref-grathwohl2020your}{Grathwohl et al. 2020}), a form of
contrastive learning, do well in this regard.

It helps to look at these findings through the lens of representation
learning with high degrees of freedom. Deep ensembles are approximate
Bayesian model averages, which are particularly effective when models
are underspecified by the available data
(\citeproc{ref-wilson2020case}{Wilson 2020}). Averaging across solutions
mitigates the risk of overrelying on a single locally optimal
representation that corresponds to semantically meaningless
explanations. Likewise, it has been shown that generating plausible
(``interpretable'') CEs is almost trivial for deep ensembles that have
undergone adversarial training (\citeproc{ref-schut2021generating}{Schut
et al. 2021}). The case for JEMs is even clearer: they optimize a hybrid
objective that induces both high predictive performance and strong
generative capacity (\citeproc{ref-grathwohl2020your}{Grathwohl et al.
2020}), resembling the idea of aligning models with plausible
explanations. This was an inspiration for CT.

\subsection{Explanatory Capacity and Robust
Learning}\label{sec-explainability-and-robust-learning}

Prior work has shown that counterfactual explanations tend to be more
meaningful (``explainable'') if the underlying model is more robust to
adversarial examples (\citeproc{ref-augustin2020adversarial}{Augustin,
Meinke, and Hein 2020}). Once again, we can make intuitive sense of this
finding if we look at adversarial training (AT) through the lens of
representation learning with high degrees of freedom: highly complex and
flexible models may learn representations that make them sensitive to
implausible or even adversarial examples
(\citeproc{ref-szegedy2013intriguing}{Szegedy et al. 2014}). Thus, by
inducing models to ``unlearn'' susceptibility to such examples,
adversarial training can effectively remove implausible explanations
from the solution space.

This interpretation of the link between explanatory capacity through
counterfactuals on the one side, and robustness to adversarial examples
on the other is backed by empirical evidence. Firstly, prior work has
shown that using counterfactual images during classifier training
improves model robustness (\citeproc{ref-sauer2021counterfactual}{Sauer
and Geiger 2021}). Similarly, related work has shown that
counterfactuals represent potentially useful training data in machine
learning tasks, especially in supervised settings where inputs may be
reasonably mapped to multiple outputs
(\citeproc{ref-abbasnejad2020counterfactual}{Abbasnejad et al. 2020}).
The authors show that augmenting the training data of (image)
classifiers can improve generalization performance. Finally, another
related work has demonstrated that counterfactual pairs tend to exist in
training data (\citeproc{ref-teney2020learning}{Teney, Abbasnedjad, and
Hengel 2020}). Hence, the proposed approach aims to identify similar
inputs with different annotations and ensure that the gradient of the
classifier aligns with the vector between such pairs of inputs using a
cosine distance loss function.

CEs have also been used to improve models in the natural language
processing domain. A well-known paper in this domain has proposed
\emph{Polyjuice} (\citeproc{ref-wu2021polyjuice}{Wu et al. 2021}), a
general-purpose CE generator for language models. The authors
demonstrate that the augmentation of training data with \emph{Polyjuice}
improves robustness in a number of tasks. Related work has introduced
the \emph{Counterfactual Adversarial Training} (CAT) framework
(\citeproc{ref-luu2023counterfactual}{Luu and Inoue 2023}), which aims
to improve generalization and robustness of language models by
generating counterfactuals for training samples that are subject to high
predictive uncertainty.

There have also been several attempts at formalizing the relationship
between counterfactual explanations and adversarial examples. Pointing
to clear similarities in how CEs and AEs are generated, prior work makes
the case for jointly studying the opaqueness and robustness problems in
representation learning
(\citeproc{ref-freiesleben2022intriguing}{Freiesleben 2022}). Formally,
the authors show that AEs can be seen as the subset of CEs for which
misclassification is achieved
(\citeproc{ref-freiesleben2022intriguing}{Freiesleben 2022}). Similarly,
others have shown that CEs and AEs are equivalent under certain
conditions (\citeproc{ref-pawelczyk2022exploring}{Pawelczyk et al.
2022}).

Two other works are closely related to ours in that they use
counterfactuals during training with the explicit goal of affecting
certain properties of the post-hoc counterfactual explanations. The
first closely related work has proposed a way to train models that
guarantee recourse to a positive target class with high probability
(\citeproc{ref-ross2021learning}{Ross, Lakkaraju, and Bastani 2024}).
The approach builds on adversarial training by explicitly inducing
susceptibility to targeted AEs for the positive class. Additionally, the
method allows for imposing a set of actionability constraints ex-ante.
For example, users can specify that certain features are immutable. A
second closely related work has introduced the first end-to-end training
pipeline that includes CEs as part of the training procedure
(\citeproc{ref-guo2023counternet}{Guo, Nguyen, and Yadav 2023}); the
\emph{CounterNet} network architecture includes a predictor and a CE
generator, where the parameters of the CE generator are learnable.
Counterfactuals are generated during each training iteration and fed
back to the predictor. In contrast, we impose no restrictions on the
artificial neural network architecture at all.

\section{Counterfactual Training}\label{sec-method}

This section introduces the counterfactual training framework, applying
ideas from contrastive and robust learning to counterfactual
explanations. CT produces models whose learned representations align
with plausible explanations that comply with user-defined actionability
constraints.

Counterfactual explanations are typically generated by solving
variations of the following optimization problem,

\begin{equation}\phantomsection\label{eq-general}{
\begin{aligned}
\min_{\mathbf{x}^\prime \in \mathcal{X}^D} \left\{  {\text{yloss}(\mathbf{M}_\theta(\mathbf{x}^{\prime}),\mathbf{y}^+)}+ \lambda {\text{reg}(\mathbf{x}^{\prime}) }  \right\} 
\end{aligned}
}\end{equation}

where \(\mathbf{M}_\theta: \mathcal{X} \mapsto \mathcal{Y}\) denotes a
classifier, \(\mathbf{x}^{\prime}\) denotes the counterfactual with
\(D\) features and \(\mathbf{y}^+\in\mathcal{Y}\) denotes some target
class. The \(\text{yloss}(\cdot)\) function quantifies the discrepancy
between current model predictions for \(\mathbf{x}^{\prime}\) and the
target class (a conventional choice is cross-entropy). Finally, we use
\(\text{reg}(\cdot)\) to denote any form of regularization used to
induce certain properties on the counterfactual. The seminal CE paper,
(\citeproc{ref-wachter2017counterfactual}{Wachter, Mittelstadt, and
Russell 2017}), proposes regularizing the distance between
counterfactuals and their original factual values to ensure that
individuals seeking recourse through CE face minimal costs in terms of
feature changes. Different variations of Equation~\ref{eq-general} have
been proposed in the literature to address many desiderata including the
ones discussed above (faithfulness, plausibility and actionability).
Much like in the seminal work
(\citeproc{ref-wachter2017counterfactual}{Wachter, Mittelstadt, and
Russell 2017}), most of these approaches rely on gradient descent to
optimize Equation~\ref{eq-general}, and this holds true for all
approaches tested in this work. We introduce them briefly in
Section~\ref{sec-experimental-setup}, but refer the reader to the
supplementary appendix for details. In the following, we describe how
counterfactuals are generated and used in CT.

\subsection{Proposed Training
Objective}\label{proposed-training-objective}

The goal of CT is to improve the explanatory capacity of models by
aligning the learned representations with faithful explanations that are
plausible and actionable. For simplicity, we refer to models with high
explanatory capacity as \textbf{explainable} in this manuscript. We
define explainability as follows:

\begin{definition}[Model
Explainability]\protect\hypertarget{def-explainability}{}\label{def-explainability}

Let \(\mathbf{M}_\theta: \mathcal{X} \mapsto \mathcal{Y}\) denote a
supervised classification model that maps from the \(D\)-dimensional
input space \(\mathcal{X}\) to representations
\(\phi(\mathbf{x};\theta)\) and finally to the \(K\)-dimensional output
space \(\mathcal{Y}\). Let \(\mathbf{x}_0^\prime\) denote a factual
input and assume that for any given input-output pair
\(\{\mathbf{x}_0^\prime,\mathbf{y}\}_i\) there exists a counterfactual
\(\mathbf{x}^{\prime} = \mathbf{x}_0^\prime + \Delta: \mathbf{M}_\theta(\mathbf{x}^{\prime}) = \mathbf{y}^{+} \neq \mathbf{y} = \mathbf{M}_\theta(\mathbf{x})\),
where \(\arg\max_y{\mathbf{y}^{+}}=y^+\) is the index of the target
class.

We say that \(\mathbf{M}_\theta\) has an \textbf{explanatory capacity}
to the extent that faithfully generated, valid counterfactuals are also
plausible and actionable. We define these properties as:

\begin{itemize}
\item
  (Faithfulness)
  \(P(\mathbf{x}^\prime \in \mathcal{X}_\theta|\mathbf{y}^+) = 1 - \delta\),
  where \(\delta\) is some small value, and
  \(\mathcal{X}_\theta|\mathbf{y}^+\) is the conditional posterior
  distribution over inputs (adapted from
  (\citeproc{ref-altmeyer2024faithful}{Altmeyer et al. 2024}), Def.
  4.1).
\item
  (Plausibility)
  \(P(\mathbf{x}^\prime \in \mathcal{X}|\mathbf{y}^+) = 1 - \delta\),
  where \(\delta\) is some small value, and \(\mathcal{X}|\mathbf{y}^+\)
  is the conditional distribution of inputs in the target class (adapted
  from (\citeproc{ref-altmeyer2024faithful}{Altmeyer et al. 2024}), Def.
  2.1).
\item
  (Actionability) Perturbations \(\Delta\) may be subject to some
  actionability constraints.
\end{itemize}

\end{definition}

Intuitively, plausible counterfactuals are consistent with the data, and
faithful counterfactuals are consistent with what the model has learned
about the input data. Actionability constraints in
Definition~\ref{def-explainability} depend on the context in which
\(\mathbf{M}_\theta\) is deployed (e.g., specified by end-users or model
owners). We consider two types of actionability constraints: on the
domain of features and on their mutability. The former naturally arise
in automated decision-making systems whenever a feature can only take a
specific range of values. For example, \emph{age} is lower bounded by
zero and upper bounded by the maximum human lifespan. Specifying such
domain constraints can also help address training instabilities commonly
associated with energy-based modelling
(\citeproc{ref-grathwohl2020your}{Grathwohl et al. 2020}). The latter
arise when a feature cannot be freely modified. Continuing the example,
\emph{age} of a person can only increase, but it may even be considered
as an immutable feature: waiting many years for an improved outcome is
hardly feasible for individuals affected by algorithmic decisions. We
choose to only consider domain and mutability constraints for individual
features \(x_d\) for \(d=1,...,D\). Of course, this is a simplification
since feature values may correlate, e.g., higher \emph{age} may be
associated with higher \emph{level of completed education}. We address
this challenge in Section~\ref{sec-discussion}, where we also explain
why we restrict this work to classification settings.

Let \(\mathbf{x}_t^\prime\) for \(t=0,...,T\) denote a counterfactual
generated through gradient descent over \(T\) iterations as originally
proposed (\citeproc{ref-wachter2017counterfactual}{Wachter, Mittelstadt,
and Russell 2017}). CT adopts gradient-based CE search in training to
generate on-the-fly model explanations \(\mathbf{x}^\prime\) for the
training samples. We use the term \emph{nascent} to denote interim
counterfactuals \(\mathbf{x}_{\text{AE}}^\prime\) that have not yet
converged. As we explain below, these nascent counterfactuals can be
stored and repurposed as adversarial examples. Conversely, we consider
counterfactuals \(\mathbf{x}_{\text{CE}}^\prime\) as \emph{mature}
explanations if they have converged within the \(T\) iterations by
reaching a pre-specified threshold, \(\tau\), for the predicted
probability of the target class:
\(\mathcal{S}(\mathbf{M}_\theta(\mathbf{x}^\prime))[y^+] \geq \tau\),
where \(\mathcal{S}\) is the softmax function.

Formally, we propose the following counterfactual training objective to
train explainable (as in Definition~\ref{def-explainability}) models,

\begin{equation}\phantomsection\label{eq-obj}{
\begin{aligned}
&\min_\theta \text{yloss}(\mathbf{M}_\theta(\mathbf{x}),\mathbf{y}) + \lambda_{\text{div}} \text{div}(\mathbf{x}^+,\mathbf{x}_{\text{CE}}^\prime,y^+;\theta) \\+ &\lambda_{\text{adv}} \text{advloss}(\mathbf{M}_\theta(\mathbf{x}_{\text{AE}}^\prime),\mathbf{y}_{\text{AE}}) + \lambda_{\text{reg}}\text{ridge}(\mathbf{x}^+,\mathbf{x}_{\text{CE}}^\prime,y;\theta)
\end{aligned}
}\end{equation}

where \(\text{yloss}(\cdot)\) is any classification loss that induces
discriminative performance (e.g., cross-entropy). The second and third
terms are explained in detail in the following subsections. For now,
they can be summarized as inducing explainability directly and
indirectly by penalizing (1) the contrastive divergence,
\(\text{div}(\cdot)\), between mature counterfactuals
\(\mathbf{x}_{\text{CE}}^\prime\) and observed samples
\(\mathbf{x}^+\in\mathcal{X}^+=\{\mathbf{x}:y=y^+\}\) in the target
class \(y^+\), and (2) the adversarial loss, \(\text{advloss}(.)\), wrt.
nascent counterfactuals \(\mathbf{x}_{\text{AE}}^\prime\) and their
corresponding labels \(\mathbf{y}_{\text{AE}}\). Finally,
\(\text{ridge}(\cdot)\) denotes a Ridge penalty (squared
\(\ell_2\)-norm) that regularizes the magnitude of the energy terms
involved in the contrastive divergence, \(\text{div}(\cdot)\), term
(\citeproc{ref-du2019implicit}{Du and Mordatch 2020}):

\begin{equation}\phantomsection\label{eq-ridge}{
\frac{1}{n_{\text{CE}}} \sum_{i=1}^{n_{\text{CE}}} \left( \mathcal{E}_{\theta}(\mathbf{x^+},y^+)^2 + \mathcal{E}_{\theta}(\mathbf{x}_{\text{CE}}^\prime,y^+)^2 \right)
}\end{equation}

The trade-offs between these components are adjusted through penalties
\(\lambda_{\text{div}}\), \(\lambda_{\text{adv}}\), and
\(\lambda_{\text{reg}}\).

The full counterfactual training regime is sketched out in
Algorithm~\ref{algo-experiment}. During each iteration, we do the
following steps. Firstly, we randomly draw a subset of
\(n_{\text{CE}} \leq n\) factuals \(\mathbf{x}_0^\prime\) from
\(\mathbf{X}\) of size \(n\), for which we uniformly draw a target class
\(y^+\) (ensuring that it does not coincide with the class currently
predicted for \(\mathbf{x}_0^\prime\)) and a corresponding training
sample from the target class,
\(\mathbf{x}^+\sim \mathbf{X}^+=\{\mathbf{x} \in \mathbf{X}:y=y^+\}\).
Secondly, we conduct the counterfactual search by solving
(Equation~\ref{eq-general}) through gradient descent. Thirdly, we sample
mini-batches \({(\mathbf{x}_i, \mathbf{y}_i)}_{i=1}^{n_b}\) from the
training dataset \(\mathcal{D}=(\mathbf{X},\mathbf{Y})\) for
conventional training and distribute the tuples composed of
counterfactuals, their target labels and corresponding training samples,
as well as adversarial examples and corresponding labels,
\(({\mathbf{x}_{\text{CE}}^\prime}_i,{y^+}_i,{\mathbf{x}_{\text{AE}}^\prime}_i,{\mathbf{y}_{\text{AE}}}_i,{\mathbf{x}^+}_i)_{i=1}^{n_{\text{CE}}}\),
across the mini-batches. Finally, we backpropagate through
(Equation~\ref{eq-obj}).

\begin{algorithm}[H]
\caption{Pseudo-Code for Counterfactual Training}
\label{algo-experiment}
\begin{algorithmic}[1]
\Require Training dataset $\mathcal{D}$, initialize model $\mathbf{M}_{\theta}$
\While{not converged}
  \State Sample $\mathbf{x}^{\prime}_0\sim \mathbf{X}$, $y^+\sim\mathcal{U}(\mathcal{Y})$ and $\mathbf{x}^+ \sim \mathbf{X}^+$
  \For{$t = 1$ to $T$}
    \State Backpropagate $\nabla_{\mathbf{x}^\prime}$ through equation (\ref{eq-general})
    \State Store $\mathbf{x}_{\text{CE}}^\prime,\mathbf{x}_{\text{AE}}^\prime,\mathbf{y}_{\text{AE}}$
  \EndFor
  \State Sample mini-batches ${(\mathbf{x}_i, \mathbf{y}_i)}_{i=1}^{n_b}$ from dataset $\mathcal{D}$
  \State Distribute $({\mathbf{x}_{\text{CE}}^\prime}_i,{y^+}_i,{\mathbf{x}_{\text{AE}}^\prime}_i,{\mathbf{y}_{\text{AE}}}_i,{\mathbf{x}^+}_i)_{i=1}^{n_{\text{CE}}}$
  \For{each batch}
    \State Backpropagate $\nabla_{\theta}$ through equation (\ref{eq-obj})
  \EndFor
\EndWhile
\State \Return $\mathbf{M}_\theta$
\end{algorithmic}
\end{algorithm}

By limiting ourselves to a subset of \(n_{\text{CE}}\) counterfactuals,
we reduce runtimes; this approach has previously been shown to improve
efficiency in the context of adversarial training
(\citeproc{ref-kurakin2017adversarialmachinelearningscale}{Kurakin,
Goodfellow, and Bengio 2017};
\citeproc{ref-kaufmann2022efficient}{Kaufmann et al. 2022}). To improve
runtimes even more, we choose to first generate counterfactuals and then
distribute them across mini-batches to benefit from greater degrees of
parallelization during the counterfactual search. Alternatively, it is
possible to generate counterfactuals separately for each
mini-batch.\footnote{During initial prototyping of CT we also tested an
  implementation that relies on generating counterfactuals and
  adversarial examples at the batch level with no discernible difference
  in outcomes, but increased training times.}

\subsection{Directly Inducing Explainability: Contrastive
Divergence}\label{sec-directly-inducing-explainability-with-contrastive-divergence}

As observed in prior related work
(\citeproc{ref-grathwohl2020your}{Grathwohl et al. 2020}), any
classifier can be re-interpreted as a joint energy-based model that
learns to discriminate output classes conditional on the observed
(training) samples from \(p(\mathbf{x})\) and the generated samples from
\(p_\theta(\mathbf{x})\). The authors show that JEMs can be trained to
perform well at both tasks by directly maximizing the joint
log-likelihood:
\(\log p_\theta(\mathbf{x},\mathbf{y})=\log p_\theta(\mathbf{y}|\mathbf{x}) + \log p_\theta(\mathbf{x})\),
where the first term can be optimized using cross-entropy as in
Equation~\ref{eq-obj}. To optimize \(\log p_\theta(\mathbf{x})\), they
minimize the contrastive divergence between the observed samples from
\(p(\mathbf{x})\) and samples generated from \(p_\theta(\mathbf{x})\).

To generate samples, the paper introducing JEMs
(\citeproc{ref-grathwohl2020your}{Grathwohl et al. 2020}) suggests
relying on Stochastic Gradient Langevin Dynamics (SGLD) with an
uninformative prior for initialization but we depart from this
methodology: we propose to leverage counterfactual explainers to
generate counterfactuals of observed training samples. Specifically, we
have:

\begin{equation}\phantomsection\label{eq-div}{
\text{div}(\mathbf{x}^+,\mathbf{x}_{\text{CE}}^\prime,y^+;\theta) = \mathcal{E}_\theta(\mathbf{x}^+,y^+) - \mathcal{E}_\theta(\mathbf{x}_{\text{CE}}^\prime,y^+)
}\end{equation}

where \(\mathcal{E}_\theta(\cdot)\) denotes the energy function defined
as
\(\mathcal{E}_\theta(\mathbf{x},y^+)=-\mathbf{M}_\theta(\mathbf{x})[y^+]\),
with \(y^+\) denoting the index of the randomly drawn target class,
\(y^+ \sim p(y)\). Conditional on the target class \(y^+\),
\(\mathbf{x}_{\text{CE}}^\prime\) denotes a mature counterfactual for a
randomly sampled factual from a non-target class generated with a
gradient-based CE generator for up to \(T\) iterations. Intuitively, the
gradient of Equation~\ref{eq-div} decreases the energy of observed
training samples (positive samples) while increasing the energy of
counterfactuals (negative samples) (\citeproc{ref-du2019implicit}{Du and
Mordatch 2020}). As the counterfactuals get more plausible
(Definition~\ref{def-explainability}) during training, these opposing
effects gradually balance each other out
(\citeproc{ref-lippe2024uvadlc}{Lippe 2024}).

Since the maturity of counterfactuals in terms of a probability
threshold is often reached before \(T\), this form of sampling is not
only more closely aligned with Definition~\ref{def-explainability}., but
can also speed up training times compared to SGLD. The departure from
SGLD also allows us to tap into the vast repertoire of explainers that
have been proposed in the literature to meet different desiderata. For
example, many methods support domain and mutability constraints. In
principle, any approach for generating CEs is viable, so long as it does
not violate the faithfulness condition. Like JEMs
(\citeproc{ref-murphy2022probabilistic}{Murphy 2022}), counterfactual
training can be viewed as a form of contrastive representation learning.

\subsection{Indirectly Inducing Explainability: Adversarial
Robustness}\label{sec-indirectly-inducing-explainability-with-adversarial-robustness}

Based on our analysis in Section~\ref{sec-lit}, counterfactuals
\(\mathbf{x}^\prime\) can be repurposed as additional training samples
(\citeproc{ref-balashankar2023improving}{Balashankar et al. 2023};
\citeproc{ref-luu2023counterfactual}{Luu and Inoue 2023}) or adversarial
examples (\citeproc{ref-freiesleben2022intriguing}{Freiesleben 2022};
\citeproc{ref-pawelczyk2022exploring}{Pawelczyk et al. 2022}). This
leaves some flexibility with regards to the choice for the
\(\text{advloss}(\cdot)\) term in Equation~\ref{eq-obj}. An intuitive
functional form, but likely not the only sensible choice, is inspired by
adversarial training:

\begin{equation}\phantomsection\label{eq-adv}{
\begin{aligned}
\text{advloss}(\mathbf{M}_\theta(\mathbf{x}_{\text{AE}}^\prime),\mathbf{y};\varepsilon)&=\text{yloss}(\mathbf{M}_\theta(\mathbf{x}_{t_\varepsilon}^\prime),\mathbf{y}) \\
t_\varepsilon &= \max_t \{t: ||\Delta_t||_\infty < \varepsilon\}
\end{aligned}
}\end{equation}

Under this choice, we consider nascent counterfactuals
\(\mathbf{x}_{\text{AE}}^\prime\) as AEs as long as the magnitude of the
perturbation at time \(t\) (\(\Delta_t\)) to any single feature is at
most \(\varepsilon\). The most strongly perturbed counterfactual
\(\mathbf{x}_{t_\varepsilon}^\prime\) that still satisfies the condition
is used as an adversarial example \(\mathbf{x}_{\text{AE}}^\prime\).
This formalization is closely aligned with seminal work on adversarial
machine learning (\citeproc{ref-szegedy2013intriguing}{Szegedy et al.
2014}), which defines an adversarial attack as an ``imperceptible
non-random perturbation''. Thus, we work with a different distinction
between CE and AE than the one proposed in prior work
(\citeproc{ref-freiesleben2022intriguing}{Freiesleben 2022}), which
considers misclassification as the distinguishing feature of adversarial
examples. One of the key observations of our work is that we can
leverage CEs during training and get AEs essentially for free to reap
the benefits of adversarial training, leading to improved adversarial
robustness and plausibility.

\subsection{Encoding Actionability Constraints}\label{sec-constraints}

Many existing counterfactual explainers support domain and mutability
constraints. In fact, both types of constraints can be implemented for
any explainer that relies on gradient descent in the feature space for
optimization (\citeproc{ref-altmeyer2023explaining}{Altmeyer, Deursen,
and Liem 2023}). In this context, domain constraints can be imposed by
simply projecting counterfactuals back to the specified domain; if the
previous gradient step resulted in updated feature values that were
out-of-domain. Similarly, mutability constraints can be enforced by
setting partial derivatives to zero to ensure that features are only
perturbed in the allowed direction, if at all.

As actionability constraints are binding at test time, we must also
impose them when generating \(\mathbf{x}^\prime\) during each training
iteration to inform model representations. Through their effect on
\(\mathbf{x}^\prime\), both types of constraints influence model
outcomes via Equation~\ref{eq-div}. It is crucial that we avoid
penalizing implausibility that arises from mutability constraints. For
any mutability-constrained feature \(d\) this can be achieved by
enforcing \(\mathbf{x}^+[d] - \mathbf{x}^\prime[d]:=0\), whenever
perturbing \(\mathbf{x}^\prime[d]\) in the direction of
\(\mathbf{x}^+[d]\) would violate mutability constraints defined for
\(d\). Specifically, we set \(\mathbf{x}^+[d] := \mathbf{x}^\prime[d]\)
if:

\begin{enumerate}
\def\labelenumi{\arabic{enumi}.}
\item
  Feature \(d\) is strictly immutable in practice.
\item
  \(\mathbf{x}^+[d]>\mathbf{x}^\prime[d]\), but \(d\) can only be
  decreased in practice.
\item
  \(\mathbf{x}^+[d]<\mathbf{x}^\prime[d]\), but \(d\) can only be
  increased in practice.
\end{enumerate}

From a Bayesian perspective, setting
\(\mathbf{x}^+[d] := \mathbf{x}^\prime[d]\) can be understood as
assuming a point mass prior for \(p(\mathbf{x}^+)\) with respect to
feature \(d\), i.e., we can model this as absolute certainty that the
value \(\mathbf{x}^+[d]\) remains the same as in the neighbor,
\(\mathbf{x}^\prime[d]\), but it could be equivalently seen as masking
changes to feature \(d\). Intuitively, we can think of this as ignoring
implausibility costs of immutable features, which effectively forces the
model to instead seek plausibility through the remaining features. This
can be expected to produce a classifier with relatively lower
sensitivity to immutable features, and the higher relative sensitivity
to mutable features should make mutability-constrained recourse less
costly (see Section~\ref{sec-experiments}). Under certain conditions,
this result also holds theoretically (for the proof, see the
supplementary appendix):

\begin{proposition}[Protecting Immutable
Features]\protect\hypertarget{prp-mtblty}{}\label{prp-mtblty}

Let
\(f_\theta(\mathbf{x})=\mathcal{S}(\mathbf{M}_\theta(\mathbf{x}))=\mathcal{S}(\Theta\mathbf{x})\)
denote a linear classifier with softmax activation \(\mathcal{S}\) where
\(y\in\{1,...,K\}=\mathcal{K}\), \(\mathbf{x} \in \mathbb{R}^D\) and
\(\Theta\) is the matrix of coefficients with
\(\theta_{k,d}=\Theta[k,d]\) denoting the coefficient on feature \(d\)
for class \(k\). Assume multivariate Gaussian class densities with a
common diagonal covariance matrix \(\Sigma_k=\Sigma\) for all
\(k \in \mathcal{K}\), then protecting an immutable feature from the
contrastive divergence penalty will result in lower classifier
sensitivity to that feature relative to the remaining features, provided
that at least one of those is discriminative and mutable.

\end{proposition}

\section{Experiments}\label{sec-experiments}

We start by introducing the experimental setup, including performance
metrics, datasets, algorithms, and explain our approach to evaluation in
Section~\ref{sec-experimental-setup}. Then, we address the research
questions (RQ). Two questions relating to the principal goals of
counterfactual training are presented in
Section~\ref{sec-experimental-results}:

\begin{exercise}[]\protect\hypertarget{exr-plaus}{}\label{exr-plaus}

To what extent does the CT objective in Equation~\ref{eq-obj} induce
models to learn plausible explanations?

\end{exercise}

\begin{exercise}[]\protect\hypertarget{exr-act}{}\label{exr-act}

To what extent does CT result in more favorable algorithmic recourse
outcomes in the presence of actionability constraints

\end{exercise}

Next, in Section~\ref{sec-pred} we consider the performance of models
trained with CT, focusing on their adversarial robustness but also
commenting on the validity of generated CEs.

\begin{exercise}[]\protect\hypertarget{exr-ar}{}\label{exr-ar}

To what extent does CT influence the adversarial robustness of trained
models?

\end{exercise}

Finally, in Section~\ref{sec-ablation} we perform an ablation of the CT
objective and evaluate its sensitivity to hyperparameters:

\begin{exercise}[]\protect\hypertarget{exr-ablation}{}\label{exr-ablation}

How does the CT objective depends on its individual components?
(\emph{ablation})

\end{exercise}

\begin{exercise}[]\protect\hypertarget{exr-hyper}{}\label{exr-hyper}

What are the effects of hyperparameter selection on counterfactual
training?

\end{exercise}

\subsection{Experimental Setup}\label{sec-experimental-setup}

Our focus is the improvement in explainability
(Definition~\ref{def-explainability}). Thus, we mainly look at the
plausibility and cost of faithfully generated counterfactuals at test
time, but several other metrics are covered in the supplementary
appendix. To measure the cost, we follow the standard proxy of distances
(\(\ell_1\)-norm) between factuals and counterfactuals. For
plausibility, we assess how similar CEs are to observed samples in the
target domain, \(\mathbf{X}^+\subset\mathcal{X}^+\). For the evaluation,
we rely on the metric proposed in prior work
(\citeproc{ref-altmeyer2024faithful}{Altmeyer et al. 2024}) with
\(\ell_1\)-norm for distances,

\begin{equation}\phantomsection\label{eq-impl-dist}{
\text{IP}(\mathbf{x}^\prime,\mathbf{X}^+) = \frac{1}{\lvert\mathbf{X}^+\rvert}\sum_{\mathbf{x} \in \mathbf{X}^+} \text{dist}(\mathbf{x}^{\prime},\mathbf{x})
}\end{equation}

and introduce a novel divergence-based adaptation,

\begin{equation}\phantomsection\label{eq-impl-div}{
\text{IP}^*(\mathbf{X}^\prime,\mathbf{X}^+) = \text{MMD}(\mathbf{X}^\prime,\mathbf{X}^+)
}\end{equation}

where \(\mathbf{X}^\prime\) denotes a collection of counterfactuals and
\(\text{MMD}(\cdot)\) is the unbiased estimate of the squared population
maximum mean discrepancy (\citeproc{ref-gretton2012kernel}{Gretton et
al. 2012}):

\begin{equation}\phantomsection\label{eq-mmd}{
\begin{aligned}
\text{MMD}(\mathbf{X}^\prime,\mathbf{X}^+) &= \frac{1}{m(m-1)}\sum_{i=1}^m\sum_{j\neq i}^m k(x_i,x_j) \\ &+ \frac{1}{n(n-1)}\sum_{i=1}^n\sum_{j\neq i}^n k(\tilde{x}_i,\tilde{x}_j) \\ &- \frac{2}{mn}\sum_{i=1}^m\sum_{j=1}^n k(x_i,\tilde{x}_j)
\end{aligned}
}\end{equation}

with a kernel function \(k(\cdot,\cdot)\). We use a characteristic
Gaussian kernel with a constant length-scale parameter of \(0.5\), which
means that the metric in Equation~\ref{eq-impl-div} is equal to zero if
and only if the two distributions are exactly the same,
\(\mathbf{X}^\prime=\mathbf{X}^+\).

To assess outcomes with respect to actionability for non-linear models,
we look at the costs of (just) valid counterfactuals in terms of their
distances from factual starting points with \(\tau=0.5\). While this is
an imperfect proxy of sensitivity, we hypothesize that CT can reduce
these costs by teaching models to seek plausibility with respect to
mutable features, much like we observe in Figure~\ref{fig-poc} in panel
(d) compared to (c). We supplement this analysis with estimates using
integrated gradients (IG)
(\citeproc{ref-sundararajan2017ig}{Sundararajan, Taly, and Yan 2017}).
To evaluate predictive performance, we use standard metrics, such as
robust accuracy estimated on adversarially perturbed data using the fast
gradient sign method (FGSM)
(\citeproc{ref-goodfellow2014explaining}{Goodfellow, Shlens, and Szegedy
2015}) and projected gradient descent (PGD)
(\citeproc{ref-madry2017towards}{Madry et al. 2017}).

We make use of nine classification datasets common in the CE/AR
literature. Four of them are synthetic with two classes and different
characteristics: linearly separable Gaussian clusters (\emph{LS}),
overlapping clusters (\emph{OL}), concentric circles (\emph{Circ}), and
interlocking moons (\emph{Moon}). Next, we have four real-world binary
tabular datasets: \emph{Adult} (Census data)
(\citeproc{ref-becker1996adult2}{Becker and Kohavi 1996}), California
housing (\emph{CH}) (\citeproc{ref-pace1997sparse}{Pace and Barry
1997}), Default of Credit Card Clients (\emph{Cred})
(\citeproc{ref-yeh2016default}{Yeh 2016}), and Give Me Some Credit
(\emph{GMSC}) (\citeproc{ref-kaggle2011give}{Kaggle 2011}). Finally, for
convenient illustration, we use the 10-class \emph{MNIST}
(\citeproc{ref-lecun1998mnist}{LeCun 1998}).

We run experiments with three gradient-based generators: \emph{Generic}
(\citeproc{ref-wachter2017counterfactual}{Wachter, Mittelstadt, and
Russell 2017}) as a simple baseline; \emph{REVISE}
(\citeproc{ref-joshi2019realistic}{Joshi et al. 2019}) that aims to
generate plausible counterfactuals using a surrogate Variational
Autoencoder (VAE); and \emph{ECCCo}
(\citeproc{ref-altmeyer2024faithful}{Altmeyer et al. 2024}), targeting
faithfulness. In all cases, we use standard logit cross-entropy loss for
\(\text{yloss}(\cdot)\) and all generators penalize the distance
(\(\ell_1\)-norm) of counterfactuals from their original factual state.
\emph{Generic} and \emph{ECCCo} search for counterfactuals directly in
the feature space; \emph{REVISE} traverses the latent space of a
variational autoencoder (VAE) fitted to the training data, so its
outputs depend on the quality of the surrogate model. In addition to the
distance penalty, \emph{ECCCo} uses a penalty that regularizes the
energy associated with the counterfactual, \(\mathbf{x}^\prime\)
(\citeproc{ref-altmeyer2024faithful}{Altmeyer et al. 2024}). We omit the
conformal set size penalty proposed in the original paper, since the
authors found that faithfulness primarily depends on the energy penalty,
freeing us from one additional hyperparameter.

Our method does not aim to be agnostic to the underlying CE generator
and, as explained in
Section~\ref{sec-directly-inducing-explainability-with-contrastive-divergence},
the selection of the CE generator can impact the explainability of
models. To evaluate the specific value of counterfactual training, we
extensively test the method using the three above-mentioned CE
generators, which are characterized by varying complexity and
desiderata, and we present the complete results in the supplementary
appendix. Indeed, we observe that \emph{ECCCo} outclasses the other two
generators as the backbone of CT, generally leading to the highest
reduction in implausibility. This is not surprising; the goals of
\emph{ECCCo} most closely align with the objectives of CT: maximally
faithful explanations should also be the most useful for feedback.
Conversely, we cannot expect the model to learn much from counterfactual
explanations that largely depend on the quality of the surrogate model
that is trained for \emph{REVISE}. Similarly, \emph{Generic} is a very
simple baseline that optimizes only for minimal changes of features
(measured in the original seminal paper
(\citeproc{ref-wachter2017counterfactual}{Wachter, Mittelstadt, and
Russell 2017}) using median absolute deviation).

Thus, while counterfactual training can be used with any gradient-based
CE generator to improve the explainability of the resulting model, in
Section~\ref{sec-experimental-results} we mainly discuss its
effectiveness with \emph{ECCCo}, the strongest identified generator,
allowing us to optimize the quality of the models. This constitutes our
treatment method, but we still present the complete results for all
generators in the supplementary appendix.

To assess the effects of CT, we investigate the improvements in
performance metrics when using it on top of a weak baseline (BL), a
naively (conventionally) trained multilayer perceptron (\emph{MLP}), as
the control method. As we hold all other things constant, this is the
best way to get a clear picture of the improvement in explainability
that can be directly attributed to CT. It is also consistent with the
evaluation practices in the related literature
(\citeproc{ref-goodfellow2014explaining}{Goodfellow, Shlens, and Szegedy
2015}; \citeproc{ref-ross2021learning}{Ross, Lakkaraju, and Bastani
2024}; \citeproc{ref-teney2020learning}{Teney, Abbasnedjad, and Hengel
2020}).

We also note that counterfactual training involves multiple objectives
but our principal goal is high explainability as in
Definition~\ref{def-explainability}, while improved robustness is a
welcome byproduct. We neither aim to outperform state-of-the-art
approaches that target any single one of these objectives, nor do we
claim that CT can achieve this. Specifically, we do not aim to beat JEMs
with respect to their generative capacity, SOTA robust neural networks
with respect to (adversarial) robustness, or (quasi-)Bayesian neural
networks with respect to uncertainty quantification. As we have already
explained in Section~\ref{sec-lit}, existing literature has shown that
all of these objectives tend to correlate (explaining some of our
positive findings), but we situate counterfactual training squarely in
the context of (counterfactual) explainability and algorithmic recourse,
where it tackles an important shortcoming of existing approaches.

In terms of computing resources, all of our experiments were executed on
a high-performance cluster. We have relied on distributed computing
across multiple central processing units (CPU); for example, the
hyperparameter grid searches were carried out on 34 CPUs with 2GB memory
each. Graphical processing units (GPU) were \emph{not} used. All
computations were performed in the Julia Programming Language
(\citeproc{ref-bezanson2017julia}{Bezanson et al. 2017}); our code base
(algorithms and experimental settings) has been open-sourced on
GitHub.\footnote{https://github.com/JuliaTrustworthyAI/CounterfactualTraining.jl}
We explain more about the hardware, software, and reproducibility
considerations in the supplementary appendix.

\subsection{Main Results}\label{sec-experimental-results}

Our main results for plausibility and actionability for \emph{MLP}
models are summarized in Table~\ref{tbl-main} that presents
counterfactual outcomes grouped by dataset along with standard errors
averaged across bootstrap samples. Asterisks (\(^*\)) are used when the
bootstrapped 99\%-confidence interval of differences in mean outcomes
does \emph{not} include zero, so the observed effects are statistically
significant at the 0.01 level. As our experimental procedure is (by
virtue of the proposed method) relatively complex, we choose to work at
this stringent alpha level to demonstrate the high reliability of
counterfactual training.

The first two columns (\(\text{IP}\) and \(\text{IP}^*\)) show the
percentage reduction in implausibility for our two metrics when using CT
on top of the weak baseline. As an example, consider the first row for
\emph{LS} data: the observed positive values indicate that faithful
counterfactuals are around 26-51\% more plausible for models trained
with CT, in line with our observations in panel (b) of
Figure~\ref{fig-poc} compared to panel (a).

The third column shows the results for a scenario when mutability
constraints are imposed on the selected features. Again, we are
comparing CT to the baseline, so reductions in the positive direction
imply that valid counterfactuals are ``cheaper'' (more actionable) when
using CT with feature protection. Relating this back to
Figure~\ref{fig-poc}, the third column represents the reduction in
distances traveled by counterfactuals in panel (d) compared to panel
(c). In the following paragraphs, we summarize the results for all
datasets.

\begin{table}

\caption{\label{tbl-main}Key evaluation metrics for valid counterfactual
along with bootstrapped standard errors for all datasets.
\textbf{Plausibility} (columns 1-2): percentage reduction in
implausibility for \textbf{IP} and \textbf{IP}\(^*\), respectively;
\textbf{Cost} / \textbf{Actionability} (column 3): percentage reduction
in costs when selected features are protected. Outcomes are aggregated
across bootstrap samples (100 rounds) and varying degrees of the energy
penalty \(\lambda_{\text{egy}}\) used for ECCCo at test time. Asterisks
(\(^*\)) indicate that the bootstrapped 99\%-confidence interval of
differences in mean outcomes does \textbf{not} include zero.}

\centering{

\begin{tabular}{
  l
  S[table-format=2.2(1.2)]
  S[table-format=3.2(3.2)]
  S[table-format=3.2(1.2)]
}
  \toprule
  \textbf{Data} & \textbf{$ \text{IP} $ $(-\%)$} & \textbf{$ \text{IP}^* $ $(-\%)$} & \textbf{Cost $(-\%)$} \\\midrule
  LS & 26.26\pm0.67 $^{*}$ & 51.28\pm2.01 $^{*}$ & 16.41\pm0.57 $^{*}$ \\
  Circ & 58.88\pm0.37 $^{*}$ & 93.84\pm6.7 $^{*}$ & 42.99\pm0.85 $^{*}$ \\
  Moon & 19.59\pm0.73 $^{*}$ & 8.0\pm9.44 $^{}$ & 5.16\pm1.0 $^{*}$ \\
  OL & -1.93\pm1.12 $^{}$ & -27.7\pm14.59 $^{}$ & 40.86\pm2.3 $^{*}$ \\\midrule
  Adult & 0.19\pm1.05 $^{}$ & 34.35\pm5.61 $^{*}$ & 4.03\pm4.03 $^{}$ \\
  CH & 10.65\pm1.47 $^{*}$ & 63.06\pm4.25 $^{*}$ & 44.23\pm1.43 $^{*}$ \\
  Cred & 10.14\pm1.59 $^{*}$ & 50.35\pm12.26 $^{*}$ & -18.17\pm4.4 $^{*}$ \\
  GMSC & 10.65\pm2.28 $^{*}$ & 24.75\pm4.84 $^{*}$ & 66.01\pm1.41 $^{*}$ \\
  MNIST & 6.36\pm1.7 $^{*}$ & -70.31\pm217.6 $^{}$ & -35.11\pm6.96 $^{*}$ \\\midrule
  Avg. & 15.64 & 25.29 & 18.49 \\\bottomrule
\end{tabular}

}

\end{table}%

\subsubsection*{\texorpdfstring{Plausibility
(RQ~\ref{exr-plaus})}{Plausibility (RQ~)}}\label{sec-plaus}
\addcontentsline{toc}{subsubsection}{Plausibility (RQ~\ref{exr-plaus})}

\emph{CT generally produces substantial and statistically significant
improvements in plausibility.}

Average reductions in \(\text{IP}\) range from around 6\% for
\emph{MNIST} to almost 60\% for \emph{Circ}. For the real-world tabular
datasets they are around 10\% for \emph{CH}, \emph{Cred} and
\emph{GMSC}; for \emph{Adult} and \emph{OL} we find no significant
impact of CT on \(\text{IP}\). The former is subject to a large
proportion of categorical features, which inhibits the generation of
large numbers of valid counterfactuals during training and may therefore
explain this finding.

\begin{figure*}

\centering{

\includegraphics[width=1\linewidth,height=\textheight,keepaspectratio]{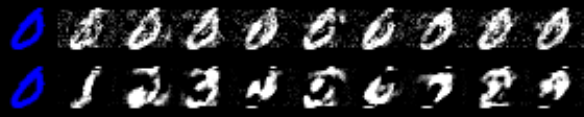}

}

\caption{\label{fig-mnist-ce}\emph{Plausibility}: BL (top row) vs CT
using the \emph{ECCCo} generator (bottom row) counterfactuals for a
randomly selected factual from class ``0'' (in blue). CT produces more
plausible counterfactuals than BL.}

\end{figure*}

Reductions in \(\text{IP}^*\) are even more substantial and generally
statistically significant, although the average degree of uncertainty is
higher than for \(\text{IP}\): reductions range from around 25\%
(\emph{GMSC}) to more than 90\% (\emph{Circ}). The only negative
findings are for \emph{OL} and \emph{MNIST}, but they are insignificant.
A qualitative inspection of the counterfactuals in
Figure~\ref{fig-mnist-ce} suggests recognizable digits for the model
trained with CT (bottom row), unlike the baseline (top row).

\begin{figure*}

\centering{

\includegraphics[width=1\linewidth,height=\textheight,keepaspectratio]{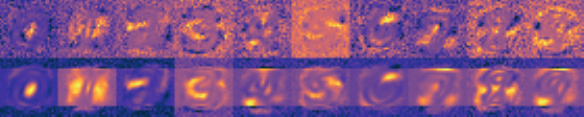}

}

\caption{\label{fig-mnist-ig}\emph{Actionability}: Sample visual
explanations (integrated gradients) for all classes in the \emph{MNIST}
dataset. Top and bottom rows of images show the results for BL and CT,
respectively. Mutability constraints are imposed on the five top and
five bottom rows of pixels. CT is less sensitive to protected features.}

\end{figure*}%

\subsubsection*{\texorpdfstring{Actionability
(RQ~\ref{exr-act})}{Actionability (RQ~)}}\label{sec-act}
\addcontentsline{toc}{subsubsection}{Actionability (RQ~\ref{exr-act})}

\emph{CT tends to improve actionability in the presence of immutable
features, but this is not guaranteed if the assumptions in
Proposition~\ref{prp-mtblty} are violated.}

For synthetic datasets, we always protect the first feature; for all
real-world tabular datasets we could identify and protect an \emph{age}
variable; for \emph{MNIST}, we protect the five top and five bottom rows
of pixels of the full image. Statistically significant reductions in
costs overwhelmingly point in the positive direction reaching up to
around 66\% for \emph{GMSC} data. Only in the case of \emph{Cred} and
\emph{MNIST}, average costs increase, most likely because any benefits
from protecting features are outweighed by an increase in costs required
for greater plausibility. With respect to \emph{MNIST} in particular,
the weak baseline is susceptible to cheap adversarial attacks that
significantly less costly to achieve that plausible counterfactuals.
Finally, the findings for \emph{Adult} are insignificant.

To further empirically evaluate the feature protection mechanism of CT
beyond linear models covered in Proposition~\ref{prp-mtblty}, we make
use of integrated gradients (IG)
(\citeproc{ref-sundararajan2017ig}{Sundararajan, Taly, and Yan 2017}).
IG calculates the contribution of each input feature towards a specific
prediction by approximating the integral of the model output with
respect to its input, using a set of samples that linearly interpolate
between a test instance and some baseline instance. This process
produces a vector of real numbers, one per input feature, which informs
about the contribution of each feature to the prediction. The selection
of an appropriate baseline is an important design decision
(\citeproc{ref-sundararajan2017ig}{Sundararajan, Taly, and Yan 2017});
to remain consistent in our evaluations, we use a baseline drawn at
random from the uniform distribution \(\mathcal{U}(-1,1)\) for all
datasets, which aligns with standard evaluation practices for IG. As the
outputs are not bounded (i.e., they are real numbers), we standardize
the integrated gradients across features to allow for a meaningful
comparison of the results for different models.

Qualitatively, the class-conditional integrated gradients in
Figure~\ref{fig-mnist-ig} suggest that CT has the expected effect even
for non-linear models: the model trained with CT (bottom row) is less
sensitive (blue) to the five top and five bottom rows of pixels that
were protected. Quantitatively, we observe substantial improvements for
seven out of nine datasets, and inconclusive results for the remaining
two datasets. Table~\ref{tbl-ig} shows the average sensitivity to
protected features measured by standardized integrated gradients for CT
and BL along with 95\% bootstrap confidence intervals: for the synthetic
datasets, we observe strong reductions in sensitivity to the protected
features for \emph{LS}, \emph{OL} and \emph{OL}, in line with
expectations. For the \emph{Moon} dataset, the effect of feature
protection is less pronounced but still in the expected direction. We
also observe that confidence intervals are in some cases much tighter
for models trained with CT: less noisy estimates for integrated
gradients likely indicate that the model is more regularized and can be
expected to behave more consistently across samples.

For real-world datasets, the sensitivity to the protected \emph{age}
variable is reduced by approximately a third for \emph{Adult}, 20\% for
\emph{CH}, and more than half for protected pixels in \emph{MNIST},
mirroring the qualitative findings in Figure~\ref{fig-mnist-ig}. In case
of \emph{Cred}, CT fully prevents the model from considering \emph{age}
as a factor in classification, with sensitivity reduced to zero. Only
for \emph{GMSC}, we observe negative impacts of CT, which we believe is
due to any or all of the following: a) data assumptions are violated; b)
the impact of other components of the CT objective outweighs expected
effects of feature protection; or c) the baseline choice applied
consistently to all datasets is not appropriate for \emph{GMSC}.

\begin{table}

\caption{\label{tbl-ig}Median sensitivity to protected features measured
by standardized integrated gradients. Square brackets enclose 95\%
bootstrap confidence intervals.}

\centering{

\begin{tabular}{l
    S[table-format=1.2]
    @{\quad[\,}S[table-format=1.2]@{,\,}S[table-format=2.2]@{\,]}
    S[table-format=5.2]
    @{\quad[\,}S[table-format=2.2]@{,\,}S[table-format=3.2]@{\,]}
}
\toprule
\textbf{Dataset} & \multicolumn{3}{c}{\textbf{CT}} & \multicolumn{3}{c}{\textbf{BL}} \\
\midrule
    LS & 0.21 & 0.20 & 0.22 & 30.69 & 12.92 & 629.20 \\[1ex]
    Circ & 6.96 & 4.88 & 20.62 & 19.20 & 6.48 & 193.92 \\[1ex]
    Moons & 0.54 & 0.41 & 0.68 & 0.66 & 0.53 & 0.92 \\[1ex]
    Over & 0.59 & 0.38 & 0.79 & 24.55 & 8.31 & 466.26 \\[1ex]
    \midrule
    Adult & 0.48 & 0.41 & 0.52 & 0.74 & 0.56 & 0.91 \\[1ex]
    CH & 0.04 & 0.01 & 0.06 & 0.05 & 0.03 & 0.09 \\[1ex]
    Cred & 0.00 & 0.00 & 0.00 & 0.20 & 0.18 & 0.25 \\[1ex]
    GMSC & 0.71 & 0.58 & 0.85 & 0.16 & 0.11 & 0.23 \\[1ex]
    MNIST & 0.17 & 0.16 & 0.17 & 0.35 & 0.33 & 0.37 \\[1ex]
    \bottomrule
\end{tabular}

}

\end{table}%

\begin{figure}

\centering{

\pandocbounded{\includegraphics[keepaspectratio]{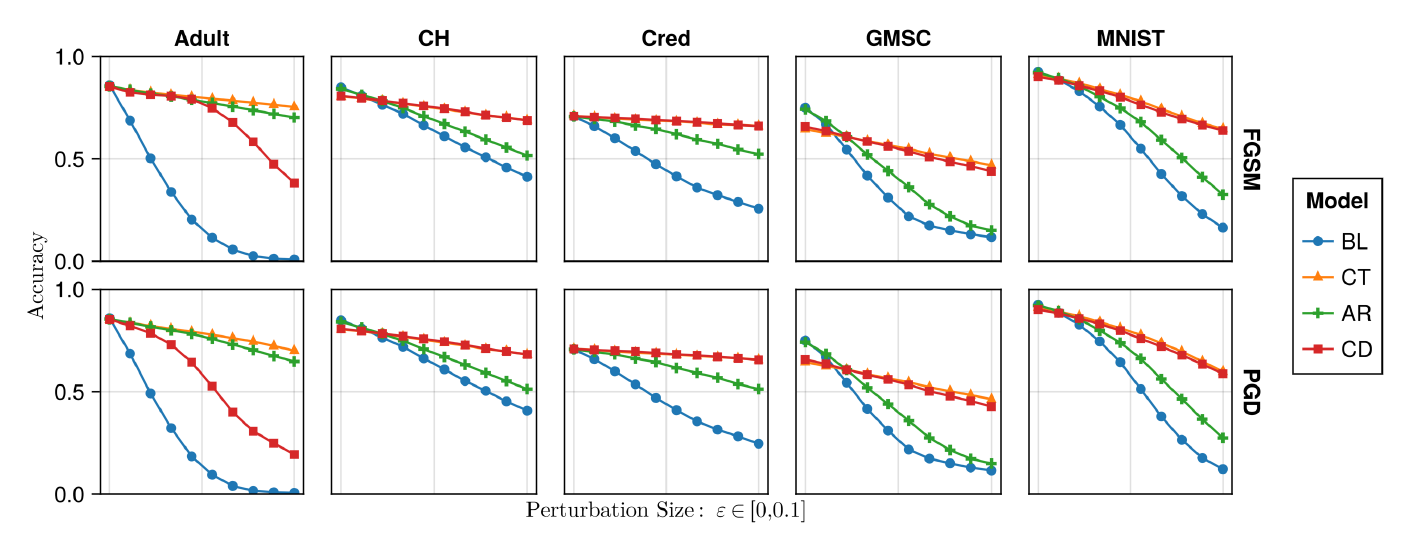}}

}

\caption{\label{fig-acc}Test accuracies on adversarially perturbed data
with varying perturbation sizes for the non-synthetic datasets.
Different training objectives are distinguished by color and shape: (1)
BL---the weak baseline; (2) CT---the full CT objective; (3) AR---a
partial CT objective without contrastive divergence; (4) CD---a partial
CT objective without adversarial loss. Top and bottom rows show the
results for FGSM and PGD (40 steps at step size \(\eta=0.01\)),
respectively.}

\end{figure}%

\subsection{Predictive Performance}\label{sec-pred}

\subsubsection*{\texorpdfstring{Adversarial Robustness
(RQ~\ref{exr-ar})}{Adversarial Robustness (RQ~)}}\label{sec-adversarial-robustness-rq3}
\addcontentsline{toc}{subsubsection}{Adversarial Robustness
(RQ~\ref{exr-ar})}

\emph{Models trained with CT are much more robust to gradient-based
adversarial attacks than conventionally-trained (weak) baselines.}

Test accuracies on clean and adversarially perturbed test data are shown
in Figure~\ref{fig-acc}. The perturbation size,
\(\varepsilon\in[0,0.1]\), increases along the horizontal axis, where
the case of \(\varepsilon=0\) corresponds to standard test accuracy for
non-perturbed data. For synthetic datasets, predictive performance is
virtually unaffected by perturbations for all models; those results are
therefore omitted from Figure~\ref{fig-acc} in favor of better
illustrations for the real-world data.

Focusing on the curves for CT and BL in Figure~\ref{fig-acc} for the
moment,\footnote{The results for AR and CD are discussed in the context
  of ablation below.} we find that standard test accuracy
(\(\varepsilon=0\)) is largely unaffected by CT, while robustness
against both types of attacks---FGSM (top row) and PGD (bottom row)---is
greatly improved: while in some cases robust accuracies for the weak
baseline drop to virtually zero (worse than random guessing) for large
enough perturbation sizes, accuracies of CT models remain remarkably
robust, even though robustness is not the primary objective of
counterfactual training. In the only case where standard accuracy on
unperturbed test data is substantially reduced for CT (\emph{GSMC}), we
note that robust accuracy decreases particularly fast for the weak
baseline as the perturbation size increases. This seems to indicate that
the standard accuracy for the weak baseline is inflated by sensitivity
to meaningless associations in the data.

We also look at the validity of generated counterfactuals, or the
proportion of counterfactuals that attain the target class, as presented
in Table~\ref{tbl-validity}. We find that in many cases CT leads to
substantial reductions in average validity, but this effect does not
seem to be strongly influenced by the imposed mutability constraints
(columns 1-2 vs columns 3-4). This result does not surprise us: by
design, CT shrinks the solution space for valid counterfactual
explanations, thus making it ``harder'' (and yet not ``more costly'') to
reach validity compared to the baseline model. As further discussed in
the supplementary appendix, this should not be seen as a shortcoming of
the method for a number of reasons: validity rates can be increased with
longer searches; costs of found solutions still generally decrease, as
we observe in our experiments; and achieving high validity does not
entail that explanations are practical for the recipients (e.g., valid
solutions may still be extremely costly)
(\citeproc{ref-venkatasubramanian2020philosophical}{Venkatasubramanian
and Alfano 2020}).

\begin{table}

\caption{\label{tbl-validity}Average validity of counterfactuals for CT
vs BL. First two columns correspond to no mutability constraints imposed
on the features; last two columns involve mutability constraints imposed
on the specified features.}

\centering{

\begin{tabular}{lSSSS}
  \toprule
  \textbf{Data} & \textbf{CT mut.} & \textbf{BL mut.} & \textbf{CT constr.} & \textbf{BL constr.} \\\midrule
  LS & 1.0 & 1.0 & 1.0 & 1.0 \\
  Circ & 1.0 & 0.51 & 0.71 & 0.48 \\
  Moon & 1.0 & 1.0 & 1.0 & 0.98 \\
  OL & 0.86 & 0.98 & 0.34 & 0.56 \\\midrule
  Adult & 0.68 & 0.99 & 0.7 & 0.99 \\
  CH & 1.0 & 1.0 & 1.0 & 1.0 \\
  Cred & 0.72 & 1.0 & 0.74 & 1.0 \\
  GMSC & 0.94 & 1.0 & 0.97 & 1.0 \\
  MNIST & 1.0 & 1.0 & 1.0 & 1.0 \\\midrule
  Avg. & 0.91 & 0.94 & 0.83 & 0.89 \\\bottomrule
\end{tabular}

}

\end{table}%

\subsection{Ablation and Hyperparameter Settings}\label{sec-ablation}

In this subsection, we use ablation studies to investigate how the
different components of the counterfactual training objective in
Equation~\ref{eq-obj} affect outcomes. Beyond this, we are also
interested in understanding how CT depends on various other
hyperparameters. To this end, we present the results from extensive grid
searches run across all synthetic datasets.

\subsubsection*{\texorpdfstring{Ablation
(RQ~\ref{exr-ablation})}{Ablation (RQ~)}}\label{sec-ablation-rq4}
\addcontentsline{toc}{subsubsection}{Ablation (RQ~\ref{exr-ablation})}

\emph{All components of the CT objective affect outcomes, even
independently, but the full objective achieves the most consistent
improvements wrt. our goals.}

We ablate the effect of both (1) the contrastive divergence component
and (2) the adversarial loss included in the full CT objective in
Equation~\ref{eq-obj}. In the following, we refer to the resulting
partial objectives as adversarial robustness (AR) and contrastive
divergence (CD), respectively. We note that AR corresponds to a form of
adversarial training and the CD objective is similar to that of a joint
energy-based model. Therefore, the ablation also serves as a comparison
of counterfactual training to stronger baselines, although we emphasize
again that we do not seek to outperform SOTA methods in the domains of
generative or robust machine learning, focusing CT squarely on models
with high explainability and actionability in the context of algorithmic
recourse.

Firstly, we find that both components play an important role in shaping
final outcomes. Both AR and CD can independently improve the
plausibility and adversarial robustness of models.

\begin{figure}

\centering{

\pandocbounded{\includegraphics[keepaspectratio]{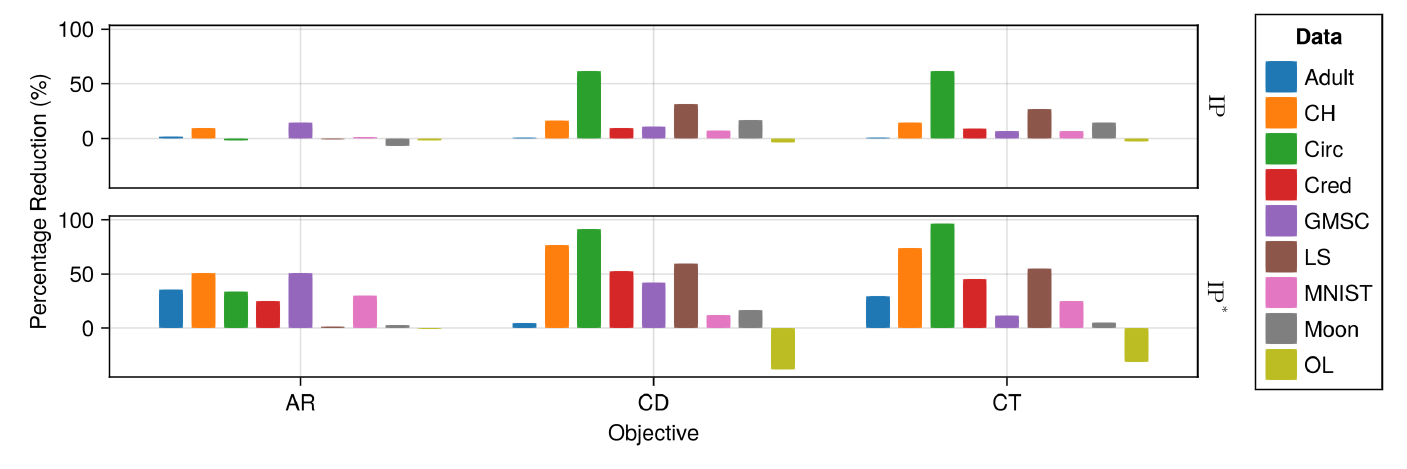}}

}

\caption{\label{fig-ablations}Percentage reductions in implausibility
for the partial (AR, CD) and full (CT) objectives compared to the weak
baseline. The results for \(\text{IP}\) and \(\text{IP}^*\) are shown in
the top and bottom graphs, respectively, and the datasets are
differentiated by color.}

\end{figure}

Concerning plausibility, Figure~\ref{fig-ablations} shows the percentage
reductions in implausibility for the partial and full objectives
compared to the weak baseline. The results for \(\text{IP}\) and
\(\text{IP}^*\) are shown in the top and bottom graphs, respectively,
and the datasets are differentiated by color. We find that in the best
identified hyperparameter settings, results for the full objective are
predominantly affected by the contrastive divergence component, but the
inclusion of adversarial loss leads to additional improvements for some
datasets (\emph{Adult}, \emph{MNIST}). We penalize contrastive
divergence twice as strongly as adversarial loss, which may explain why
the former dominates. The outcome for \emph{Adult}, in particular,
demonstrates the benefit of including both components: as noted earlier,
the large proportion of categorical features in this dataset seems to
inhibit the generation of valid counterfactuals, which in turn appears
to diminish the effect of the contrastive divergence component.

Looking at AR alone, we find that it produces mixed results for
\(\text{IP}\), with strong positive results nonetheless dominating
overall, reflecting previous findings from the related literature. In
particular, for real-world tabular datasets, adversarial robustness
seems to substantially benefit plausibility. In these cases, the
inclusion of the AR component in the full objective also helps to
substantially improve outcomes in relation to the partial CD objective:
improvements in plausibility for the \emph{Adult} and \emph{MNIST}
datasets are notably higher for full CT. In some cases---most notably
\emph{GMSC} and \emph{Cred}---the full CT objective does not outperform
the partial objectives, but still achieves the highest levels of
adversarial robustness (Figure~\ref{fig-acc}).

Zooming in on adversarial robustness, we find that the full CT objective
consistently outperforms the partial objectives, which individually
yield improvements. Consistent with the existing literature on JEMs
(\citeproc{ref-grathwohl2020your}{Grathwohl et al. 2020}), CD yields
substantially more robust models than the weak baseline at varying
perturbation sizes (Figure~\ref{fig-acc}). Similarly, AR yields
consistent improvements in robustness, as expected. Still, we observe
that in cases where either CD or AR show signs of degrading robust
accuracy at higher perturbation sizes, the full CT objective maintains
robustness. Much like in the context of plausibility, CT benefits from
both components, highlighting the effectiveness of our approach to
reusing nascent counterfactuals as AEs.

In summary, we find that the full CT objective strikes a balance between
both components, thereby leading to the most consistent improvements
with respect to plausibility and adversarial robustness.

\subsubsection*{\texorpdfstring{Hyperparameter settings
(RQ~\ref{exr-hyper})}{Hyperparameter settings (RQ~)}}\label{sec-hyperparameter-settings-rq5}
\addcontentsline{toc}{subsubsection}{Hyperparameter settings
(RQ~\ref{exr-hyper})}

\emph{CT is quite sensitive to the choice of a CE generator and its
hyperparameters but (1) we observe manageable patterns, and (2) we can
usually identify settings that improve either plausibility or
actionability, and typically both of them at the same time.}

We evaluate the impacts of three types of hyperparameters on CT. In the
following, we focus on the highlights and make the full results
available in the supplementary appendix.

Firstly, we find that optimal results are generally obtained when using
\emph{ECCCo} to generate counterfactuals. Conversely, using a generator
that may inhibit faithfulness (\emph{REVISE}), regularly yields smaller
improvements in plausibility and is more likely to even increase
implausibility. The results of the grid search for \emph{REVISE} also
exhibit higher variability than the results for \emph{ECCCo} and
\emph{Generic}. As argued above, this finding confirms our intuition
that maximally faithful explanations are most suitable for
counterfactual training.

Concerning hyperparameters that guide the gradient-based counterfactual
search, we find that increasing \(T\), the maximum number of steps,
generally yields better outcomes because more CEs can mature. Relatedly,
we also find that the effectiveness and stability of CT is positively
associated with the total number of counterfactuals generated during
each training epoch. The impact of \(\tau\), the decision threshold, is
more difficult to predict. On ``harder'' datasets it may be difficult to
satisfy high \(\tau\) for any given sample (i.e., also factuals) and so
increasing this threshold does not seem to correlate with better
outcomes. In fact, \(\tau=0.5\) generally leads to optimal results as it
is associated with high proportions of mature counterfactuals. This is
likely because the special case of \(\tau=0.5\) corresponds to equal
class probabilities, so a counterfactual is considered mature when the
logit for the target class is higher than the logits for all other
classes.

Secondly, the strength of the energy regularization,
\(\lambda_{\text{reg}}\), is highly impactful and should be set
sufficiently high to avoid common problems associated with exploding
gradients. The sensitivity with respect to \(\lambda_{\text{div}}\) and
\(\lambda_{\text{adv}}\) is much less evident. While high values of
\(\lambda_{\text{reg}}\) may increase the variability in outcomes when
combined with high values of \(\lambda_{\text{div}}\) or
\(\lambda_{\text{adv}}\), this effect is not particularly pronounced.
These results mirror our observations from the ablation studies and lend
further weight to the argument that CT benefits from both components.

Finally, we also observe desired improvements when CT was combined with
conventional training and employed only for the final 50\% of epochs of
the complete training process. Put differently, CT can improve the
explainability of models in a post-hoc, fine-tuning manner.

\section{Discussion}\label{sec-discussion}

As our results indicate, counterfactual training achieves its objective
of producing models that are more explainable. Nonetheless, these
advantages come with certain limitations.

\emph{Immutable features may have proxies.} We propose a method to
modify the sensitivity of a model to certain features, and thus increase
the actionability of the generated CEs. However, it requires that model
owners define the mutability constraints for (all) features considered
by the model. Even if all immutable features are protected, there may
exist proxies that are theoretically mutable (and hence should not be
protected) but preserve enough information about the principals to
hinder these protections. Delineating actionability is a major open
challenge in the AR literature (see, e.g.,
(\citeproc{ref-venkatasubramanian2020philosophical}{Venkatasubramanian
and Alfano 2020})) impacting the capacity of CT to fulfill its intended
goal.

\emph{Interventions on features may have implications for fairness.}
Modifying the sensitivity of a model to certain features may also have
implications for the fair and equitable treatment of decision subjects.
Model owners could misuse this solution by enforcing explanations based
on features that are more difficult to modify by some (group of)
decision subjects. For example, consider the \emph{Adult} dataset used
in our experiments, where \emph{workclass} or \emph{education} may be
more difficult to change for underprivileged groups. When applied
irresponsibly, CT could result in an unfairly assigned burden of
recourse (\citeproc{ref-sharma2020certifai}{Sharma, Henderson, and Ghosh
2020}), threatening the equality of opportunity in the system
(\citeproc{ref-bell2024fairness}{Bell et al. 2024}). Nonetheless, these
phenomena are not specific to CT.

\emph{Plausibility is costly.} As noted before, more plausible
counterfactuals are inevitably more costly
(\citeproc{ref-altmeyer2024faithful}{Altmeyer et al. 2024}). CT improves
plausibility and robustness, but this can negatively affect average
costs and validity whenever cheap, implausible, and adversarial
explanations are removed from the solution space.

\emph{CT increases training times.} Just like contrastive and robust
learning, CT is more resource-intensive than conventional regimes. Three
factors mitigate this effect: (1) CT yields itself to parallel
execution; (2) it amortizes the cost of CEs for the training samples;
and (3) our preliminary findings suggest that it can be used to
fine-tune conventionally-trained models.

We also highlight three key directions for future research. Firstly, it
is an interesting challenge to extend CT beyond classification settings.
Our formulation relies on the distinction between target and non-target
classes, requiring the output space to be discrete. Thus, it does not
apply to ML tasks where the change in outcome cannot be readily
discretized. Classification remains the focus of CE and algorithmic
recourse research; other settings have attracted some interest (e.g.,
regression (\citeproc{ref-spooner2021counterfactual}{Spooner et al.
2021})), but there is little consensus on how to extend the notion of
CEs.

Secondly, our analysis covers CE generators with different
characteristics, but it is interesting to extend it to more algorithms,
including ones that do not rely on computationally costly gradient-based
optimization. This should reduce training costs while possibly
preserving the benefits of CT.

Finally, we believe that it is possible to considerably improve
hyperparameter selection procedures. Our method benefits from the tuning
of certain key hyperparameters but we have relied exclusively on grid
searches. Future work on CT could benefit from more sophisticated
approaches. Notably, CT is iterative, which makes methods such as
Bayesian or gradient-based optimization applicable (see, e.g.,
(\citeproc{ref-bischl2023hyperparameter}{Bischl et al. 2023})).

\section{Conclusion}\label{sec-conclusion}

State-of-the-art machine learning models are prone to learning complex
representations that cannot be interpreted by humans. Existing work on
counterfactual explanations has largely focused on designing tools to
generate plausible and actionable explanations for any model. In this
work, we instead hold models accountable for delivering such
explanations. We introduce counterfactual training: a novel training
regime that integrates recent advances in contrastive learning,
adversarial robustness, and CE to incentivize highly explainable models.
Through theoretical results and extensive experiments, we demonstrate
that CT satisfies this goal while promoting adversarial robustness of
models. Explanations generated from CT-based models are both more
plausible (compliant with the underlying data-generating process) and
more actionable (compliant with user-specified mutability constraints),
and thus meaningful to recipients. In turn, our work highlights the
value of simultaneously improving models and their explanations.

\section*{Acknowledgment}\label{acknowledgment}
\addcontentsline{toc}{section}{Acknowledgment}

Some of the authors were partially funded by ICAI AI for Fintech
Research, an ING---TU Delft collaboration. Research reported in this
work was partially facilitated by computational resources and support of
the DelftBlue high-performance computing cluster at TU Delft
(\citeproc{ref-DHPC2022}{(DHPC) 2022}).

\section*{References}\label{references}
\addcontentsline{toc}{section}{References}

\phantomsection\label{refs}

\newpage{}

\begin{appendices}

\section{Notation}\label{notation}

\subsection{Variables and Parameters}\label{variables-and-parameters}

Below we provide an overview of some notation used frequently throughout
the paper:

\begin{itemize}
\tightlist
\item
  \(\mathcal{Y}\): The output domain.
\item
  \(y^+\): The target class and also the index of the target class.
\item
  \(y^-\): The non-target class and also the index of non-the target
  class.
\item
  \(\mathcal{X}\): The input domain.
\item
  \(\mathbf{x}\): a single training sample.
\item
  \(\mathbf{x}^\prime\): a counterfactual.
\item
  \(t=1,...,T\): Step indicator for counterfactual search iterations.
\item
  \(\mathbf{x}_{\text{AE}}^\prime\): a nascent counterfactual, defined
  as a counterfactual that has not yet converged.
\item
  \(\mathbf{x}_{\text{CE}}^\prime\): a mature counterfactual, defined as
  a counterfactual that has either passed a threshold probability or
  exhausted all \(T\) steps.
\item
  \(\mathbf{x}^+\): a training sample in the target class
  (ground-truth).
\item
  \(\mathbf{y}^+\): The one-hot encoded output vector for the target
  class.
\item
  \(\theta\): Model parameters (unspecified).
\item
  \(\Theta\): Matrix of parameters.
\item
  \(\mathbf{M}(\cdot)\): linear predictions (logits) of the classifier.
\end{itemize}

\subsection{Formulas}\label{formulas}

\subsubsection{Maximum Mean Discrepancy}\label{maximum-mean-discrepancy}

Maximum mean discrepancy is defined as follows,

\begin{equation}\phantomsection\label{eq-mmd}{
\begin{aligned}
\text{MMD}({X}^\prime,\tilde{X}^\prime) &= \frac{1}{m(m-1)}\sum_{i=1}^m\sum_{j\neq i}^m k(x_i,x_j) \\ &+ \frac{1}{n(n-1)}\sum_{i=1}^n\sum_{j\neq i}^n k(\tilde{x}_i,\tilde{x}_j) \\ &- \frac{2}{mn}\sum_{i=1}^m\sum_{j=1}^n k(x_i,\tilde{x}_j)
\end{aligned}
}\end{equation}

where \(k(\cdot,\cdot)\) is a kernel function
(\citeproc{ref-gretton2012kernel}{Gretton et al. 2012}). We make use of
a Gaussian kernel with a constant length-scale parameter of \(0.5\). In
our implementation, Equation~\ref{eq-mmd} is by default applied to the
entire subset of the training data for which \(y=y^+\).

\subsection{Other Conventions}\label{other-conventions}

In some place of this appendix, we use the terms \emph{full/Full}
(i.e.~the full CT objective) and \emph{vanilla/Vanilla} (i.e.~vanilla
training objective) to refer to models trained with counterfactual
training (\emph{CT}) and the baseline (\emph{BL}), respectively.

\section{Technical Details of Our
Approach}\label{technical-details-of-our-approach}

\subsection{Generating Counterfactuals through Gradient
Descent}\label{sec-app-ce}

In this section, we provide some additional background on gradient-based
counterfactual generators (Section~\ref{sec-app-ce-background}) and
discuss how we define convergence in this context
(Section~\ref{sec-app-conv}).

\subsubsection{Background}\label{sec-app-ce-background}

Gradient-based counterfactual search was originally proposed by Wachter,
Mittelstadt, and Russell
(\citeproc{ref-wachter2017counterfactual}{2017}). A more general version
of the counterfactual search objective presented in the paper is as
follows,

\[
\begin{aligned}
\min_{\mathbf{z}^\prime \in \mathcal{Z}^L} \left\{  {\text{yloss}(\mathbf{M}_{\theta}(g(\mathbf{z}^\prime)),\mathbf{y}^+)}+ \lambda {\text{reg}(g(\mathbf{z}^\prime)) }  \right\} 
\end{aligned} 
\]

where \(g: \mathcal{Z} \mapsto \mathcal{X}\) is an invertible function
that maps from the \(L\)-dimensional counterfactual state space to the
feature space and \(\text{reg}(\cdot)\) denotes one or more penalties
that are used to induce certain properties of the counterfactual
outcome. As above, \(\mathbf{y}^+\) denotes the target output and
\(\mathbf{M}_{\theta}(\mathbf{x})\) returns the logit predictions of the
underlying classifier for \(\mathbf{x}=g(\mathbf{z})\).

For all generators used in this work we use standard logit crossentropy
loss for \(\text{yloss}(\cdot)\). All generators also penalize the
distance (\(\ell_1\)-norm) of counterfactuals from their original
factual state. For \emph{Generic} and \emph{ECCCo}, we have
\(\mathcal{Z}:=\mathcal{X}\) and
\(g(\mathbf{z})=g(\mathbf{z})^{-1}=\mathbf{z}\), that is counterfactual
are searched directly in the feature space. Conversely, \emph{REVISE}
traverses the latent space of a variational autoencoder (VAE) fitted to
the training data, where \(g(\cdot)\) corresponds to the decoder
(\citeproc{ref-joshi2019realistic}{Joshi et al. 2019}). In addition to
the distance penalty, \emph{ECCCo} uses a penalty that regularizes the
energy associated with the counterfactual, \(\mathbf{x}^\prime\)
(\citeproc{ref-altmeyer2024faithful}{Altmeyer et al. 2024}). We omit the
conformal set size penalty proposed in the original paper, since a) the
authors found faithfulness to primarily depend on the energy penalty and
hence this alleviates us from one additional hyperparameter.

\subsubsection{Convergence}\label{sec-app-conv}

An important consideration when generating counterfactual explanations
using gradient-based methods is how to define convergence. Two common
choices are to 1) perform gradient descent over a fixed number of
iterations \(T\), or 2) conclude the search as soon as the predicted
probability for the target class has reached a pre-determined threshold,
\(\tau\):
\(\mathcal{S}(\mathbf{M}_\theta(\mathbf{x}^\prime))[y^+] \geq \tau\). We
prefer the latter for our purposes, because it explicitly defines
convergence in terms of the black-box model, \(\mathbf{M}(\mathbf{x})\).

Defining convergence in this way allows for a more intuitive
interpretation of the resulting counterfactual outcomes than with fixed
\(T\). Specifically, it allows us to think of counterfactuals as
explaining `high-confidence' predictions by the model for the target
class \(y^+\). Depending on the context and application, different
choices of \(\tau\) can be considered as representing `high-confidence'
predictions.

\subsection{Protecting Mutability Constraints with Linear
Classifiers}\label{sec-app-constraints}

In the main paper, we explain that to avoid penalizing implausibility
that arises due to mutability constraints, we impose a point mass prior
on \(p(\mathbf{x})\) for the corresponding feature. We argue that this
approach induces models to be relatively less sensitive to immutable
features, propose a theoretical result supporting this and provide
empirical evidence that strengthens our argument (both in the main paper
and additional findings in this appendix). Below we derive the
analytical results in Proposition in the main paper.

\begin{proof}
Let \(d_{\text{mtbl}}\) and \(d_{\text{immtbl}}\) denote some mutable
and immutable feature, respectively. Suppose that
\(\mu_{y^-,d_{\text{immtbl}}} < \mu_{y^+,d_{\text{immtbl}}}\) and
\(\mu_{y^-,d_{\text{mtbl}}} > \mu_{y^+,d_{\text{mtbl}}}\), where
\(\mu_{k,d}\) denotes the conditional sample mean of feature \(d\) in
class \(k\). In words, we assume that the immutable feature tends to
take lower values for samples in the non-target class \(y^-\) than in
the target class \(y^+\). We assume the opposite to hold for the mutable
feature.

Assuming multivariate Gaussian class densities with common diagonal
covariance matrix \(\Sigma_k=\Sigma\) for all \(k \in \mathcal{K}\), we
have for the log likelihood ratio between any two classes
\(k,m \in \mathcal{K}\) (\citeproc{ref-hastie2009elements}{Hastie,
Tibshirani, and Friedman 2009}):

\begin{equation}\phantomsection\label{eq-loglike}{
\log \frac{p(k|\mathbf{x})}{p(m|\mathbf{x})}=\mathbf{x}^\intercal \Sigma^{-1}(\mu_{k}-\mu_{m})  + \text{const}
}\end{equation}

By independence of \(x_1,...,x_D\), the full log-likelihood ratio
decomposes into:

\begin{equation}\phantomsection\label{eq-loglike-decomp}{
\log \frac{p(k|\mathbf{x})}{p(m|\mathbf{x})} = \sum_{d=1}^D \frac{\mu_{k,d}-\mu_{m,d}}{\sigma_{d}^2} x_{d} + \text{const}
}\end{equation}

By the properties of our classifier (\emph{multinomial logistic
regression}), we have:

\begin{equation}\phantomsection\label{eq-multi}{
\log \frac{p(k|\mathbf{x})}{p(m|\mathbf{x})} = \sum_{d=1}^D \left( \theta_{k,d} - \theta_{m,d} \right)x_d + \text{const}
}\end{equation}

where \(\theta_{k,d}=\Theta[k,d]\) denotes the coefficient on feature
\(d\) for class \(k\).

Based on Equation~\ref{eq-loglike-decomp} and Equation~\ref{eq-multi} we
can identify that
\((\mu_{k,d}-\mu_{m,d}) \propto (\theta_{k,d} - \theta_{m,d})\) under
the assumptions we made above. Hence, we have that
\((\theta_{y^-,d_{\text{immtbl}}} - \theta_{y^+,d_{\text{immtbl}}}) < 0\)
and
\((\theta_{y^-,d_{\text{mtbl}}} - \theta_{y^+,d_{\text{mtbl}}}) > 0\).

Let \(\mathbf{x}^\prime\) denote some randomly chosen individual from
class \(y^-\) and let \(y^+ \sim p(y)\) denote the randomly chosen
target class. Then the partial derivative of the contrastive divergence
penalty with respect to coefficient \(\theta_{y^+,d}\) is equal to

\begin{equation}\phantomsection\label{eq-grad}{
\frac{\partial}{\partial\theta_{y^+,d}} \left(\text{div}(\mathbf{x}^+,\mathbf{x^\prime},\mathbf{y};\theta)\right) = \frac{\partial}{\partial\theta_{y^+,d}} \left( \left(-\mathbf{M}_\theta(\mathbf{x}^+)[y^+]\right) - \left(-\mathbf{M}_\theta(\mathbf{x}^\prime)[y^+]\right) \right) = x_{d}^\prime - x^+_{d}
}\end{equation}

and equal to zero everywhere else.

Since \((\mu_{y^-,d_{\text{immtbl}}} < \mu_{y^+,d_{\text{immtbl}}})\) we
are more likely to have
\((x_{d_{\text{immtbl}}}^\prime - x^+_{d_{\text{immtbl}}}) < 0\) than
vice versa at initialization. Similarly, we are more likely to have
\((x_{d_{\text{mtbl}}}^\prime - x^+_{d_{\text{mtbl}}}) > 0\) since
\((\mu_{y^-,d_{\text{mtbl}}} > \mu_{y^+,d_{\text{mtbl}}})\).

This implies that if we do not protect feature \(d_{\text{immtbl}}\),
the contrastive divergence penalty will decrease
\(\theta_{y^-,d_{\text{immtbl}}}\) thereby exacerbating the existing
effect
\((\theta_{y^-,d_{\text{immtbl}}} - \theta_{y^+,d_{\text{immtbl}}}) < 0\).
In words, not protecting the immutable feature would have the
undesirable effect of making the classifier more sensitive to this
feature, in that it would be more likely to predict class \(y^-\) as
opposed to \(y^+\) for lower values of \(d_{\text{immtbl}}\).

By the same rationale, the contrastive divergence penalty can generally
be expected to increase \(\theta_{y^-,d_{\text{mtbl}}}\) exacerbating
\((\theta_{y^-,d_{\text{mtbl}}} - \theta_{y^+,d_{\text{mtbl}}}) > 0\).
In words, this has the effect of making the classifier more sensitive to
the mutable feature, in that it would be more likely to predict class
\(y^-\) as opposed to \(y^+\) for higher values of \(d_{\text{mtbl}}\).

Thus, our proposed approach of protecting feature \(d_{\text{immtbl}}\)
has the net affect of decreasing the classifier's sensitivity to the
immutable feature relative to the mutable feature (i.e.~no change in
sensitivity for \(d_{\text{immtbl}}\) relative to increased sensitivity
for \(d_{\text{mtbl}}\)).
\end{proof}

\subsection{Domain Constraints}\label{domain-constraints}

We apply domain constraints on counterfactuals during training and
evaluation. There are at least two good reasons for doing so. Firstly,
within the context of explainability and algorithmic recourse,
real-world attributes are often domain constrained: the \emph{age}
feature, for example, is lower bounded by zero and upper bounded by the
maximum human lifespan. Secondly, domain constraints help mitigate
training instabilities commonly associated with energy-based modelling
(\citeproc{ref-grathwohl2020your}{Grathwohl et al. 2020};
\citeproc{ref-altmeyer2024faithful}{Altmeyer et al. 2024}).

For our image datasets, features are pixel values and hence the domain
is constrained by the lower and upper bound of values that pixels can
take depending on how they are scaled (in our case \([-1,1]\)). For all
other features \(d\) in our synthetic and tabular datasets, we
automatically infer domain constraints
\([x_d^{\text{LB}},x_d^{\text{UB}}]\) as follows,

\begin{equation}\phantomsection\label{eq-domain}{
\begin{aligned}
x_d^{\text{LB}} &= \arg\min_{x_d} \{\mu_d - n_{\sigma_d}\sigma_d, \arg \min_{x_d} x_d\} \\
x_d^{\text{UB}} &= \arg\max_{x_d} \{\mu_d + n_{\sigma_d}\sigma_d, \arg \max_{x_d} x_d\} 
\end{aligned}
}\end{equation}

where \(\mu_d\) and \(\sigma_d\) denote the sample mean and standard
deviation of feature \(d\). We set \(n_{\sigma_d}=3\) across the board
but higher values and hence wider bounds may be appropriate depending on
the application.

\subsection{Training Hyperparameters}\label{sec-app-training}

Note~\ref{nte-train-default} presents the default hyperparameters used
during training.

\begin{tcolorbox}[enhanced jigsaw, arc=.35mm, colback=white, bottomrule=.15mm, toptitle=1mm, leftrule=.75mm, titlerule=0mm, toprule=.15mm, breakable, colframe=quarto-callout-note-color-frame, left=2mm, opacityback=0, bottomtitle=1mm, opacitybacktitle=0.6, title={Note \ref*{nte-train-default}space Training Phase}, colbacktitle=quarto-callout-note-color!10!white, rightrule=.15mm, coltitle=black]

\quartocalloutnte{nte-train-default} 

\begin{itemize}
\tightlist
\item
  Meta Parameters:

  \begin{itemize}
  \tightlist
  \item
    Generator: \texttt{ecco}
  \item
    Model: \texttt{mlp}
  \end{itemize}
\item
  Model:

  \begin{itemize}
  \tightlist
  \item
    Activation: \texttt{relu}
  \item
    No.~Hidden: \texttt{32}
  \item
    No.~Layers: \texttt{1}
  \end{itemize}
\item
  Training Parameters:

  \begin{itemize}
  \tightlist
  \item
    Burnin: \texttt{0.0}
  \item
    Class Loss: \texttt{logitcrossentropy}
  \item
    Convergence: \texttt{threshold}
  \item
    Generator Parameters:

    \begin{itemize}
    \tightlist
    \item
      Decision Threshold: \texttt{0.75}
    \item
      \(\lambda_{\text{cst}}\): \texttt{0.001}
    \item
      \(\lambda_{\text{egy}}\): \texttt{5.0}
    \item
      Learning Rate: \texttt{0.25}
    \item
      Maximum Iterations: \texttt{30}
    \item
      Optimizer: \texttt{sgd}
    \item
      Type: \texttt{ECCo}
    \end{itemize}
  \item
    \(\lambda_{\text{adv}}\): \texttt{0.25}
  \item
    \(\lambda_{\text{clf}}\): \texttt{1.0}
  \item
    \(\lambda_{\text{div}}\): \texttt{0.5}
  \item
    \(\lambda_{\text{reg}}\): \texttt{0.1}
  \item
    Learning Rate: \texttt{0.001}
  \item
    No.~Counterfactuals: \texttt{1000}
  \item
    No.~Epochs: \texttt{100}
  \item
    Objective: \texttt{full}
  \item
    Optimizer: \texttt{adam}
  \end{itemize}
\end{itemize}

\end{tcolorbox}

\subsection{Evaluation Details}\label{sec-app-eval}

\subsubsection{Counterfactual Outcomes}\label{counterfactual-outcomes}

For all of our counterfactual evaluations, we proceed as follows: for
each dataset we run \(J\) bootstrap rounds (``No.~Runs'') to account for
stochasticity (Note~\ref{nte-eval-default}); for each bootstrap round,
we randomly draw factual and target pairs; then, for each model, we draw
samples from the test set (with replacement) for which the model
predicts the randomly chosen factual class; finally, we generate
multiple counterfactuals (``No.~Counterfactuals'') and evaluate the
outcomes (Note~\ref{nte-eval-default}). This is in line with standard
practice in the related literature on CE (see e.g. Schut et al.
(\citeproc{ref-schut2021generating}{2021})). For our final results
presented in the main paper, we rely on held-out test sets for
evaluation. For tuning purposes we rely on training and/or validation
sets.

Note~\ref{nte-eval-default} presents the default hyperparameters used
during evaluation for tuning purposes. For the main results presented in
the paper, we use larger evaluations, specifically:

\begin{itemize}
\tightlist
\item
  ``No.~Runs'': We set the number of bootstrap rounds to \(J=100\) for
  all datasets.
\item
  ``No.~Individuals'': In each round we draw 1,250, 500 and 125 samples
  for synthetic datasets, real-world tabular datasets and \emph{MNIST},
  respectively, across five different values for the strength of the
  energy penalty of \emph{ECCCo} at test time,
  \(\lambda_{\text{egy}}\in\{0.1, 0.5, 1.0, 5.0, 10.0\}\).
\end{itemize}

\begin{tcolorbox}[enhanced jigsaw, arc=.35mm, colback=white, bottomrule=.15mm, toptitle=1mm, leftrule=.75mm, titlerule=0mm, toprule=.15mm, breakable, colframe=quarto-callout-note-color-frame, left=2mm, opacityback=0, bottomtitle=1mm, opacitybacktitle=0.6, title={Note \ref*{nte-eval-default}space Evaluation Phase}, colbacktitle=quarto-callout-note-color!10!white, rightrule=.15mm, coltitle=black]

\quartocalloutnte{nte-eval-default} 

\begin{itemize}
\tightlist
\item
  Convergence: \texttt{threshold}
\item
  Decision Threshold: \texttt{0.95}
\item
  Maximum Iterations: \texttt{50}
\item
  No.~Individuals: \texttt{100}
\item
  No.~Runs: \texttt{5}
\end{itemize}

\end{tcolorbox}

\subsubsection{Predictive Performance}\label{predictive-performance}

To assess (robust) predictive performance, we evaluate model accuracy on
(adversarially perturbed) test data. To generate adversarial examples we
use the Fast Gradient Sign Method (FGSM)
(\citeproc{ref-goodfellow2014explaining}{Goodfellow, Shlens, and Szegedy
2015}). For the main results in the paper, we choose a range of values
\(\epsilon=[0.0,0.1]\). In some places of this appendix, you will also
find predictive performance evaluations in terms of the F1-score.

\FloatBarrier

\section{Details on Main Experiments}\label{sec-app-main}

\subsection{Final Hyperparameters}\label{final-hyperparameters}

As discussed the main paper, CT is sensitive to certain hyperparameter
choices. We study the effect of many hyperparameters extensively in
Section~\ref{sec-app-grid} of this appendix. For the main results, we
tune a small set of key hyperparameters (Section~\ref{sec-app-tune}).
The final choices for the main results are presented for each data set
in Table~\ref{tbl-final-params} along with training, test and batch
sizes.

\begin{table}

\caption{\label{tbl-final-params}Final hyperparameters used for the main
results presented in the main paper. Any hyperparameter not shown here
is set to its default value (Note~\ref{nte-train-default}).}

\centering{

\begin{tabular}{cccccccc}
  \toprule
  \textbf{Data} & \textbf{No. Train} & \textbf{No. Test} & \textbf{Batchsize} & \textbf{Domain} & \textbf{Decision Threshold} & \textbf{No. Counterfactuals} & \textbf{$\lambda_{\text{reg}}$} \\\midrule
  LS & 3600 & 600 & 30 & none & 0.5 & 1000 & 0.01 \\
  Circ & 3600 & 600 & 30 & none & 0.5 & 1000 & 0.5 \\
  Moon & 3600 & 600 & 30 & none & 0.9 & 1000 & 0.25 \\
  OL & 3600 & 600 & 30 & none & 0.5 & 1000 & 0.25 \\\midrule
  Adult & 26049 & 5010 & 1000 & none & 0.75 & 5000 & 0.25 \\
  CH & 16504 & 3101 & 1000 & none & 0.5 & 5000 & 0.25 \\
  Cred & 10617 & 1923 & 1000 & none & 0.5 & 5000 & 0.25 \\
  GMSC & 13371 & 2474 & 1000 & none & 0.5 & 5000 & 0.5 \\
  MNIST & 11000 & 2000 & 1000 & (-1.0, 1.0) & 0.5 & 5000 & 0.01 \\\bottomrule
\end{tabular}

}

\end{table}%

\subsubsection{Confidence Intervals}\label{confidence-intervals}

Table~\ref{tbl-ci} present the exact confidence intervals (99\%) for the
difference in mean outcomes on which we base our assessment of
statistical significance in the main paper. Grouped by evaluation
metrics (Variable) and dataset (Data), the table presents the mean
outcomes for CT and BT and finally the lower bound (LB) and upper bound
(UB) of the confidence interval. To compute the intervals, we used the
percentile method for bootstrapped confidence intervals: the lower and
upper bound represent the \(\alpha/2\)- and \((1-\alpha / 2)-\)quantile
of the bootstrap distribution, respectively, for \(\alpha=0.01\).

\begin{table}

\caption{\label{tbl-ci}Mean outcomes for CT and BL along with
bootstrapped confidence intervals (99\%) for difference in mean outcomes
grouped by dataset and evaluation metric. Column LB and UB show the
lower and upper bound of the intervals, respectively, and computed using
the percentile method (for significance, interval should not include
zero). The underlying counterfactual evaluations are the same as the
ones used to produce the main table in the paper.}

\centering{

\begin{tabular}{lccccc}
  \toprule
  \textbf{Variable} & \textbf{Data} & \textbf{CT} & \textbf{BL} & \textbf{LB} & \textbf{UB} \\\midrule
  Cost & Adult & 2.19 & 2.28 & -0.32 & 0.11 \\
  Cost & CH & 1.37 & 2.46 & -1.18 & -1.0 \\
  Cost & Circ & 0.7 & 1.22 & -0.55 & -0.49 \\
  Cost & Cred & 2.7 & 2.29 & 0.16 & 0.6 \\
  Cost & GMSC & 1.03 & 3.04 & -2.37 & -1.86 \\
  Cost & LS & 3.75 & 4.48 & -0.8 & -0.67 \\
  Cost & MNIST & 72.08 & 53.42 & 11.15 & 26.68 \\
  Cost & Moon & 1.52 & 1.6 & -0.12 & -0.05 \\
  Cost & OL & 1.55 & 2.62 & -1.25 & -0.9 \\
  $ \text{IP}^* $ & Adult & 0.07 & 0.11 & -0.06 & -0.02 \\
  $ \text{IP}^* $ & CH & 0.02 & 0.06 & -0.05 & -0.03 \\
  $ \text{IP}^* $ & Circ & 0.0 & 0.0 & -0.01 & -0.0 \\
  $ \text{IP}^* $ & Cred & 0.03 & 0.06 & -0.05 & -0.01 \\
  $ \text{IP}^* $ & GMSC & 0.05 & 0.07 & -0.02 & -0.01 \\
  $ \text{IP}^* $ & LS & 0.11 & 0.23 & -0.13 & -0.11 \\
  $ \text{IP}^* $ & MNIST & 0.02 & 0.02 & -0.07 & 0.07 \\
  $ \text{IP}^* $ & Moon & 0.02 & 0.02 & -0.01 & 0.0 \\
  $ \text{IP}^* $ & OL & 0.12 & 0.09 & -0.01 & 0.05 \\
  $ \text{IP} $ & Adult & 15.13 & 15.16 & -0.42 & 0.39 \\
  $ \text{IP} $ & CH & 6.72 & 7.52 & -1.05 & -0.6 \\
  $ \text{IP} $ & Circ & 0.97 & 2.36 & -1.44 & -1.35 \\
  $ \text{IP} $ & Cred & 19.79 & 22.02 & -3.17 & -1.42 \\
  $ \text{IP} $ & GMSC & 7.24 & 8.1 & -1.26 & -0.38 \\
  $ \text{IP} $ & LS & 2.51 & 3.4 & -0.95 & -0.84 \\
  $ \text{IP} $ & MNIST & 261.05 & 278.84 & -27.38 & -7.51 \\
  $ \text{IP} $ & Moon & 1.37 & 1.71 & -0.36 & -0.3 \\
  $ \text{IP} $ & OL & 4.52 & 4.44 & -0.03 & 0.19 \\\bottomrule
\end{tabular}

}

\end{table}%

\subsubsection{Qualitative Findings for Image
Data}\label{qualitative-findings-for-image-data}

\begin{figure}

\begin{minipage}{0.46\linewidth}

\begin{figure}[H]

\centering{

\pandocbounded{\includegraphics[keepaspectratio]{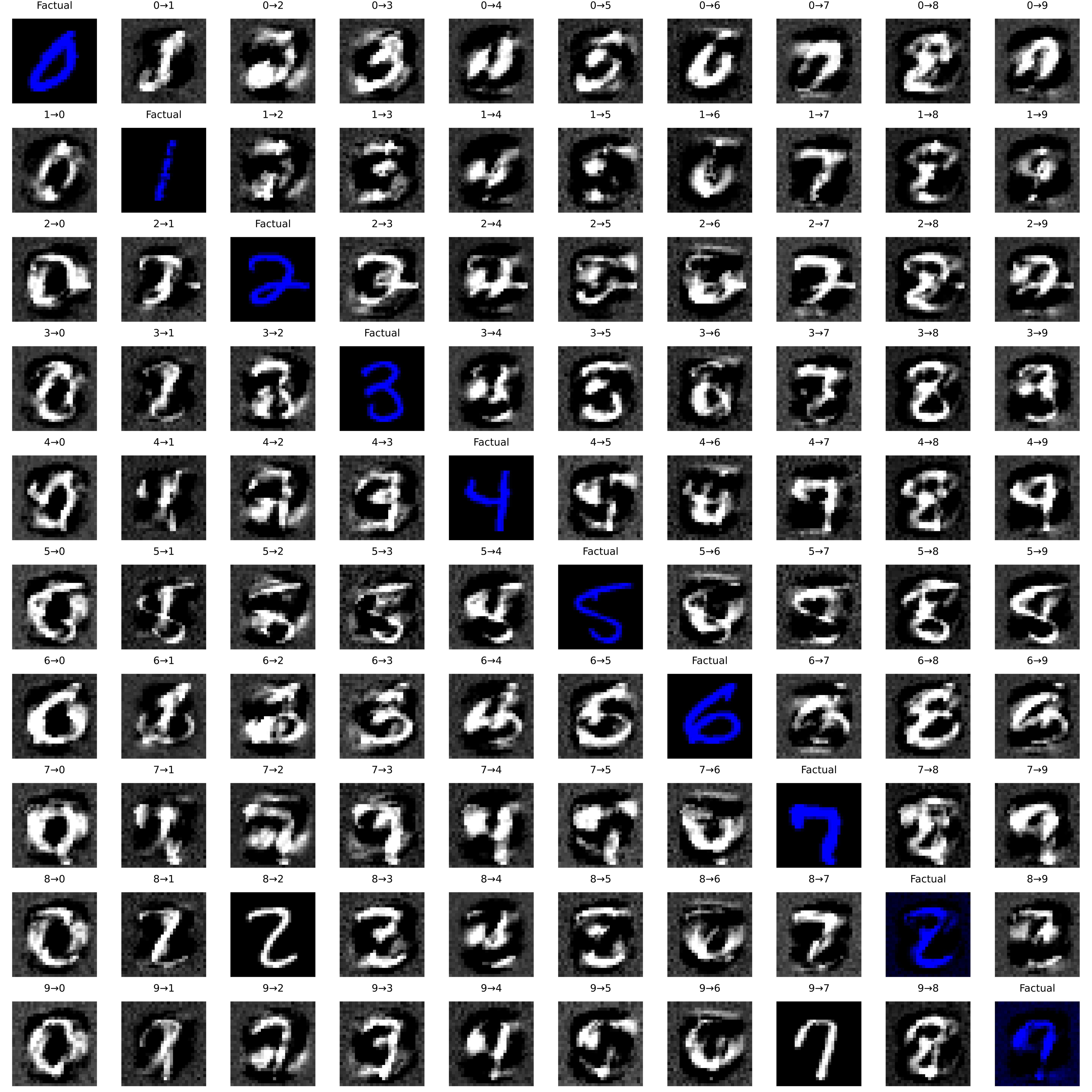}}

}

\caption{\label{fig-mnist}Counterfactual images for \emph{MLP} with
counterfactual training. Factual images are shown on the diagonal, with
the corresponding counterfactual for each target class (columns) in that
same row. The underlying generator, \emph{ECCCo}, aims to generate
counterfactuals that are faithful to the model
(\citeproc{ref-altmeyer2024faithful}{Altmeyer et al. 2024}).}

\end{figure}%

\end{minipage}%
\begin{minipage}{0.09\linewidth}
~\end{minipage}%
\begin{minipage}{0.46\linewidth}

\begin{figure}[H]

\centering{

\pandocbounded{\includegraphics[keepaspectratio]{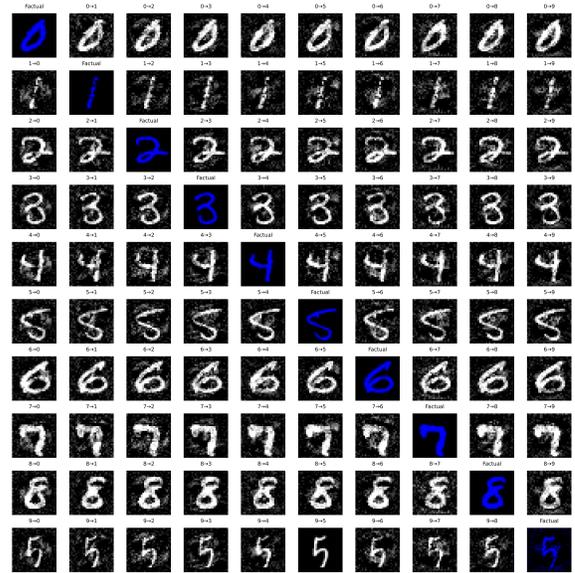}}

}

\caption{\label{fig-mnist-vanilla}The same setup, factuals, model
architecture and generator as in Figure~\ref{fig-mnist}, but the model
was trained conventionally.}

\end{figure}%

\end{minipage}%

\end{figure}

Figure~\ref{fig-mnist} shows much more plausible (faithful)
counterfactuals for a model with CT than the model with conventional
training (Figure~\ref{fig-mnist-vanilla}).

\subsubsection{Integrated Gradients}\label{integrated-gradients}

We make use of integrated gradients (IG) proposed by Sundararajan, Taly,
and Yan (\citeproc{ref-sundararajan2017ig}{2017}) to empirically
evaluate the feature protection mechanism in CT. We choose this approach
because it produces theoretically sound results, works well for
non-linear models, and remains relatively inexpensive.

IG calculates the contribution of each input feature towards a specific
prediction by approximating the integral of the model output with
respect to its input, using a set of samples that linearly interpolate
between a test instance and some baseline instance
(\citeproc{ref-sundararajan2017ig}{Sundararajan, Taly, and Yan 2017}).
This process produces a vector of real numbers, one per input feature,
which informs about the contribution of each feature to the prediction.
For example:

\begin{itemize}
\tightlist
\item
  a large positive value indicates that a feature has strong positive
  influence on the classification (i.e., increases the score for a
  class);
\item
  a small negative value indicates that a feature has weak negative
  influence on the classification (i.e., decreases the score for a
  class).
\end{itemize}

To calculate the contributions, IG compares the output to a baseline.
The selection of an appropriate baseline is an important design decision
--- it should produce a ``neutral'' prediction to avoid capturing
effects that cannot be directly attributed to the model
(\citeproc{ref-sundararajan2017ig}{Sundararajan, Taly, and Yan 2017};
\citeproc{ref-sturmfels2020visualizing}{Sturmfels, Lundberg, and Lee
2020}). To remain consistent in our evaluations, we use a baseline drawn
at random from a uniform distribution, \(\mathcal{U}(-1,1)\), for all
datasets. This aligns with standard evaluation practices for IG.

We run IG on models trained on all datasets to compare their sensitivity
to features that were protected using CT:

\begin{itemize}
\tightlist
\item
  for synthetic datasets, this is always the first feature
\item
  for real-world tabular datasets, this is always \emph{age}
\item
  for MNIST, this is first five and last five rows of pixels
\end{itemize}

As IG outputs are not bounded (i.e., they are arbitrary real numbers),
it becomes a challenge to meaningfully compare IG outputs of different
models --- ones that are trained conventionally, and ones that underwent
counterfactual training. For our purposes, we observe with reference to
our Proposition, that we are interested estimating changes in the
relative contribution of protected features compared to mutable ones.
Thus, to meaningfully compare integrated gradients for different models
and to accommodate for variable ranges of outputs in absolute terms, we
standardize the integrated gradients across features.

\begin{figure}

\centering{

\pandocbounded{\includegraphics[keepaspectratio]{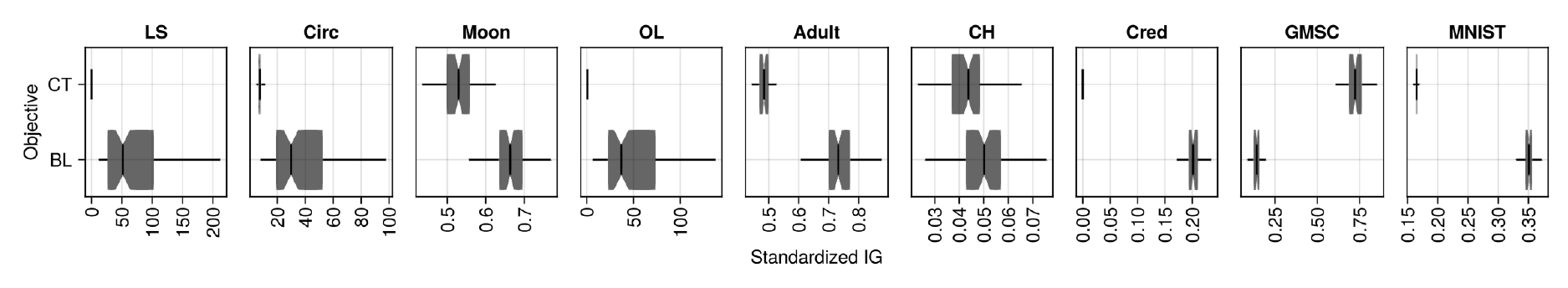}}

}

\caption{\label{fig-ig-box}Interquartile ranges of bootstrap outcomes
for sensitivity to protected features measured by standardized
integrated gradients.}

\end{figure}

Let \(\mathbf{g}_d\) denote the estimated IG for feature \(d\). Then in
the case of 2D synthetic datasets we find that taking the absolute value
of the outputs, \(|\mathbf{g}_d|\), and then dividing them by a
\(\max(\mathbf{g}) - \min(\mathbf{g})\) term allows us to make the most
meaningful comparison. In the case of real-world datasets we choose to
normalize the values to a \([0,1]\) range instead. We compare the
(average) sensitivity to the features that were protected for CT models.
Once again we use bootstrapping (100 rounds, 2500 samples per round) to
establish the significance of our results (Figure~\ref{fig-ig-box}).

\subsubsection{Costs and Validity}\label{costs-and-validity}

In Table~\ref{tbl-panel}, we present additional outcomes for common
evaluation metrics: Table~\ref{tbl-costs} presents the average reduction
in costs of counterfactuals for CT vs.~BL with no mutability
constraints, i.e.~corresponding to the first two columns in the main
table of the paper; Table~\ref{tbl-val} shows the corresponding average
validities; finally, Table~\ref{tbl-val-mtbl} shows average validities
for the case with mutability constraints, i.e.~corresponding to the
third columns in the main table of the paper.

As noted in the discussion section of the main paper, we observe mixed
results results here. Average costs in terms of distances from factual
values decrease for most datasets, which is positively surprising since
improved plausibility requires counterfactuals to travel further into
the target domain than minimum distance counterfactuals. It appears that
in these cases faithful counterfactuals for the baseline model still end
up far away from their initial starting points, but not close enough for
samples in the target domain to be plausible. In that sense, CT can be
seen to improve both plausibility and costs for faithful CE. In some
cases though (\emph{LS}, \emph{CH}, \emph{MNIST}), we do seem to observe
the tradeoff between plausibility and costs play out, as we would expect
(compare panels (a) and (b) of Figure 1 in the main paper for
reference).

Concerning validity, we find that can lead to substantial reductions and
only increases average validity compared to the baseline in one case
(\emph{Circ}). As noted in the discussion section of the main paper,
this result does not surprise us: by design, CT shrinks the solution
space for valid counterfactual explanations, thus making it ``harder''
to reach validity compared to the baseline model. Note that for a number
of reasons this should not be seen as problematic:

\begin{enumerate}
\def\labelenumi{\arabic{enumi}.}
\tightlist
\item
  Validity of gradient-based CE is a function on the number of steps and
  the step size which we both kept fixed during evaluation: simply
  adjusting \(T=50\) to higher values or choosing a larger step size
  will lead to higher rates of validity.
\item
  Even though reaching validity is sometimes ``harder'' in terms of the
  necessary number of steps for a given step size, we have already shown
  that the average distances that counterfactuals need to travel
  decrease for most datasets. Users care about costs in terms of feature
  distances, not search iteration steps.
\item
  From a philosophical perspective on algorithmic recourse, validity in
  and off itself is not a sufficient desideratum for counterfactuals. In
  fact, Venkatasubramanian and Alfano
  (\citeproc{ref-venkatasubramanian2020philosophical}{2020}) propose
  introducing an upper bound on costs of the flipset (i.e.~the set of
  valid CE), arguing that valid but highly costly counterfactuals are
  not useful to individuals in practice. In a similar fashion, it could
  be argued that there should be an upper bound on the implausibility of
  counterfactuals in the flipset.
\end{enumerate}

\begin{table}

\caption{\label{tbl-panel}Costs and validity.}

\begin{minipage}{0.40\linewidth}

\subcaption{\label{tbl-costs}Reduction in average costs for CT vs.~the
baseline. Results correspond to the case with no mutability constraints
in the main table of the paper.}

\centering{

\begin{tabular}{
l
S[table-format=2.2(1.2)]
}
  \toprule
  \textbf{Data} & \textbf{Cost $(-\%)$} \\\midrule
  LS & -27.11\pm0.75 $^{*}$ \\
  Circ & 40.17\pm0.85 $^{*}$ \\
  Moon & 32.54\pm1.23 $^{*}$ \\
  OL & 12.08\pm1.58 $^{*}$ \\\midrule
  Adult & -4.59\pm2.54 $^{}$ \\
  CH & -33.04\pm1.96 $^{*}$ \\
  Cred & 27.43\pm1.05 $^{*}$ \\
  GMSC & -22.4\pm3.64 $^{*}$ \\
  MNIST & -40.71\pm7.02 $^{*}$ \\\midrule
  Avg. & -1.74 \\\bottomrule
\end{tabular}

}

\end{minipage}%
\begin{minipage}{0.20\linewidth}
~\end{minipage}%
\begin{minipage}{0.40\linewidth}

\subcaption{\label{tbl-val}Average validities of counterfactuals for CT
and BL. Unconstrained case.}

\centering{

\begin{tabular}{lcc}
  \toprule
  \textbf{Data} & \textbf{CT} & \textbf{BL} \\\midrule
  LS & 1.0 & 1.0 \\
  Circ & 1.0 & 0.51 \\
  Moon & 1.0 & 1.0 \\
  OL & 0.86 & 0.98 \\\midrule
  Adult & 0.68 & 0.99 \\
  CH & 1.0 & 1.0 \\
  Cred & 0.72 & 1.0 \\
  GMSC & 0.94 & 1.0 \\
  MNIST & 1.0 & 1.0 \\\bottomrule
\end{tabular}

}

\end{minipage}%
\newline
\begin{minipage}{\linewidth}

\subcaption{\label{tbl-val-mtbl}Average validities of counterfactuals
for CT and BL. Mutability constrained case.}

\centering{

\begin{tabular}{lcc}
  \toprule
  \textbf{Data} & \textbf{CT} & \textbf{BL} \\\midrule
  LS & 1.0 & 1.0 \\
  Circ & 0.71 & 0.48 \\
  Moon & 1.0 & 0.98 \\
  OL & 0.34 & 0.56 \\\midrule
  Adult & 0.7 & 0.99 \\
  CH & 1.0 & 1.0 \\
  Cred & 0.74 & 1.0 \\
  GMSC & 0.97 & 1.0 \\
  MNIST & 1.0 & 1.0 \\\bottomrule
\end{tabular}

}

\end{minipage}%

\end{table}%

\FloatBarrier

\section{Grid Searches}\label{sec-app-grid}

To assess the hyperparameter sensitivity of our proposed training regime
we ran multiple large grid searches for all of our synthetic datasets.
We have grouped these grid searches into multiple categories:

\begin{enumerate}
\def\labelenumi{\arabic{enumi}.}
\tightlist
\item
  \textbf{Generator Parameters} (Section~\ref{sec-app-grid-gen}):
  Investigates the effect of changing hyperparameters that affect the
  counterfactual outcomes during the training phase.
\item
  \textbf{Penalty Strengths} (Section~\ref{sec-app-grid-pen}):
  Investigates the effect of changing the penalty strengths in our
  proposed training objective.
\item
  \textbf{Other Parameters} (Section~\ref{sec-app-grid-train}):
  Investigates the effect of changing other training parameters,
  including the total number of generated counterfactuals in each epoch.
\end{enumerate}

We begin by summarizing the high-level findings in
Section~\ref{sec-app-grid-hl}. For each of the categories,
Section~\ref{sec-app-grid-gen} to Section~\ref{sec-app-grid-train} then
present all details including the exact parameter grids, average
predictive performance outcomes and key evaluation metrics for the
generated counterfactuals.

\subsection{Evaluation Details}\label{evaluation-details}

To measure predictive performance, we compute the accuracy and F1-score
for all models on test data (Table~\ref{tbl-acc-gen},
Table~\ref{tbl-acc-pen}, Table~\ref{tbl-acc-train}). With respect to
explanatory performance, we report here our findings for the
(im)plausibility and cost of counterfactuals at test time. Since the
computation of our proposed divergence-based adaption (\(\text{IP}^*\))
is memory-intensive, we rely on the distance-based metric for the grid
searches. For the counterfactual evaluation, we draw factual samples
from the training data for the grid searches to avoid data leakage with
respect to our final results reported in the body of the paper.
Specifically, we want to avoid choosing our default hyperparameters
based on results on the test data. Since we are optimizing for
explainability, not predictive performance, we still present test
accuracy and F1-scores.

\subsubsection{Predictive Performance}\label{predictive-performance-1}

We find that CT is associated with little to no decrease in average
predictive performance for our synthetic datasets: test accuracy and
F1-scores decrease by at most \textasciitilde1 percentage point, but
generally much less (Table~\ref{tbl-acc-gen}, Table~\ref{tbl-acc-pen},
Table~\ref{tbl-acc-train}). Variation across hyperparameters is
negligible as indicated by small standard deviations for these metrics
across the board.

\subsubsection{Counterfactual Outcomes}\label{sec-app-grid-hl}

Overall, we find that counterfactual training achieves it key objectives
consistently across all hyperparameter settings and also broadly across
datasets: plausibility is improved by up to 60 percent (\%) for the
\emph{Circles} data (e.g.
Figure~\ref{fig-grid-gen_params-plaus-circles}), 25-30\% for the
\emph{Moons} data (e.g. Figure~\ref{fig-grid-gen_params-plaus-moons})
and 10-20\% for the \emph{Linearly Separable} data (e.g.
Figure~\ref{fig-grid-gen_params-plaus-lin_sep}). At the same time, the
average costs of faithful counterfactuals are reduced in many cases by
around 20-25\% for \emph{Circles} (e.g.
Figure~\ref{fig-grid-gen_params-cost-circles}) and up to 50\% for
\emph{Moons} (e.g. Figure~\ref{fig-grid-gen_params-cost-moons}). For the
\emph{Linearly Separable} data, costs are generally increased although
typically by less than 10\% (e.g.
Figure~\ref{fig-grid-gen_params-cost-lin_sep}), which reflects a common
tradeoff between costs and plausibility
(\citeproc{ref-altmeyer2024faithful}{Altmeyer et al. 2024}).

We do observe strong sensitivity to certain hyperparameters, with clear
an manageable patterns. Concerning generator parameters, we firstly find
that using \emph{REVISE} to generate counterfactuals during training
typically yields the worst outcomes out of all generators, often leading
to a substantial decrease in plausibility. This finding can be
attributed to the fact that \emph{REVISE} effectively assigns the task
of learning plausible explanations from the model itself to a surrogate
VAE. In other words, counterfactuals generated by \emph{REVISE} are less
faithful to the model that \emph{ECCCo} and \emph{Generic}, and hence we
would expect them to be a less effective and, in fact, potentially
detrimental role in our training regime. Secondly, we observe that
allowing for a higher number of maximum steps \(T\) for the
counterfactual search generally yields better outcomes. This is
intuitive, because it allows more counterfactuals to reach maturity in
any given iteration. Looking in particular at the results for
\emph{Linearly Separable}, it seems that higher values for \(T\) in
combination with higher decision thresholds (\(\tau\)) yields the best
results when using \emph{ECCCo}. But depending on the degree of class
separability of the underlying data, a high decision-threshold can also
affect results adversely, as evident from the results for the
\emph{Overlapping} data (Figure~\ref{fig-grid-gen_params-plaus-over}):
here we find that CT generally fails to achieve its objective because
only a tiny proportion of counterfactuals ever reaches maturity.

Regarding penalty strengths, we find that the strength of the energy
regularization, \(\lambda_{\text{reg}}\) is a key hyperparameter, while
sensitivity with respect to \(\lambda_{\text{div}}\) and
\(\lambda_{\text{adv}}\) is much less evident. In particular, we observe
that not regularizing energy enough or at all typically leads to poor
performance in terms of decreased plausibility and increased costs, in
particular for \emph{Circles} (Figure~\ref{fig-grid-pen-plaus-circles}),
\emph{Linearly Separable} (Figure~\ref{fig-grid-pen-plaus-lin_sep}) and
\emph{Overlapping} (Figure~\ref{fig-grid-pen-plaus-over}). High values
of \(\lambda_{\text{reg}}\) can increase the variability in outcomes, in
particular when combined with high values for \(\lambda_{\text{div}}\)
and \(\lambda_{\text{adv}}\), but this effect is less pronounced.

Finally, concerning other hyperparameters we observe that the
effectiveness and stability of CT is positively associated with the
number of counterfactuals generated during each training epoch, in
particular for \emph{Circles}
(Figure~\ref{fig-grid-train-plaus-circles}) and \emph{Moons}
(Figure~\ref{fig-grid-train-plaus-moons}). We further find that a higher
number of training epochs is beneficial as expected, where we tested
training models for 50 and 100 epochs. Interestingly, we find that it is
not necessary to employ CT during the entire training phase to achieve
the desired improvements in explainability: specifically, we have tested
training models conventionally during the first half of training before
switching to CT after this initial burn-in period.

\subsection{Generator Parameters}\label{sec-app-grid-gen}

The hyperparameter grid with varying generator parameters during
training is shown in Note~\ref{nte-gen-params-final-run-train}. The
corresponding evaluation grid used for these experiments is shown in
Note~\ref{nte-gen-params-final-run-eval}.

\begin{tcolorbox}[enhanced jigsaw, arc=.35mm, colback=white, bottomrule=.15mm, toptitle=1mm, leftrule=.75mm, titlerule=0mm, toprule=.15mm, breakable, colframe=quarto-callout-note-color-frame, left=2mm, opacityback=0, bottomtitle=1mm, opacitybacktitle=0.6, title={Note \ref*{nte-gen-params-final-run-train}space Training Phase}, colbacktitle=quarto-callout-note-color!10!white, rightrule=.15mm, coltitle=black]

\quartocalloutnte{nte-gen-params-final-run-train} 

\begin{itemize}
\tightlist
\item
  Generator Parameters:

  \begin{itemize}
  \tightlist
  \item
    Decision Threshold: \texttt{0.75,\ 0.9,\ 0.95}
  \item
    \(\lambda_{\text{egy}}\): \texttt{0.1,\ 0.5,\ 5.0,\ 10.0,\ 20.0}
  \item
    Maximum Iterations: \texttt{5,\ 25,\ 50}
  \end{itemize}
\item
  Generator: \texttt{ecco,\ generic,\ revise}
\item
  Model: \texttt{mlp}
\item
  Training Parameters:

  \begin{itemize}
  \tightlist
  \item
    Objective: \texttt{full,\ vanilla}
  \end{itemize}
\end{itemize}

\end{tcolorbox}

\begin{tcolorbox}[enhanced jigsaw, arc=.35mm, colback=white, bottomrule=.15mm, toptitle=1mm, leftrule=.75mm, titlerule=0mm, toprule=.15mm, breakable, colframe=quarto-callout-note-color-frame, left=2mm, opacityback=0, bottomtitle=1mm, opacitybacktitle=0.6, title={Note \ref*{nte-gen-params-final-run-eval}space Evaluation Phase}, colbacktitle=quarto-callout-note-color!10!white, rightrule=.15mm, coltitle=black]

\quartocalloutnte{nte-gen-params-final-run-eval} 

\begin{itemize}
\tightlist
\item
  Generator Parameters:

  \begin{itemize}
  \tightlist
  \item
    \(\lambda_{\text{egy}}\): \texttt{0.1,\ 0.5,\ 1.0,\ 5.0,\ 10.0}
  \end{itemize}
\end{itemize}

\end{tcolorbox}

\subsubsection{Predictive Performance}\label{predictive-performance-2}

Predictive performance measures for this grid search are shown in
Table~\ref{tbl-acc-gen}.

\begin{longtable}{ccccc}

\caption{\label{tbl-acc-gen}Predictive performance measures by dataset
and objective averaged across training-phase parameters
(Note~\ref{nte-gen-params-final-run-train}) and evaluation-phase
parameters (Note~\ref{nte-gen-params-final-run-eval}).}

\tabularnewline

  \toprule
  \textbf{Dataset} & \textbf{Variable} & \textbf{Objective} & \textbf{Mean} & \textbf{Se} \\\midrule
  \endfirsthead
  \toprule
  \textbf{Dataset} & \textbf{Variable} & \textbf{Objective} & \textbf{Mean} & \textbf{Se} \\\midrule
  \endhead
  \bottomrule
  \multicolumn{5}{r}{Continuing table below.}\\
  \bottomrule
  \endfoot
  \endlastfoot
  Circ & Accuracy & Full & 1.0 & 0.0 \\
  Circ & Accuracy & Vanilla & 1.0 & 0.0 \\
  Circ & F1-score & Full & 1.0 & 0.0 \\
  Circ & F1-score & Vanilla & 1.0 & 0.0 \\
  LS & Accuracy & Full & 1.0 & 0.0 \\
  LS & Accuracy & Vanilla & 1.0 & 0.0 \\
  LS & F1-score & Full & 1.0 & 0.0 \\
  LS & F1-score & Vanilla & 1.0 & 0.0 \\
  Moon & Accuracy & Full & 1.0 & 0.0 \\
  Moon & Accuracy & Vanilla & 1.0 & 0.0 \\
  Moon & F1-score & Full & 1.0 & 0.0 \\
  Moon & F1-score & Vanilla & 1.0 & 0.0 \\
  OL & Accuracy & Full & 0.91 & 0.0 \\
  OL & Accuracy & Vanilla & 0.92 & 0.0 \\
  OL & F1-score & Full & 0.91 & 0.0 \\
  OL & F1-score & Vanilla & 0.92 & 0.0 \\\bottomrule

\end{longtable}

\subsubsection{Plausibility}\label{plausibility}

\begin{figure}

\centering{

\includegraphics[width=0.8\linewidth,height=\textheight,keepaspectratio]{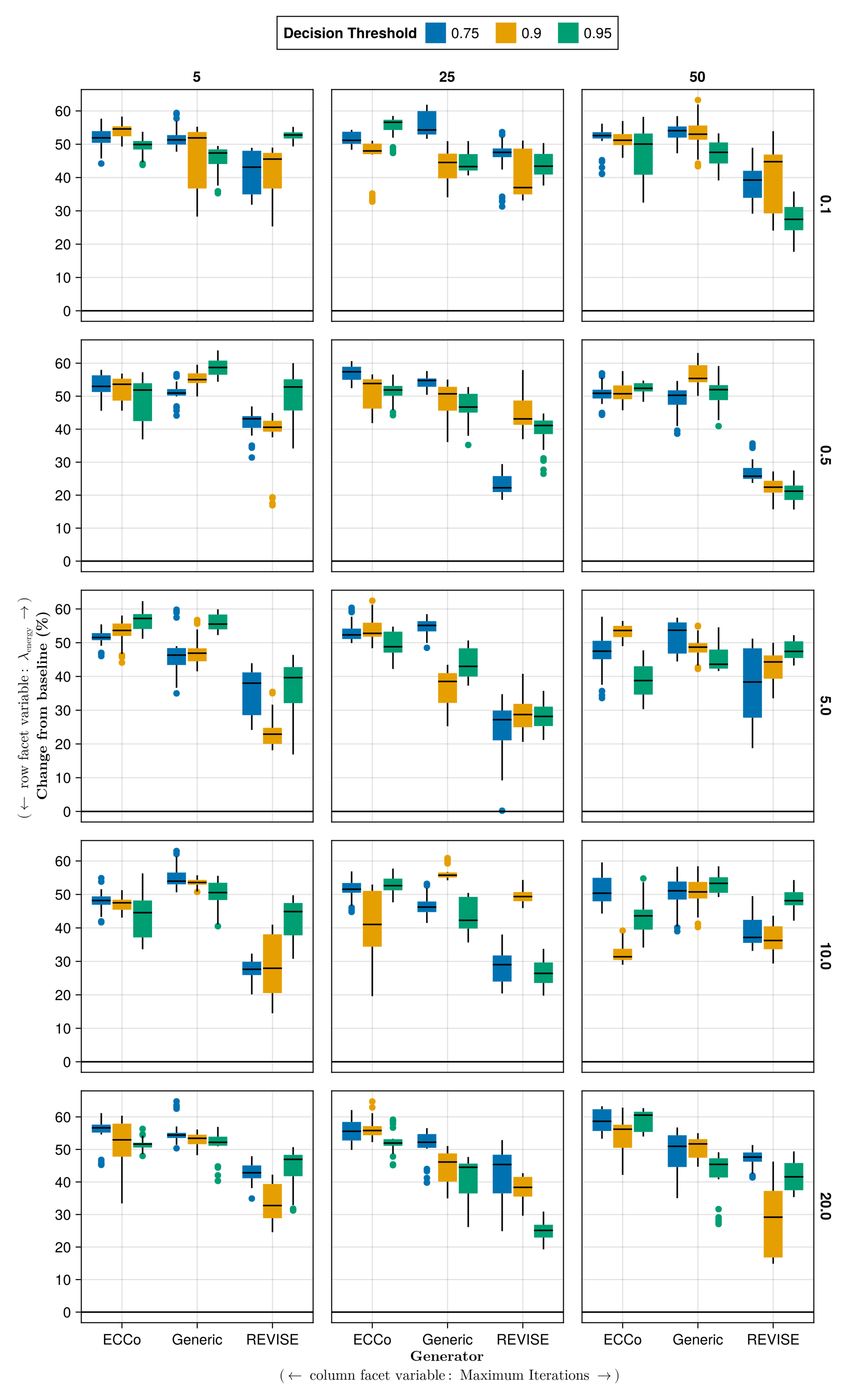}

}

\caption{\label{fig-grid-gen_params-plaus-circles}Average outcomes for
the plausibility measure across hyperparameters. This shows the \%
change from the baseline model for the distance-based implausibility
metric (\(\text{IP}\)). Boxplots indicate the variation across
evaluation runs and test settings (varying parameters for \emph{ECCCo}).
Data: Circles.}

\end{figure}

\begin{figure}

\centering{

\includegraphics[width=0.8\linewidth,height=\textheight,keepaspectratio]{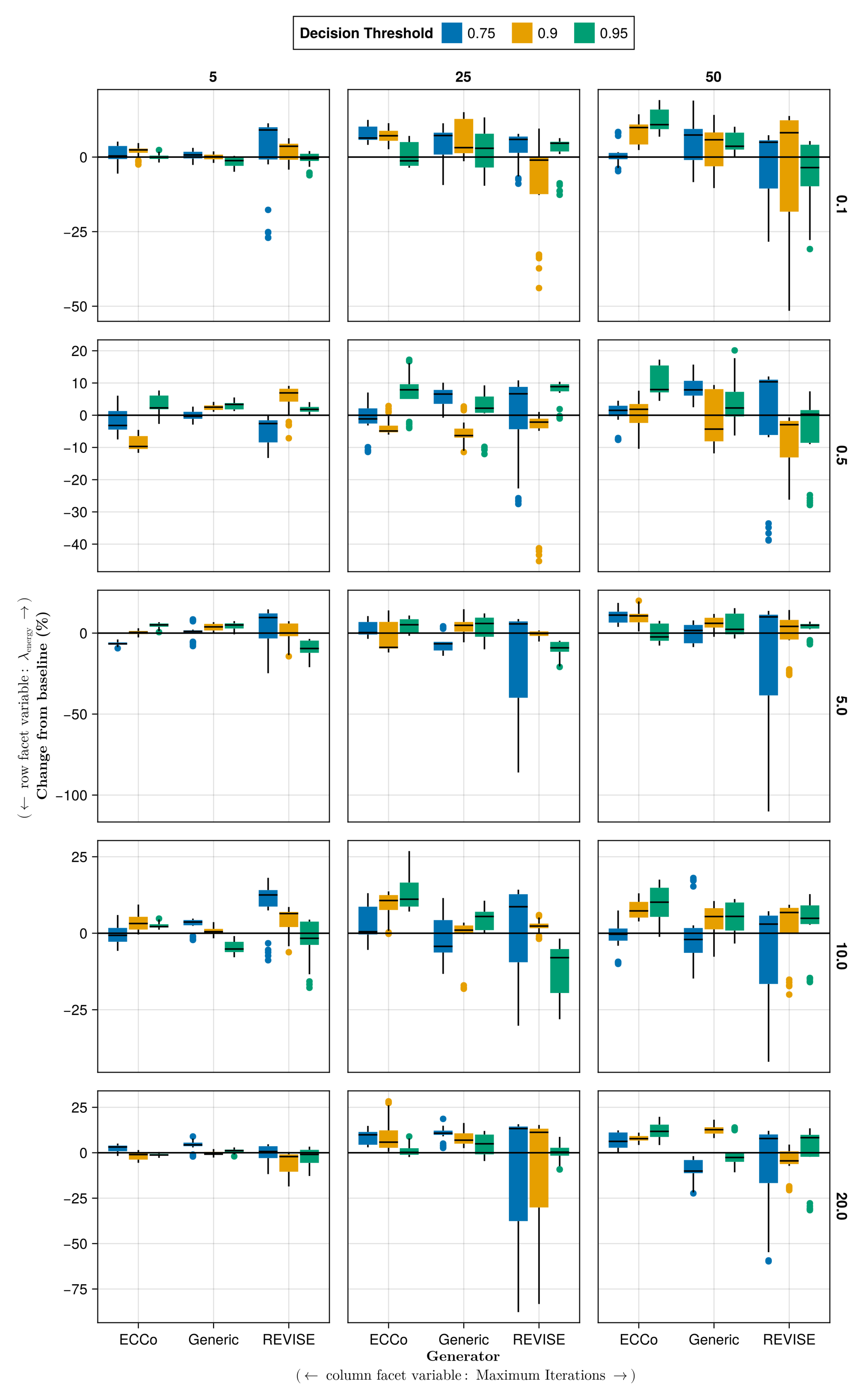}

}

\caption{\label{fig-grid-gen_params-plaus-lin_sep}Average outcomes for
the plausibility measure across hyperparameters. This shows the \%
change from the baseline model for the distance-based implausibility
metric (\(\text{IP}\)). Boxplots indicate the variation across
evaluation runs and test settings (varying parameters for \emph{ECCCo}).
Data: Linearly Separable.}

\end{figure}

\begin{figure}

\centering{

\includegraphics[width=0.8\linewidth,height=\textheight,keepaspectratio]{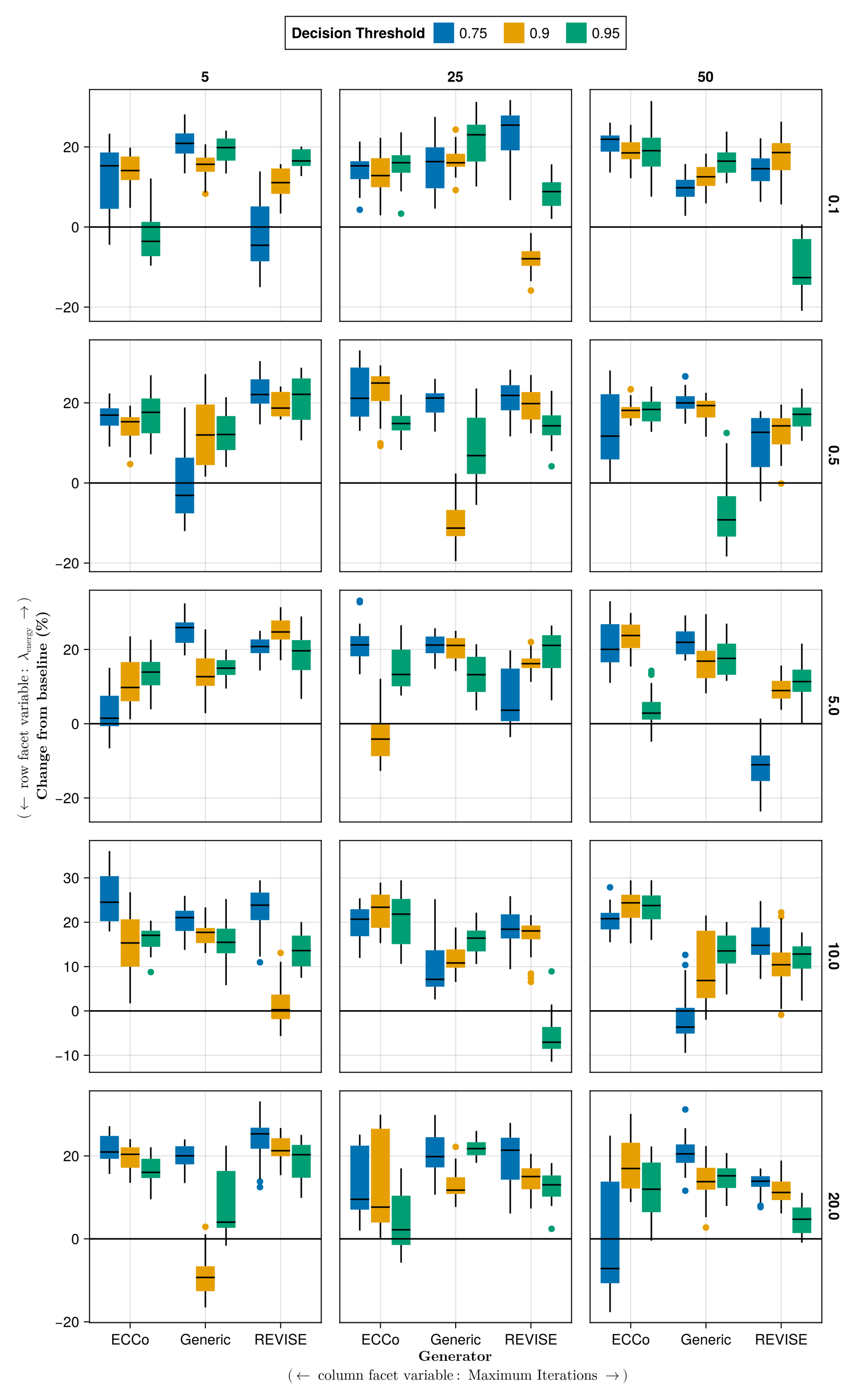}

}

\caption{\label{fig-grid-gen_params-plaus-moons}Average outcomes for the
plausibility measure across hyperparameters. This shows the \% change
from the baseline model for the distance-based implausibility metric
(\(\text{IP}\)). Boxplots indicate the variation across evaluation runs
and test settings (varying parameters for \emph{ECCCo}). Data: Moons.}

\end{figure}

\begin{figure}

\centering{

\includegraphics[width=0.8\linewidth,height=\textheight,keepaspectratio]{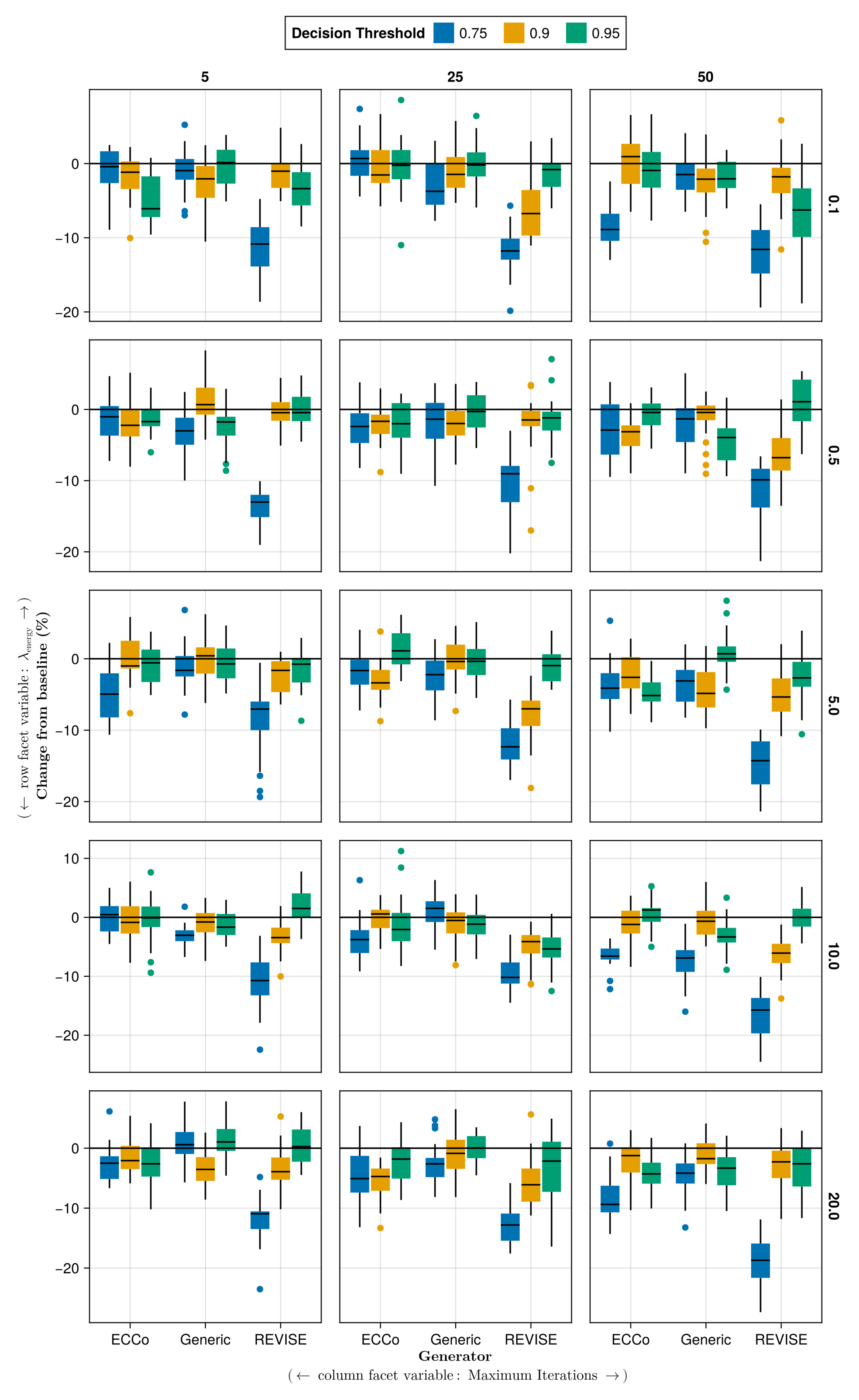}

}

\caption{\label{fig-grid-gen_params-plaus-over}Average outcomes for the
plausibility measure across hyperparameters. This shows the \% change
from the baseline model for the distance-based implausibility metric
(\(\text{IP}\)). Boxplots indicate the variation across evaluation runs
and test settings (varying parameters for \emph{ECCCo}). Data:
Overlapping.}

\end{figure}

The results with respect to the plausibility measure are shown in
Figure~\ref{fig-grid-gen_params-plaus-circles} to
Figure~\ref{fig-grid-gen_params-plaus-over}.

\subsubsection{Cost}\label{cost}

\begin{figure}

\centering{

\includegraphics[width=0.8\linewidth,height=\textheight,keepaspectratio]{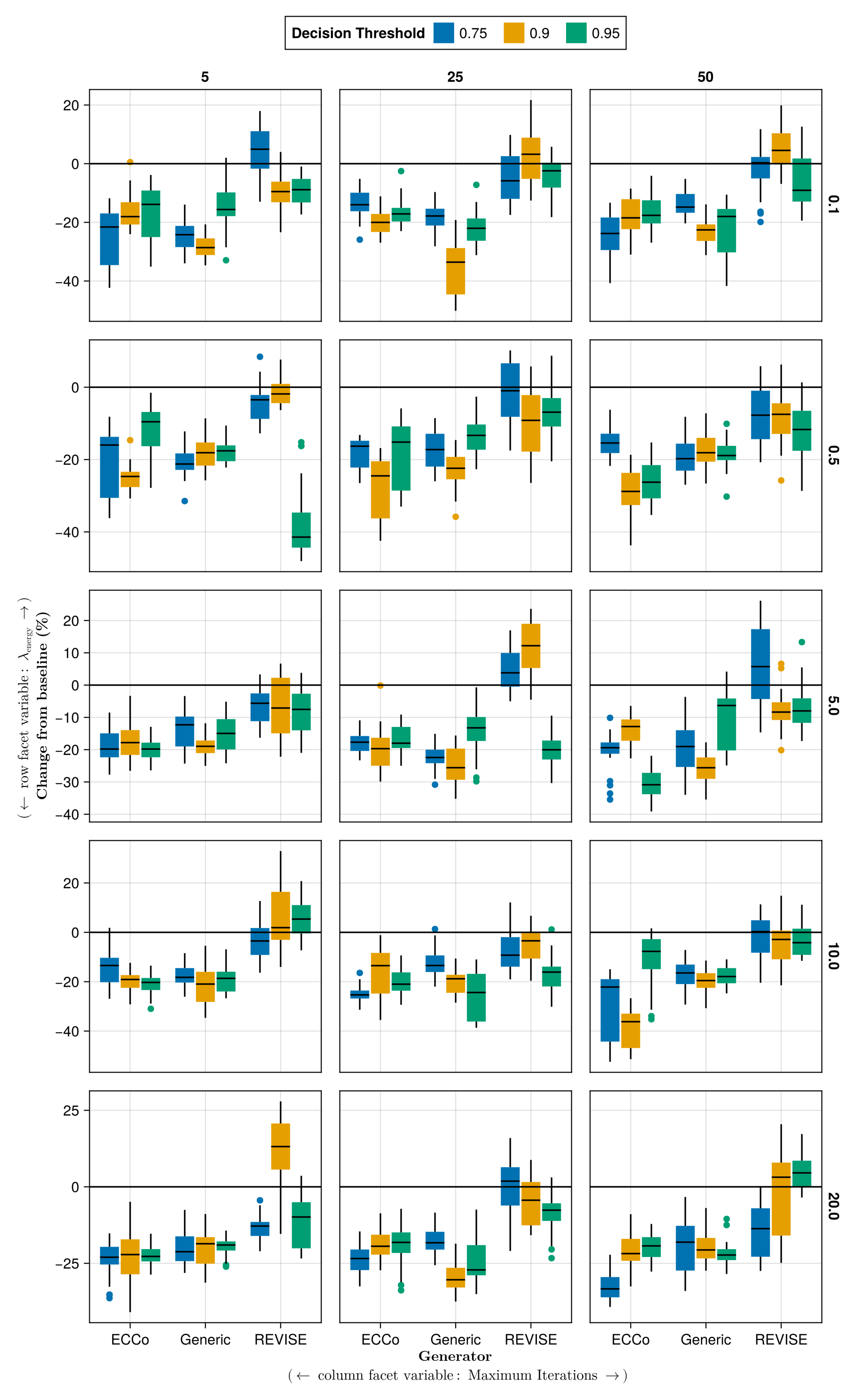}

}

\caption{\label{fig-grid-gen_params-cost-circles}Average outcomes for
the cost measure across hyperparameters. This shows the \% change from
the baseline model for the distance-based cost metric
(\citeproc{ref-wachter2017counterfactual}{Wachter, Mittelstadt, and
Russell 2017}). Boxplots indicate the variation across evaluation runs
and test settings (varying parameters for \emph{ECCCo}). Data: Circles.}

\end{figure}

\begin{figure}

\centering{

\includegraphics[width=0.8\linewidth,height=\textheight,keepaspectratio]{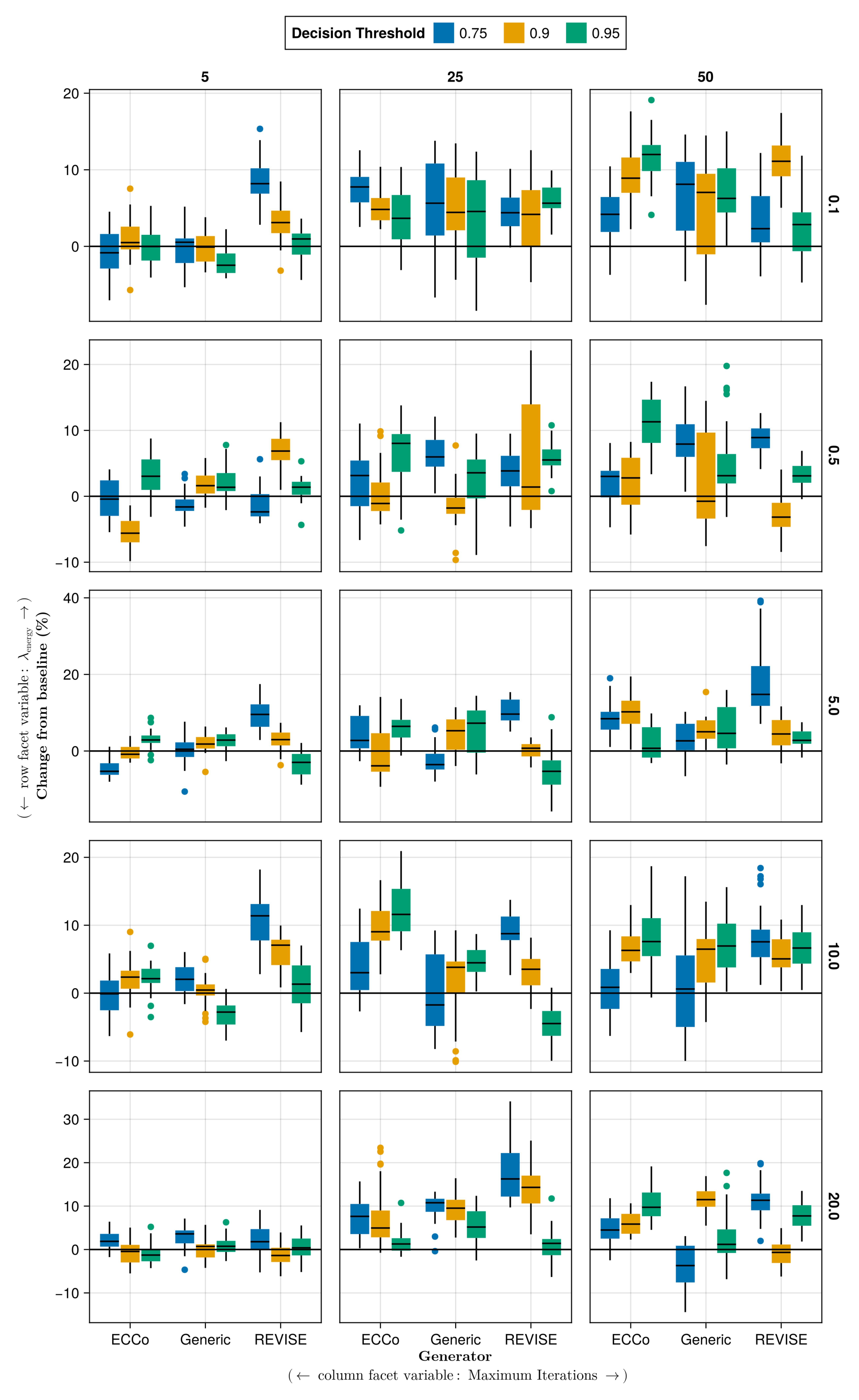}

}

\caption{\label{fig-grid-gen_params-cost-lin_sep}Average outcomes for
the cost measure across hyperparameters. This shows the \% change from
the baseline model for the distance-based cost metric
(\citeproc{ref-wachter2017counterfactual}{Wachter, Mittelstadt, and
Russell 2017}). Boxplots indicate the variation across evaluation runs
and test settings (varying parameters for \emph{ECCCo}). Data: Linearly
Separable.}

\end{figure}

\begin{figure}

\centering{

\includegraphics[width=0.8\linewidth,height=\textheight,keepaspectratio]{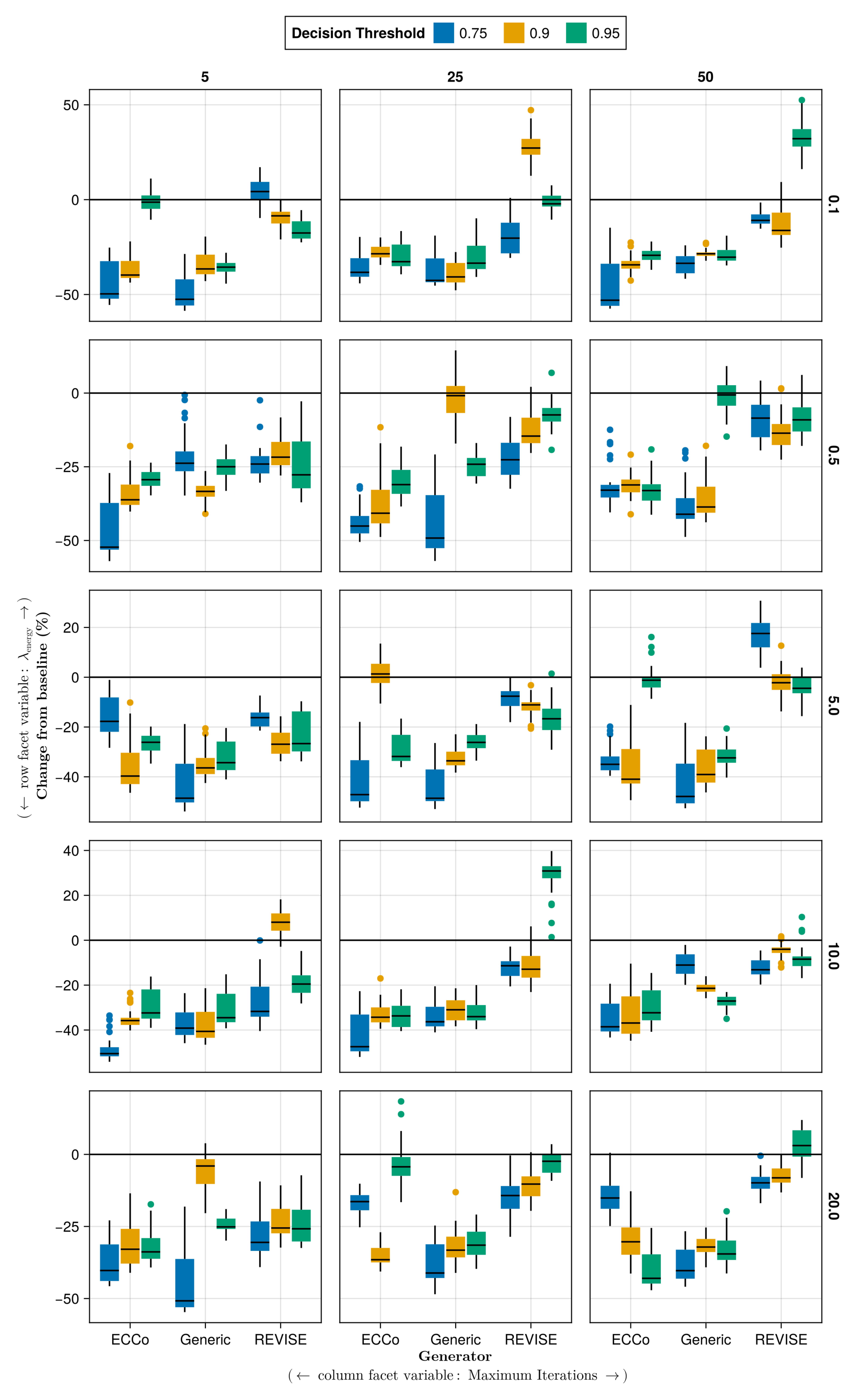}

}

\caption{\label{fig-grid-gen_params-cost-moons}Average outcomes for the
cost measure across hyperparameters. This shows the \% change from the
baseline model for the distance-based cost metric
(\citeproc{ref-wachter2017counterfactual}{Wachter, Mittelstadt, and
Russell 2017}). Boxplots indicate the variation across evaluation runs
and test settings (varying parameters for \emph{ECCCo}). Data: Moons.}

\end{figure}

\begin{figure}

\centering{

\includegraphics[width=0.8\linewidth,height=\textheight,keepaspectratio]{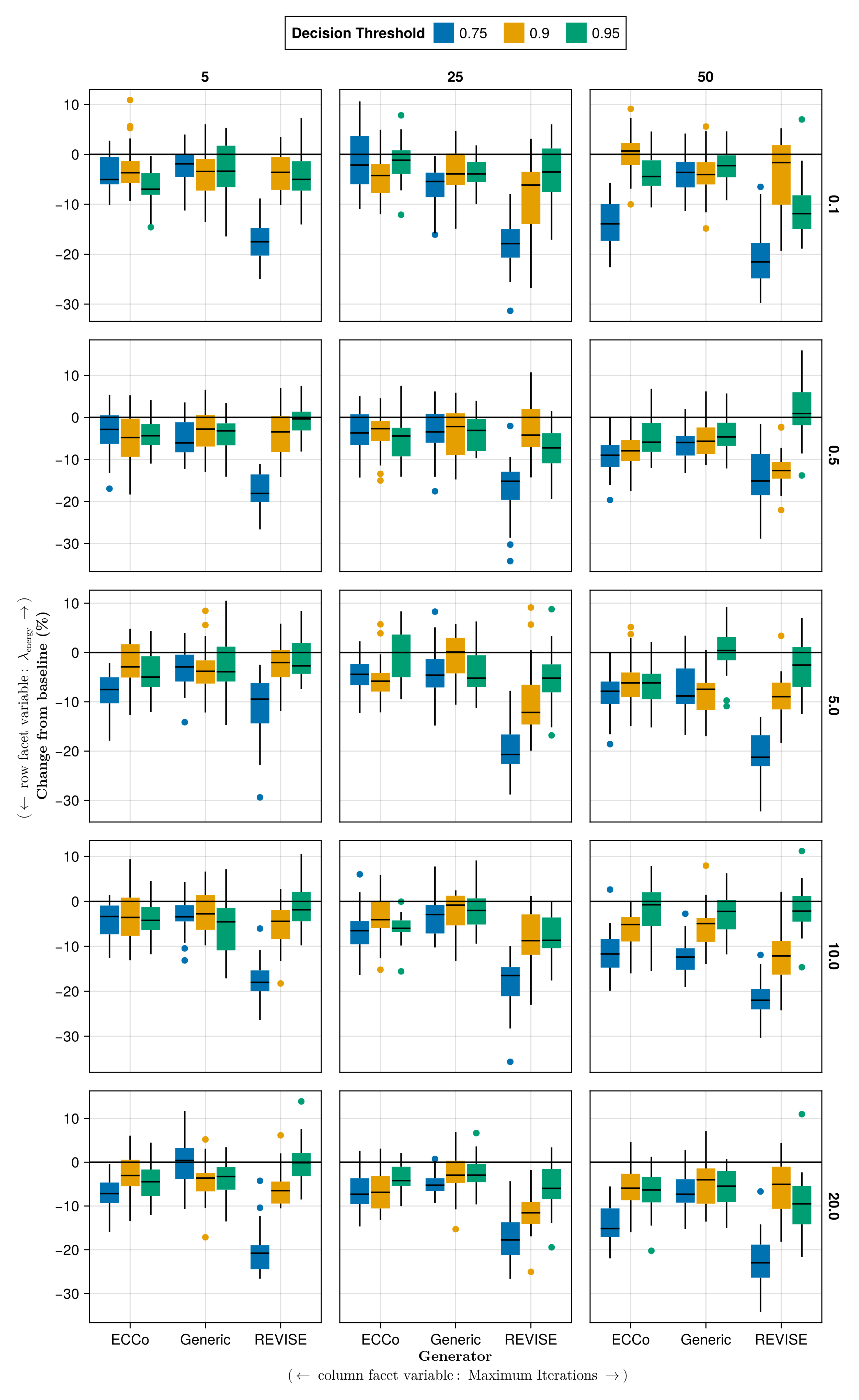}

}

\caption{\label{fig-grid-gen_params-cost-over}Average outcomes for the
cost measure across hyperparameters. This shows the \% change from the
baseline model for the distance-based cost metric
(\citeproc{ref-wachter2017counterfactual}{Wachter, Mittelstadt, and
Russell 2017}). Boxplots indicate the variation across evaluation runs
and test settings (varying parameters for \emph{ECCCo}). Data:
Overlapping.}

\end{figure}

The results with respect to the cost measure are shown in
Figure~\ref{fig-grid-gen_params-cost-circles} to
Figure~\ref{fig-grid-gen_params-cost-over}.

\subsection{Penalty Strengths}\label{sec-app-grid-pen}

The hyperparameter grid with varying penalty strengths during training
is shown in Note~\ref{nte-pen-final-run-train}. The corresponding
evaluation grid used for these experiments is shown in
Note~\ref{nte-pen-final-run-eval}.

\begin{tcolorbox}[enhanced jigsaw, arc=.35mm, colback=white, bottomrule=.15mm, toptitle=1mm, leftrule=.75mm, titlerule=0mm, toprule=.15mm, breakable, colframe=quarto-callout-note-color-frame, left=2mm, opacityback=0, bottomtitle=1mm, opacitybacktitle=0.6, title={Note \ref*{nte-pen-final-run-train}space Training Phase}, colbacktitle=quarto-callout-note-color!10!white, rightrule=.15mm, coltitle=black]

\quartocalloutnte{nte-pen-final-run-train} 

\begin{itemize}
\tightlist
\item
  Generator: \texttt{ecco,\ generic,\ revise}
\item
  Model: \texttt{mlp}
\item
  Training Parameters:

  \begin{itemize}
  \tightlist
  \item
    \(\lambda_{\text{adv}}\): \texttt{0.1,\ 0.25,\ 1.0}
  \item
    \(\lambda_{\text{div}}\): \texttt{0.01,\ 0.1,\ 1.0}
  \item
    \(\lambda_{\text{reg}}\): \texttt{0.0,\ 0.01,\ 0.1,\ 0.25,\ 0.5}
  \item
    Objective: \texttt{full,\ vanilla}
  \end{itemize}
\end{itemize}

\end{tcolorbox}

\begin{tcolorbox}[enhanced jigsaw, arc=.35mm, colback=white, bottomrule=.15mm, toptitle=1mm, leftrule=.75mm, titlerule=0mm, toprule=.15mm, breakable, colframe=quarto-callout-note-color-frame, left=2mm, opacityback=0, bottomtitle=1mm, opacitybacktitle=0.6, title={Note \ref*{nte-pen-final-run-eval}space Evaluation Phase}, colbacktitle=quarto-callout-note-color!10!white, rightrule=.15mm, coltitle=black]

\quartocalloutnte{nte-pen-final-run-eval} 

\begin{itemize}
\tightlist
\item
  Generator Parameters:

  \begin{itemize}
  \tightlist
  \item
    \(\lambda_{\text{egy}}\): \texttt{0.1,\ 0.5,\ 1.0,\ 5.0,\ 10.0}
  \end{itemize}
\end{itemize}

\end{tcolorbox}

\subsubsection{Predictive Performance}\label{predictive-performance-3}

Predictive performance measures for this grid search are shown in
Table~\ref{tbl-acc-pen}.

\begin{longtable}{ccccc}

\caption{\label{tbl-acc-pen}Predictive performance measures by dataset
and objective averaged across training-phase parameters
(Note~\ref{nte-pen-final-run-train}) and evaluation-phase parameters
(Note~\ref{nte-pen-final-run-eval}).}

\tabularnewline

  \toprule
  \textbf{Dataset} & \textbf{Variable} & \textbf{Objective} & \textbf{Mean} & \textbf{Se} \\\midrule
  \endfirsthead
  \toprule
  \textbf{Dataset} & \textbf{Variable} & \textbf{Objective} & \textbf{Mean} & \textbf{Se} \\\midrule
  \endhead
  \bottomrule
  \multicolumn{5}{r}{Continuing table below.}\\
  \bottomrule
  \endfoot
  \endlastfoot
  Circ & Accuracy & Full & 0.99 & 0.01 \\
  Circ & Accuracy & Vanilla & 1.0 & 0.0 \\
  Circ & F1-score & Full & 0.99 & 0.01 \\
  Circ & F1-score & Vanilla & 1.0 & 0.0 \\
  LS & Accuracy & Full & 1.0 & 0.01 \\
  LS & Accuracy & Vanilla & 1.0 & 0.0 \\
  LS & F1-score & Full & 1.0 & 0.01 \\
  LS & F1-score & Vanilla & 1.0 & 0.0 \\
  Moon & Accuracy & Full & 0.99 & 0.04 \\
  Moon & Accuracy & Vanilla & 1.0 & 0.01 \\
  Moon & F1-score & Full & 0.99 & 0.04 \\
  Moon & F1-score & Vanilla & 1.0 & 0.01 \\
  OL & Accuracy & Full & 0.91 & 0.02 \\
  OL & Accuracy & Vanilla & 0.92 & 0.0 \\
  OL & F1-score & Full & 0.91 & 0.02 \\
  OL & F1-score & Vanilla & 0.92 & 0.0 \\\bottomrule

\end{longtable}

\subsubsection{Plausibility}\label{plausibility-1}

\begin{figure}

\centering{

\includegraphics[width=0.8\linewidth,height=\textheight,keepaspectratio]{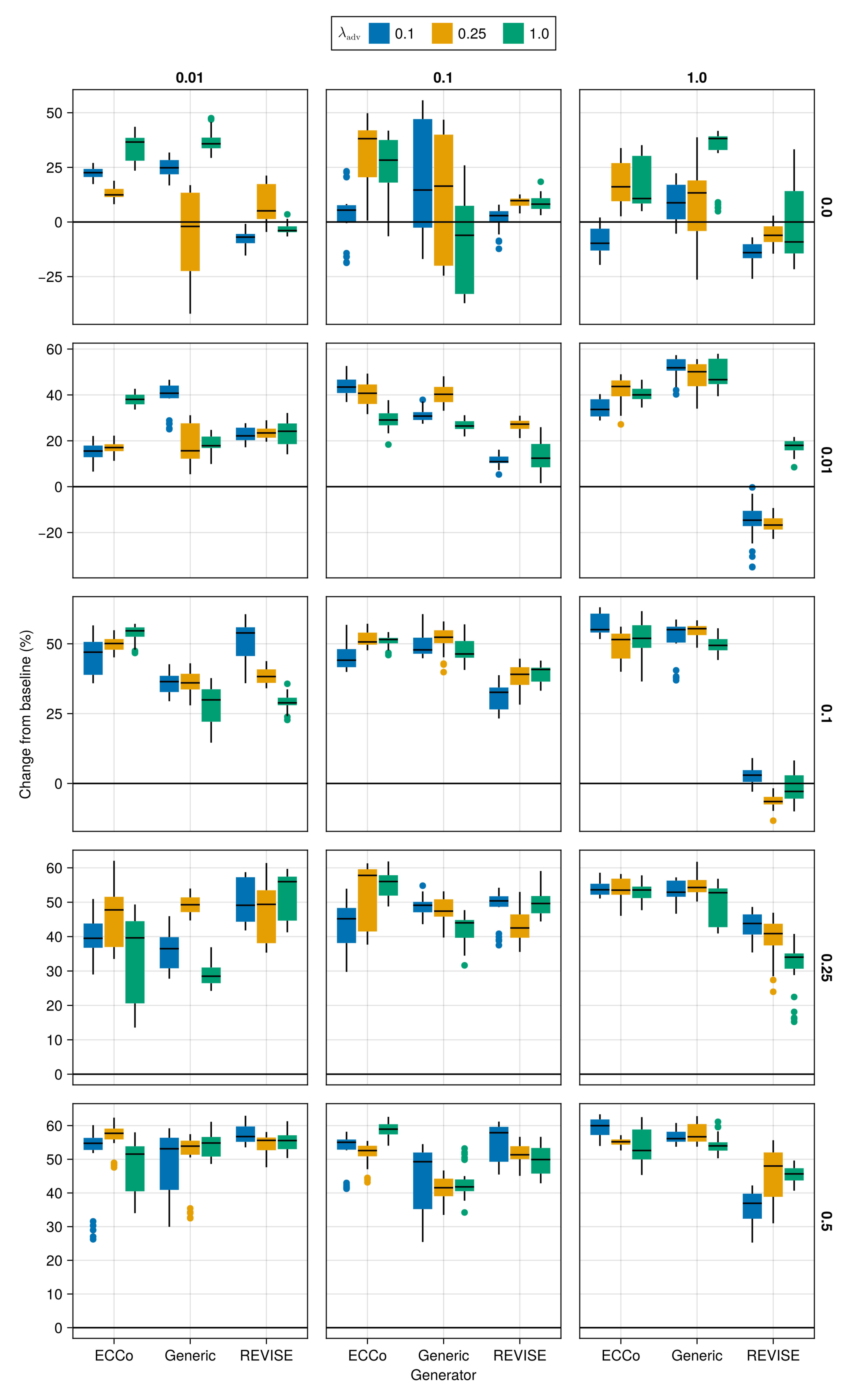}

}

\caption{\label{fig-grid-pen-plaus-circles}Average outcomes for the
plausibility measure across hyperparameters. This shows the \% change
from the baseline model for the distance-based implausibility metric
(\(\text{IP}\)). Boxplots indicate the variation across evaluation runs
and test settings (varying parameters for \emph{ECCCo}). Data: Circles.}

\end{figure}

\begin{figure}

\centering{

\includegraphics[width=0.8\linewidth,height=\textheight,keepaspectratio]{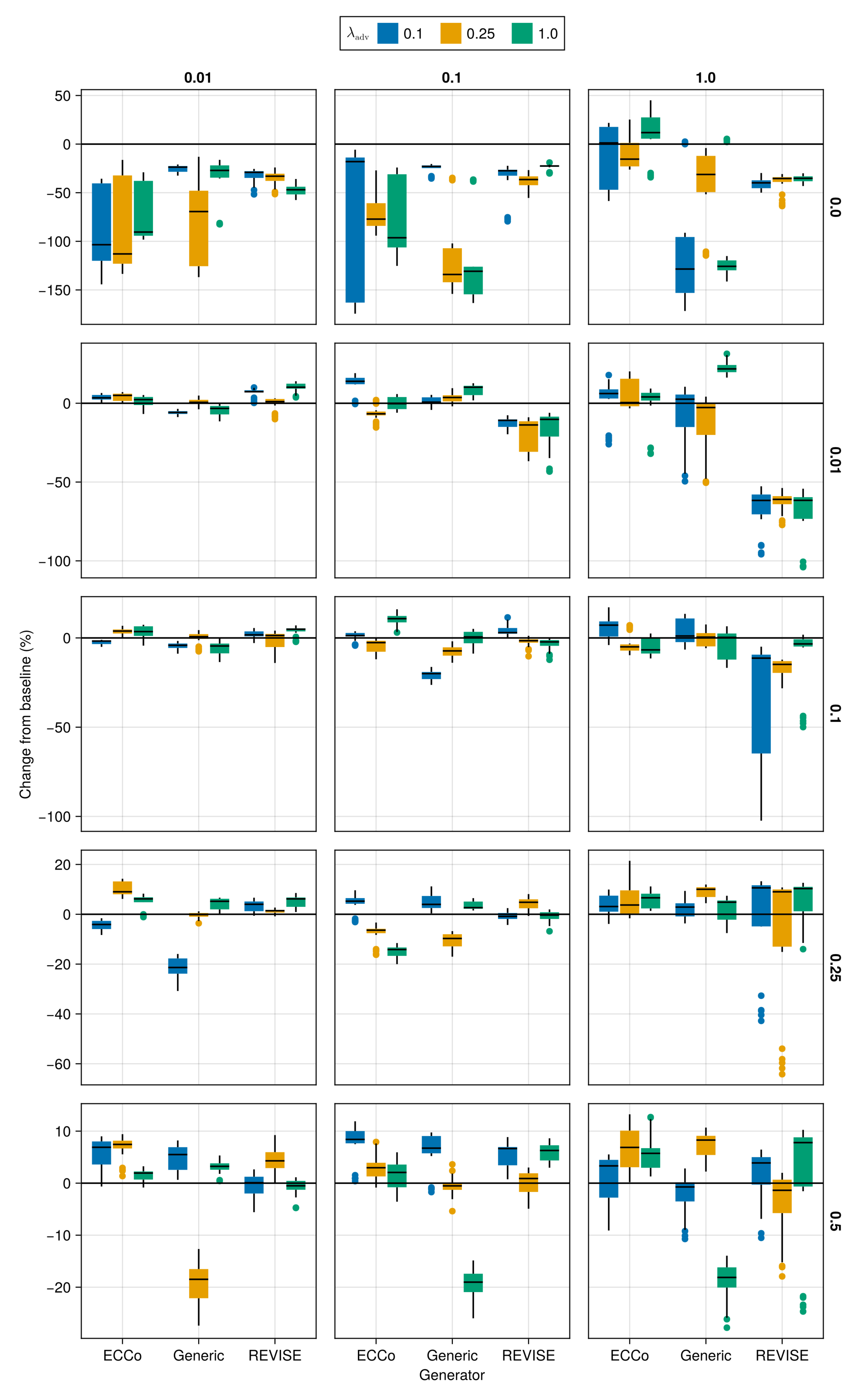}

}

\caption{\label{fig-grid-pen-plaus-lin_sep}Average outcomes for the
plausibility measure across hyperparameters. This shows the \% change
from the baseline model for the distance-based implausibility metric
(\(\text{IP}\)). Boxplots indicate the variation across evaluation runs
and test settings (varying parameters for \emph{ECCCo}). Data: Linearly
Separable.}

\end{figure}

\begin{figure}

\centering{

\includegraphics[width=0.8\linewidth,height=\textheight,keepaspectratio]{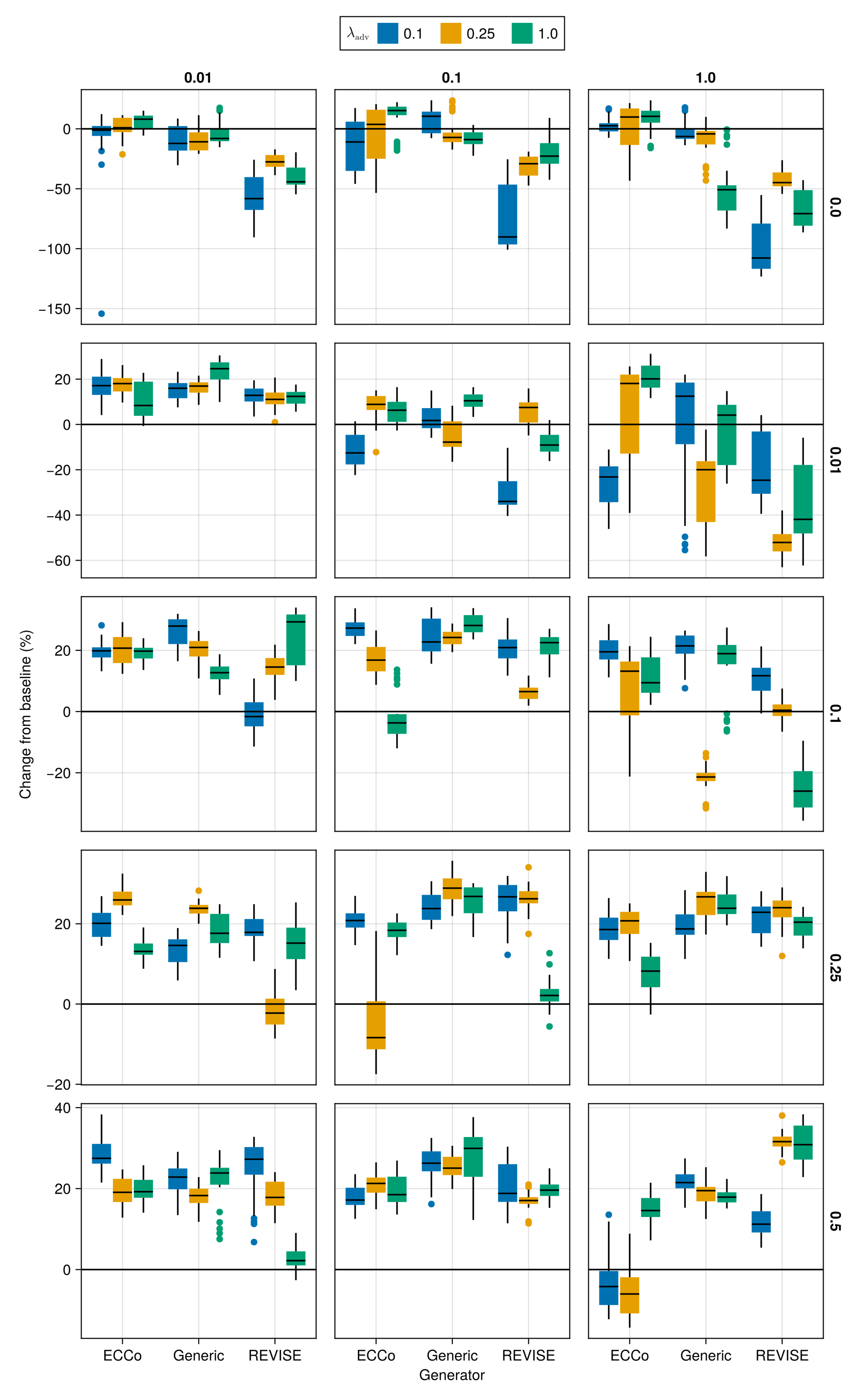}

}

\caption{\label{fig-grid-pen-plaus-moons}Average outcomes for the
plausibility measure across hyperparameters. This shows the \% change
from the baseline model for the distance-based implausibility metric
(\(\text{IP}\)). Boxplots indicate the variation across evaluation runs
and test settings (varying parameters for \emph{ECCCo}). Data: Moons.}

\end{figure}

\begin{figure}

\centering{

\includegraphics[width=0.8\linewidth,height=\textheight,keepaspectratio]{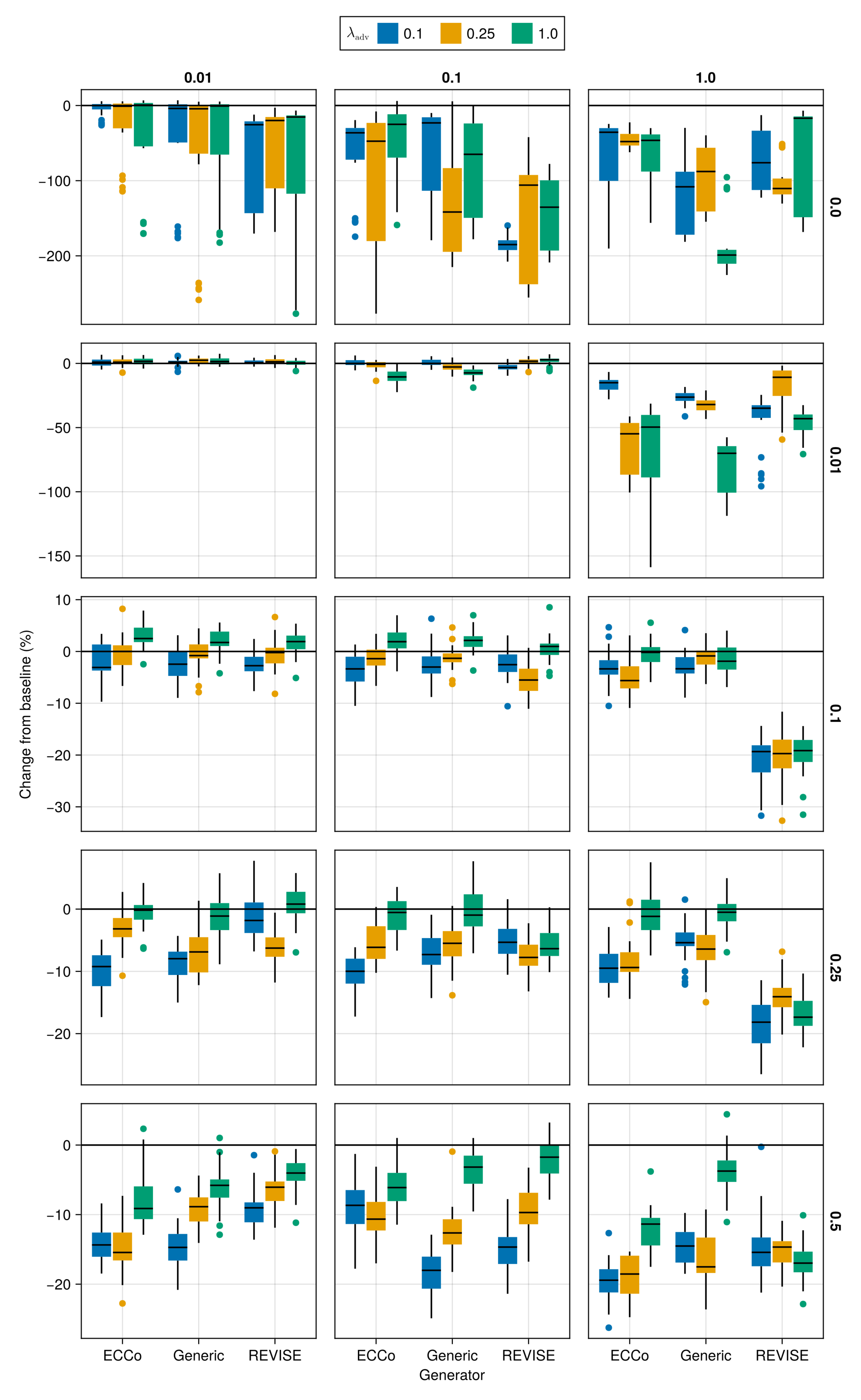}

}

\caption{\label{fig-grid-pen-plaus-over}Average outcomes for the
plausibility measure across hyperparameters. This shows the \% change
from the baseline model for the distance-based implausibility metric
(\(\text{IP}\)). Boxplots indicate the variation across evaluation runs
and test settings (varying parameters for \emph{ECCCo}). Data:
Overlapping.}

\end{figure}

The results with respect to the plausibility measure are shown in
Figure~\ref{fig-grid-pen-plaus-circles} to
Figure~\ref{fig-grid-pen-plaus-over}.

\subsubsection{Cost}\label{cost-1}

\begin{figure}

\centering{

\includegraphics[width=0.8\linewidth,height=\textheight,keepaspectratio]{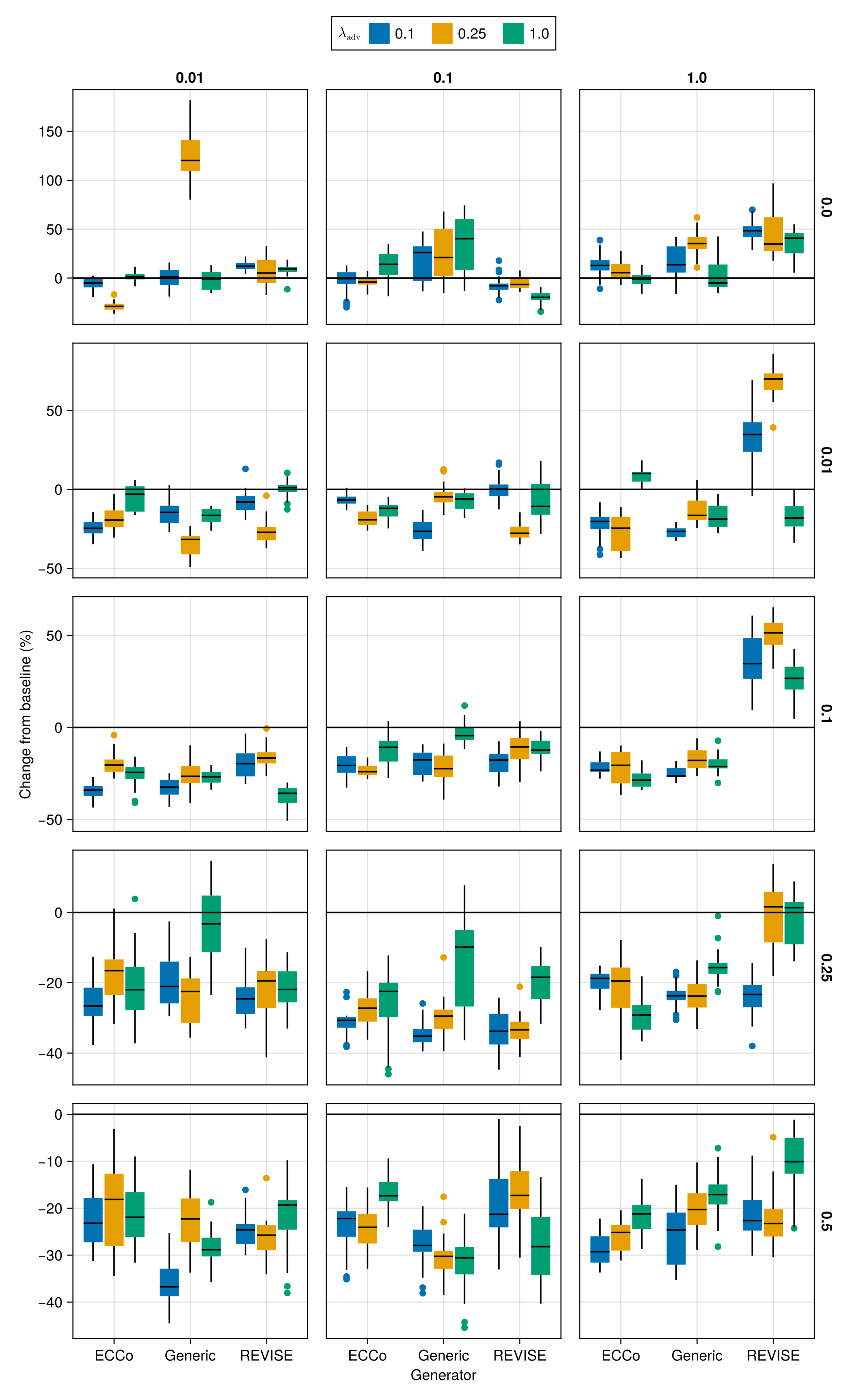}

}

\caption{\label{fig-grid-pen-cost-circles}Average outcomes for the cost
measure across hyperparameters. This shows the \% change from the
baseline model for the distance-based cost metric
(\citeproc{ref-wachter2017counterfactual}{Wachter, Mittelstadt, and
Russell 2017}). Boxplots indicate the variation across evaluation runs
and test settings (varying parameters for \emph{ECCCo}). Data: Circles.}

\end{figure}

\begin{figure}

\centering{

\includegraphics[width=0.8\linewidth,height=\textheight,keepaspectratio]{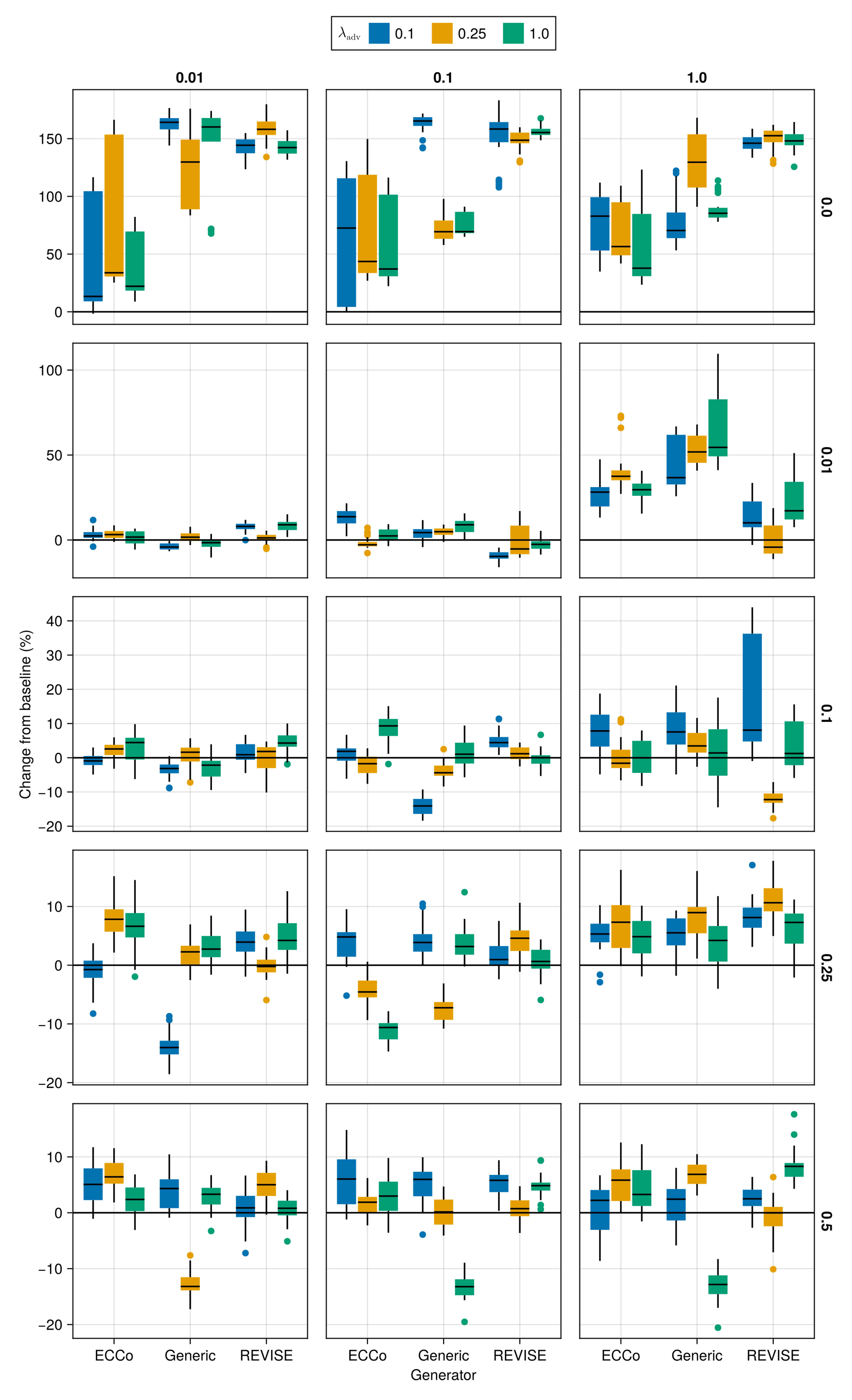}

}

\caption{\label{fig-grid-pen-cost-lin_sep}Average outcomes for the cost
measure across hyperparameters. This shows the \% change from the
baseline model for the distance-based cost metric
(\citeproc{ref-wachter2017counterfactual}{Wachter, Mittelstadt, and
Russell 2017}). Boxplots indicate the variation across evaluation runs
and test settings (varying parameters for \emph{ECCCo}). Data: Linearly
Separable.}

\end{figure}

\begin{figure}

\centering{

\includegraphics[width=0.8\linewidth,height=\textheight,keepaspectratio]{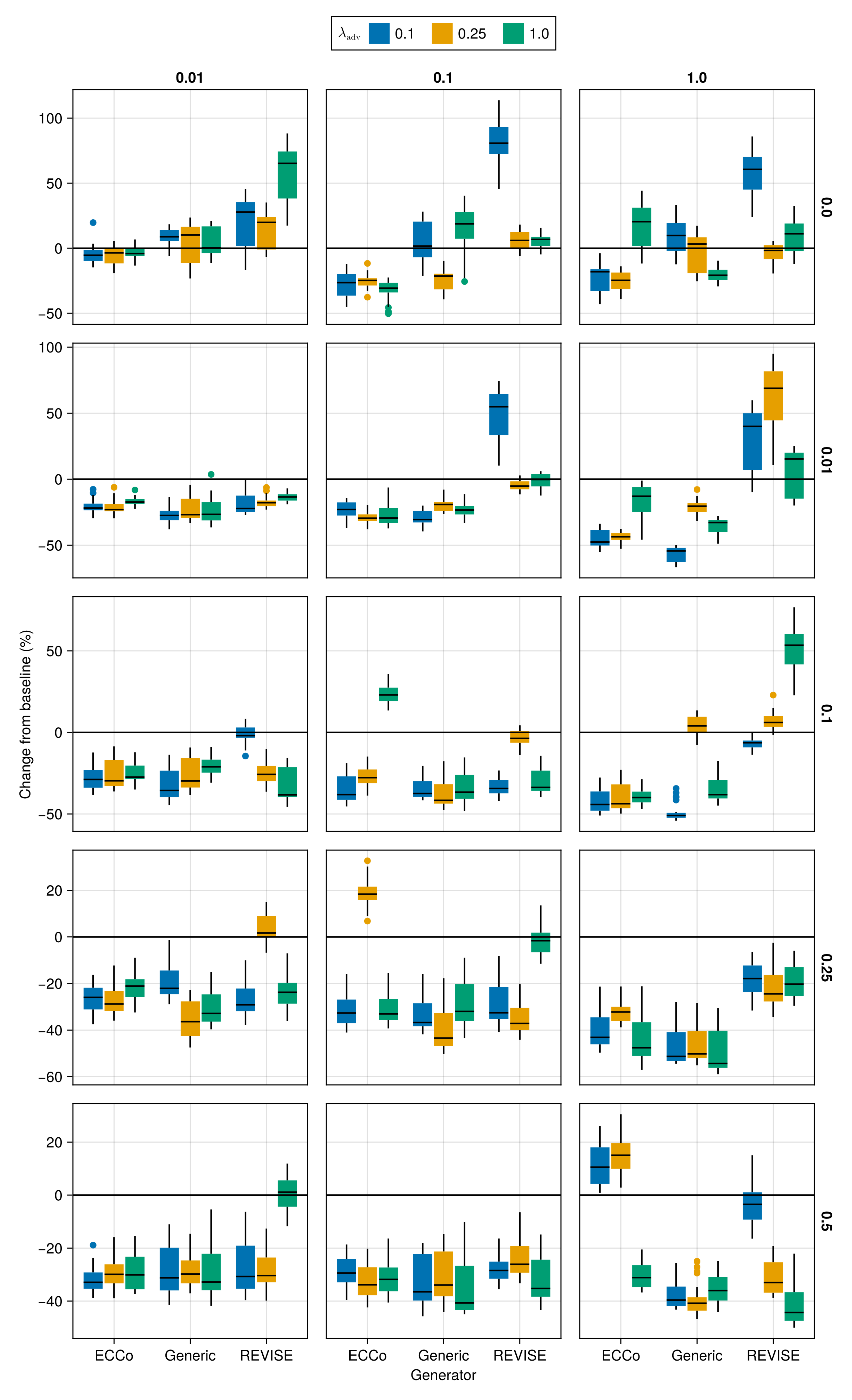}

}

\caption{\label{fig-grid-pen-cost-moons}Average outcomes for the cost
measure across hyperparameters. This shows the \% change from the
baseline model for the distance-based cost metric
(\citeproc{ref-wachter2017counterfactual}{Wachter, Mittelstadt, and
Russell 2017}). Boxplots indicate the variation across evaluation runs
and test settings (varying parameters for \emph{ECCCo}). Data: Moons.}

\end{figure}

\begin{figure}

\centering{

\includegraphics[width=0.8\linewidth,height=\textheight,keepaspectratio]{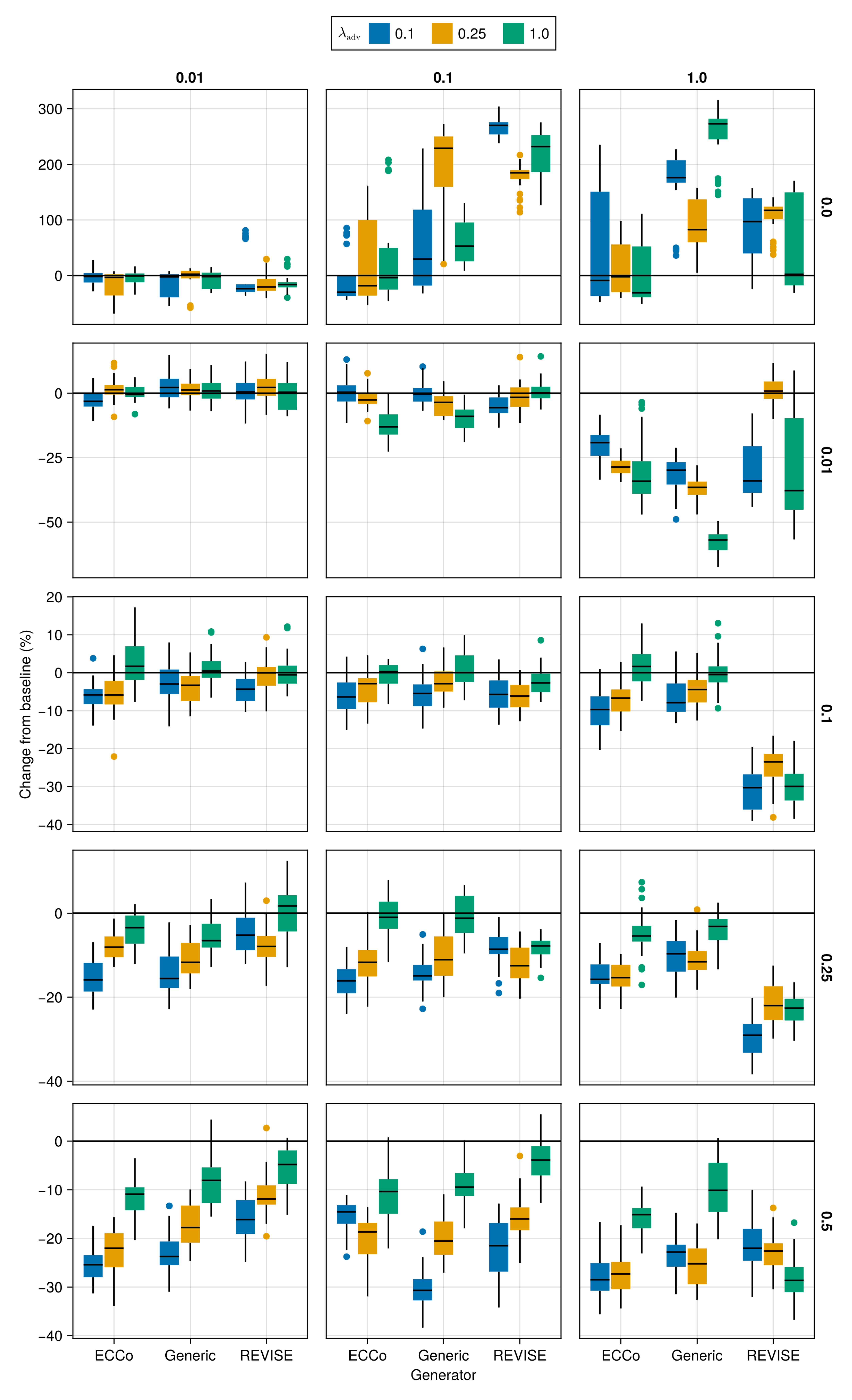}

}

\caption{\label{fig-grid-pen-cost-over}Average outcomes for the cost
measure across hyperparameters. This shows the \% change from the
baseline model for the distance-based cost metric
(\citeproc{ref-wachter2017counterfactual}{Wachter, Mittelstadt, and
Russell 2017}). Boxplots indicate the variation across evaluation runs
and test settings (varying parameters for \emph{ECCCo}). Data:
Overlapping.}

\end{figure}

The results with respect to the cost measure are shown in
Figure~\ref{fig-grid-pen-cost-circles} to
Figure~\ref{fig-grid-pen-cost-over}.

\subsection{Other Parameters}\label{sec-app-grid-train}

The hyperparameter grid with other varying training parameters is shown
in Note~\ref{nte-train-final-run-train}. The corresponding evaluation
grid used for these experiments is shown in
Note~\ref{nte-train-final-run-eval}.

\begin{tcolorbox}[enhanced jigsaw, arc=.35mm, colback=white, bottomrule=.15mm, toptitle=1mm, leftrule=.75mm, titlerule=0mm, toprule=.15mm, breakable, colframe=quarto-callout-note-color-frame, left=2mm, opacityback=0, bottomtitle=1mm, opacitybacktitle=0.6, title={Note \ref*{nte-train-final-run-train}space Training Phase}, colbacktitle=quarto-callout-note-color!10!white, rightrule=.15mm, coltitle=black]

\quartocalloutnte{nte-train-final-run-train} 

\begin{itemize}
\tightlist
\item
  Generator: \texttt{ecco,\ generic,\ revise}
\item
  Model: \texttt{mlp}
\item
  Training Parameters:

  \begin{itemize}
  \tightlist
  \item
    Burnin: \texttt{0.0,\ 0.5}
  \item
    No.~Counterfactuals: \texttt{100,\ 1000}
  \item
    No.~Epochs: \texttt{50,\ 100}
  \item
    Objective: \texttt{full,\ vanilla}
  \end{itemize}
\end{itemize}

\end{tcolorbox}

\begin{tcolorbox}[enhanced jigsaw, arc=.35mm, colback=white, bottomrule=.15mm, toptitle=1mm, leftrule=.75mm, titlerule=0mm, toprule=.15mm, breakable, colframe=quarto-callout-note-color-frame, left=2mm, opacityback=0, bottomtitle=1mm, opacitybacktitle=0.6, title={Note \ref*{nte-train-final-run-eval}space Evaluation Phase}, colbacktitle=quarto-callout-note-color!10!white, rightrule=.15mm, coltitle=black]

\quartocalloutnte{nte-train-final-run-eval} 

\begin{itemize}
\tightlist
\item
  Generator Parameters:

  \begin{itemize}
  \tightlist
  \item
    \(\lambda_{\text{egy}}\): \texttt{0.1,\ 0.5,\ 1.0,\ 5.0,\ 10.0}
  \end{itemize}
\end{itemize}

\end{tcolorbox}

\subsubsection{Predictive Performance}\label{predictive-performance-4}

Predictive performance measures for this grid search are shown in
Table~\ref{tbl-acc-train}.

\begin{longtable}{ccccc}

\caption{\label{tbl-acc-train}Predictive performance measures by dataset
and objective averaged across training-phase parameters
(Note~\ref{nte-train-final-run-train}) and evaluation-phase parameters
(Note~\ref{nte-train-final-run-eval}).}

\tabularnewline

  \toprule
  \textbf{Dataset} & \textbf{Variable} & \textbf{Objective} & \textbf{Mean} & \textbf{Se} \\\midrule
  \endfirsthead
  \toprule
  \textbf{Dataset} & \textbf{Variable} & \textbf{Objective} & \textbf{Mean} & \textbf{Se} \\\midrule
  \endhead
  \bottomrule
  \multicolumn{5}{r}{Continuing table below.}\\
  \bottomrule
  \endfoot
  \endlastfoot
  Circ & Accuracy & Full & 0.99 & 0.0 \\
  Circ & Accuracy & Vanilla & 1.0 & 0.0 \\
  Circ & F1-score & Full & 0.99 & 0.0 \\
  Circ & F1-score & Vanilla & 1.0 & 0.0 \\
  LS & Accuracy & Full & 1.0 & 0.0 \\
  LS & Accuracy & Vanilla & 1.0 & 0.0 \\
  LS & F1-score & Full & 1.0 & 0.0 \\
  LS & F1-score & Vanilla & 1.0 & 0.0 \\
  Moon & Accuracy & Full & 1.0 & 0.01 \\
  Moon & Accuracy & Vanilla & 0.99 & 0.02 \\
  Moon & F1-score & Full & 1.0 & 0.01 \\
  Moon & F1-score & Vanilla & 0.99 & 0.02 \\
  OL & Accuracy & Full & 0.91 & 0.01 \\
  OL & Accuracy & Vanilla & 0.92 & 0.0 \\
  OL & F1-score & Full & 0.91 & 0.01 \\
  OL & F1-score & Vanilla & 0.92 & 0.0 \\\bottomrule

\end{longtable}

\subsubsection{Plausibility}\label{plausibility-2}

\begin{figure}

\centering{

\includegraphics[width=0.8\linewidth,height=\textheight,keepaspectratio]{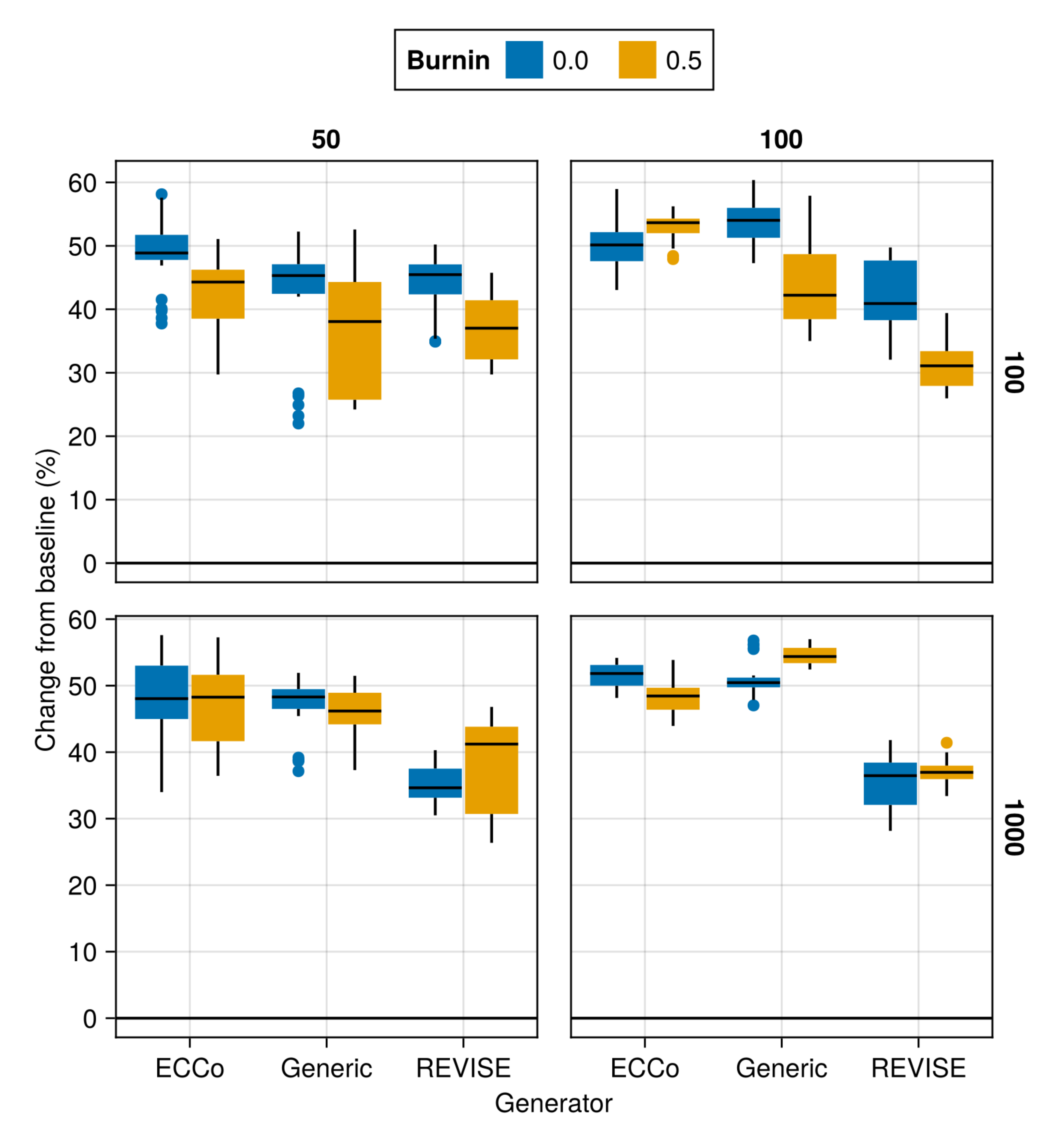}

}

\caption{\label{fig-grid-train-plaus-circles}Average outcomes for the
plausibility measure across hyperparameters. This shows the \% change
from the baseline model for the distance-based implausibility metric
(\(\text{IP}\)). Boxplots indicate the variation across evaluation runs
and test settings (varying parameters for \emph{ECCCo}). Data: Circles.}

\end{figure}

\begin{figure}

\centering{

\includegraphics[width=0.8\linewidth,height=\textheight,keepaspectratio]{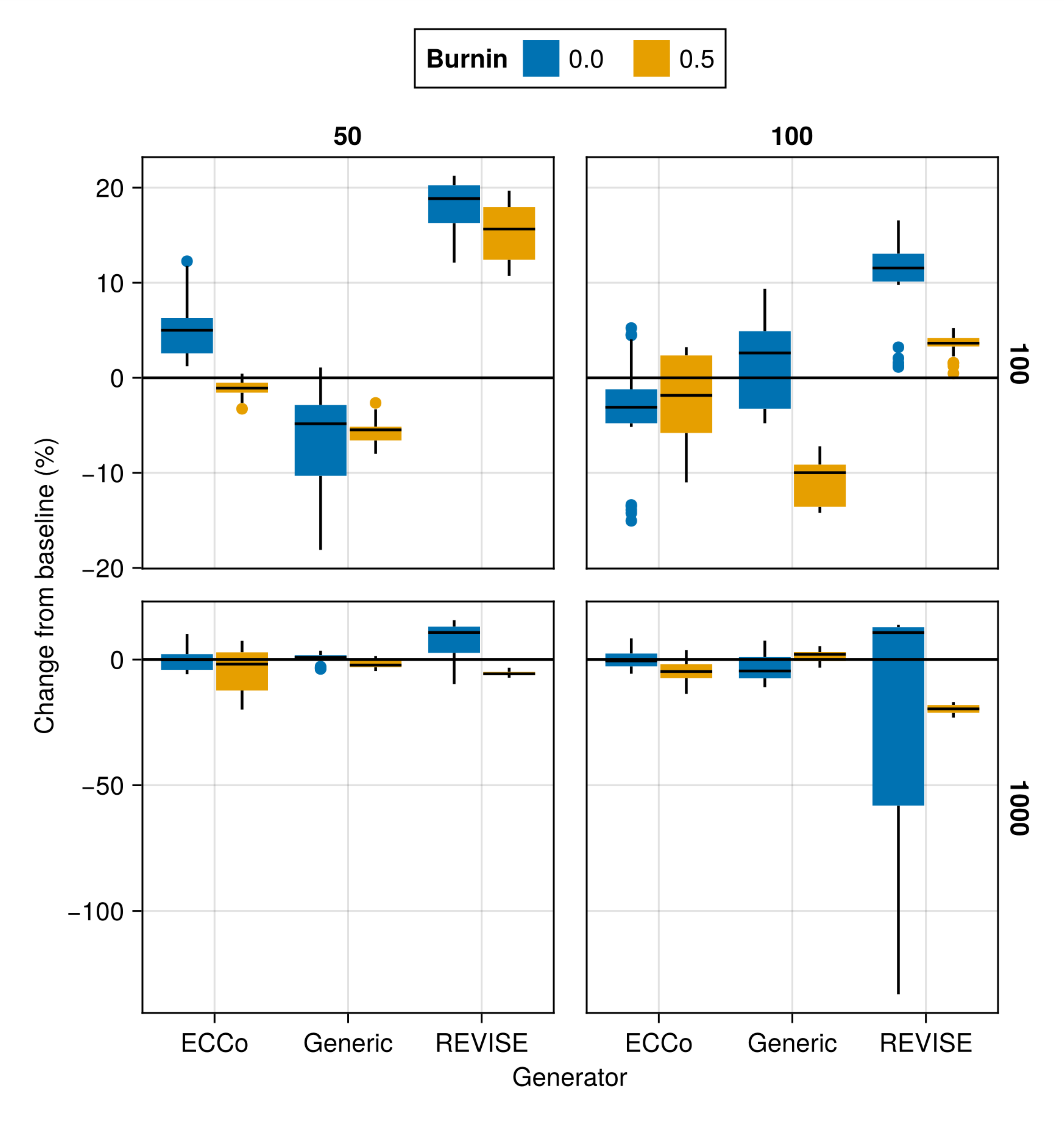}

}

\caption{\label{fig-grid-train-plaus-lin_sep}Average outcomes for the
plausibility measure across hyperparameters. This shows the \% change
from the baseline model for the distance-based implausibility metric
(\(\text{IP}\)). Boxplots indicate the variation across evaluation runs
and test settings (varying parameters for \emph{ECCCo}). Data: Linearly
Separable.}

\end{figure}

\begin{figure}

\centering{

\includegraphics[width=0.8\linewidth,height=\textheight,keepaspectratio]{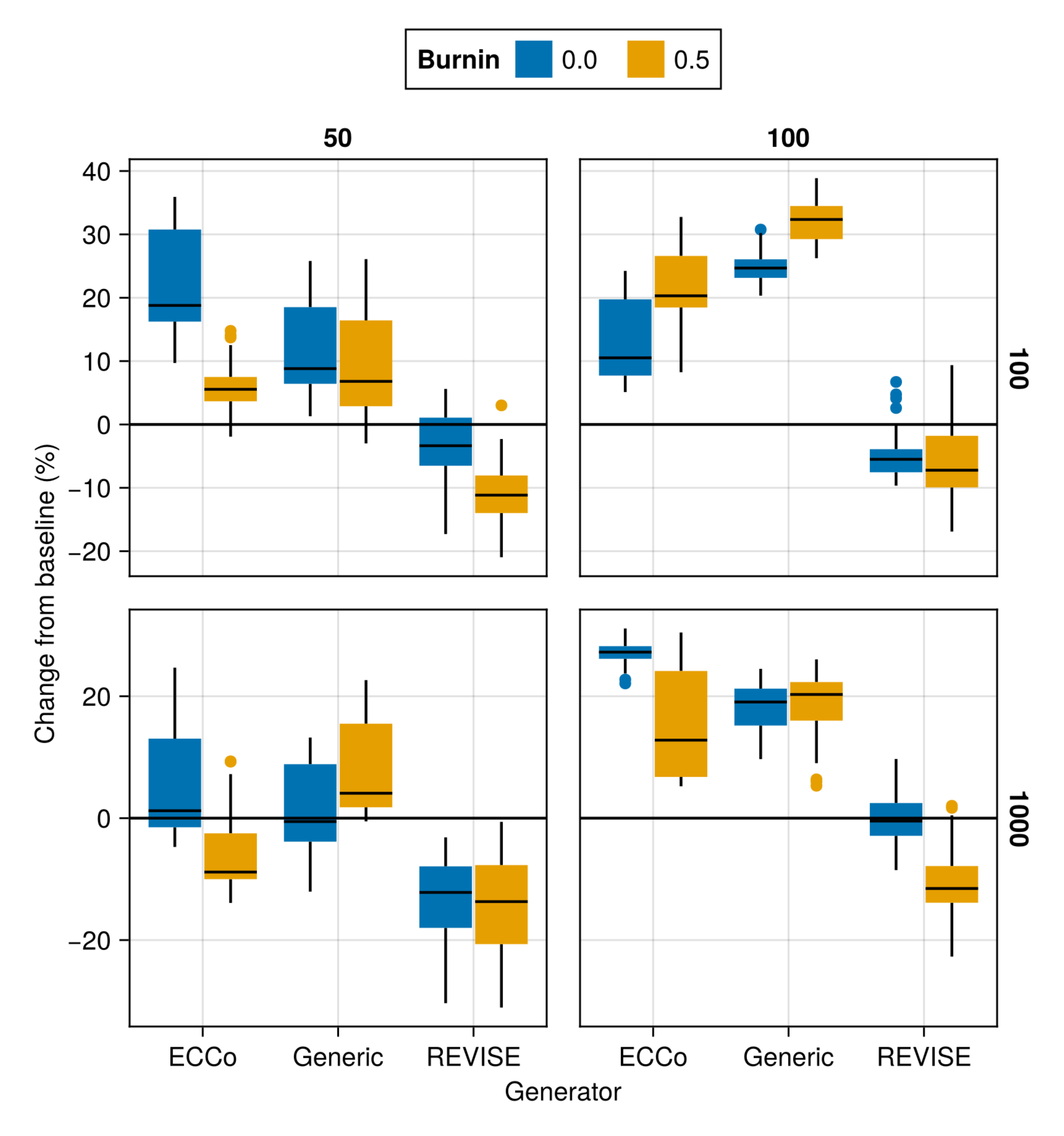}

}

\caption{\label{fig-grid-train-plaus-moons}Average outcomes for the
plausibility measure across hyperparameters. This shows the \% change
from the baseline model for the distance-based implausibility metric
(\(\text{IP}\)). Boxplots indicate the variation across evaluation runs
and test settings (varying parameters for \emph{ECCCo}). Data: Moons.}

\end{figure}

\begin{figure}

\centering{

\includegraphics[width=0.8\linewidth,height=\textheight,keepaspectratio]{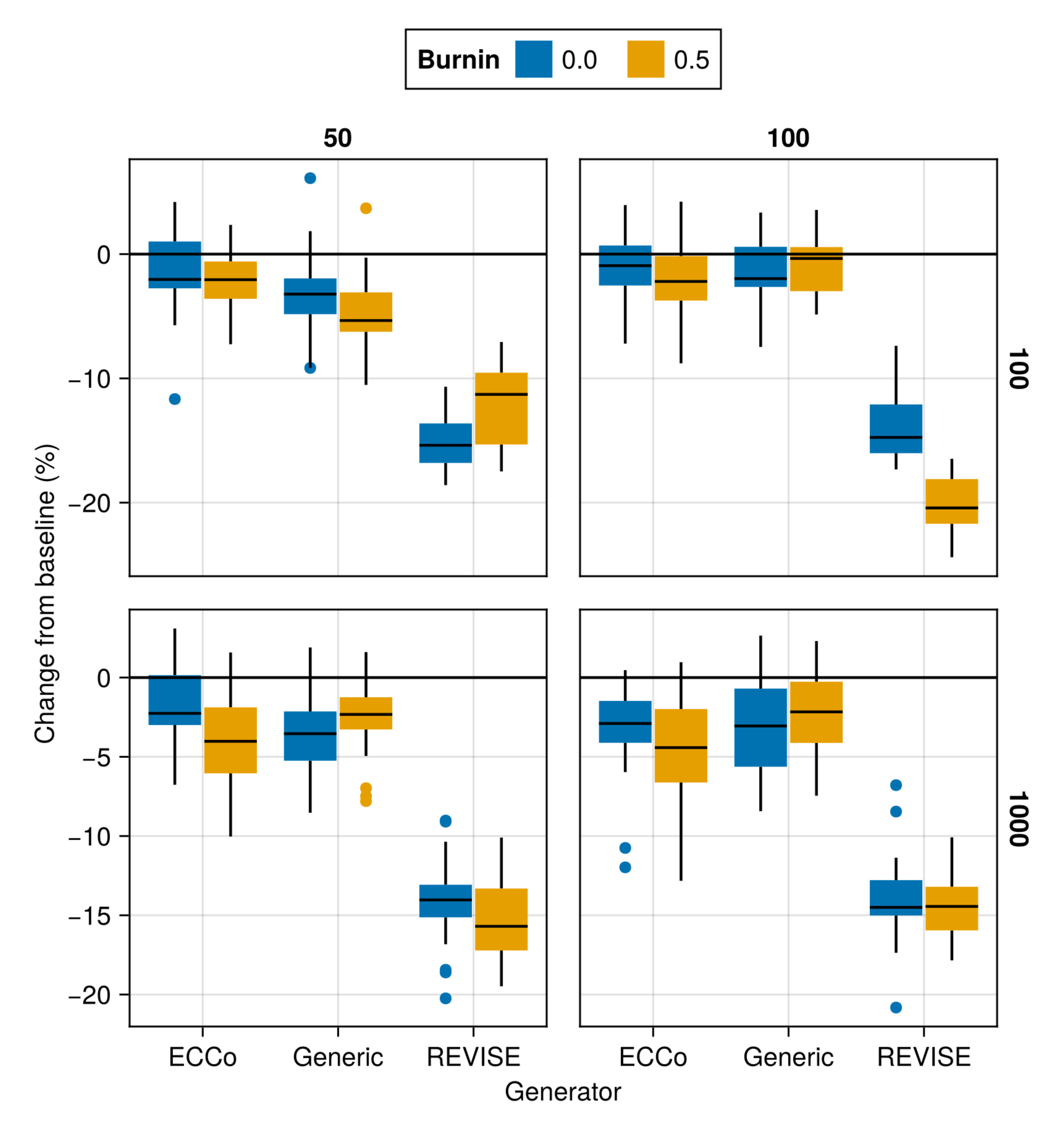}

}

\caption{\label{fig-grid-train-plaus-over}Average outcomes for the
plausibility measure across hyperparameters. This shows the \% change
from the baseline model for the distance-based implausibility metric
(\(\text{IP}\)). Boxplots indicate the variation across evaluation runs
and test settings (varying parameters for \emph{ECCCo}). Data:
Overlapping.}

\end{figure}

The results with respect to the plausibility measure are shown in
Figure~\ref{fig-grid-train-plaus-circles} to
Figure~\ref{fig-grid-train-plaus-over}.

\subsubsection{Cost}\label{cost-2}

\begin{figure}

\centering{

\includegraphics[width=0.8\linewidth,height=\textheight,keepaspectratio]{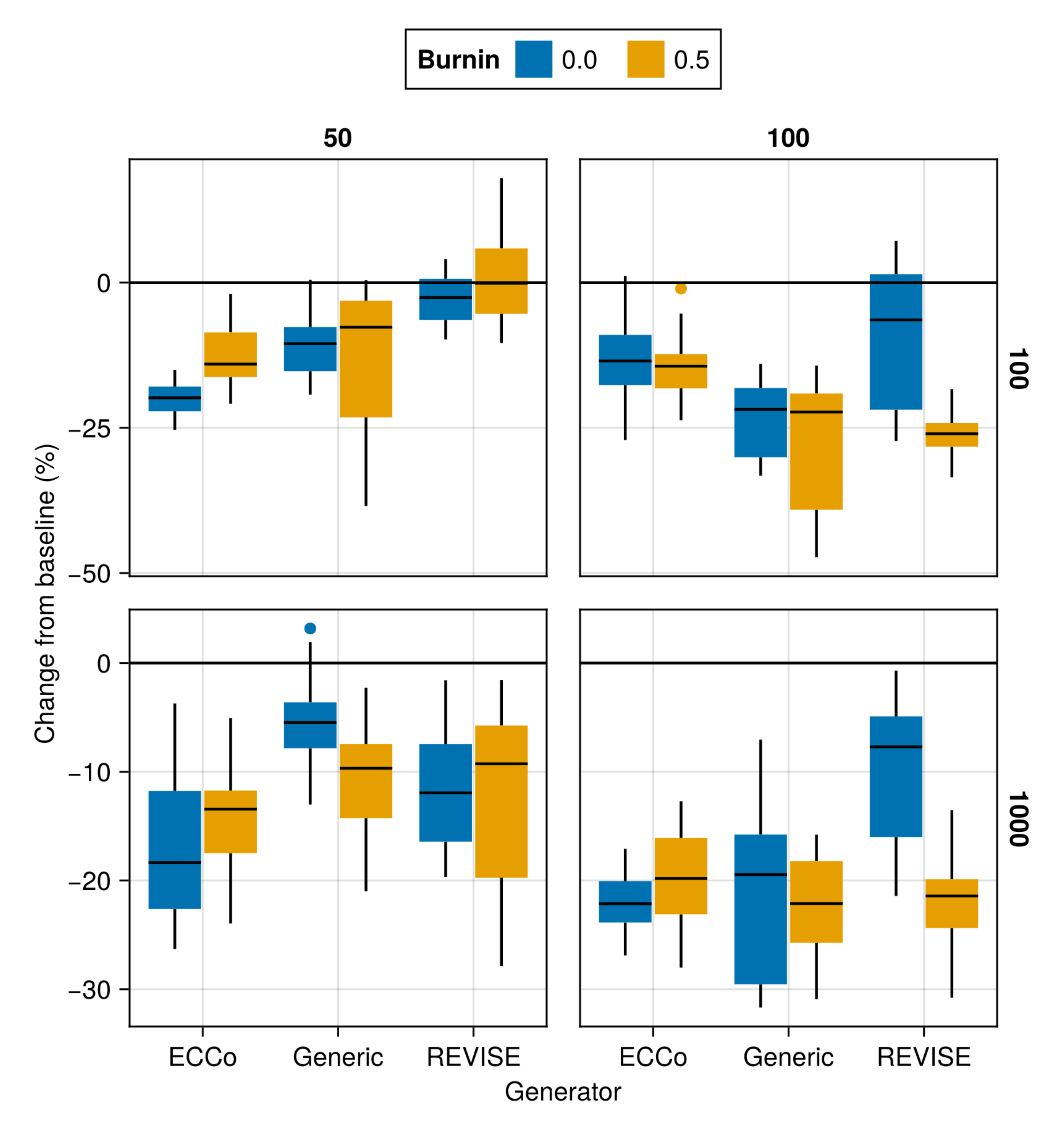}

}

\caption{\label{fig-grid-train-cost-circles}Average outcomes for the
cost measure across hyperparameters. This shows the \% change from the
baseline model for the distance-based cost metric
(\citeproc{ref-wachter2017counterfactual}{Wachter, Mittelstadt, and
Russell 2017}). Boxplots indicate the variation across evaluation runs
and test settings (varying parameters for \emph{ECCCo}). Data: Circles.}

\end{figure}

\begin{figure}

\centering{

\includegraphics[width=0.8\linewidth,height=\textheight,keepaspectratio]{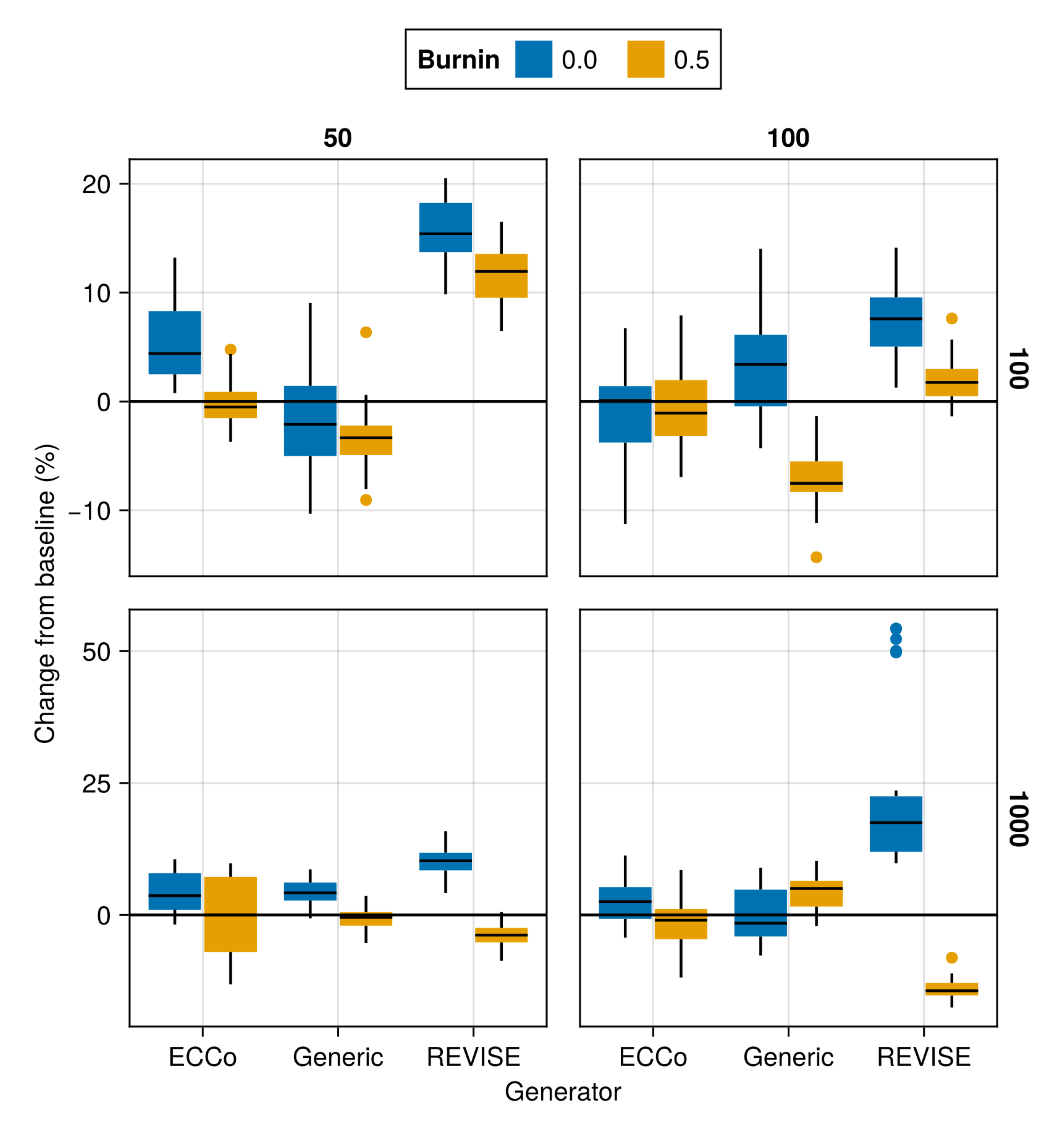}

}

\caption{\label{fig-grid-train-cost-lin_sep}Average outcomes for the
cost measure across hyperparameters. This shows the \% change from the
baseline model for the distance-based cost metric
(\citeproc{ref-wachter2017counterfactual}{Wachter, Mittelstadt, and
Russell 2017}). Boxplots indicate the variation across evaluation runs
and test settings (varying parameters for \emph{ECCCo}). Data: Linearly
Separable.}

\end{figure}

\begin{figure}

\centering{

\includegraphics[width=0.8\linewidth,height=\textheight,keepaspectratio]{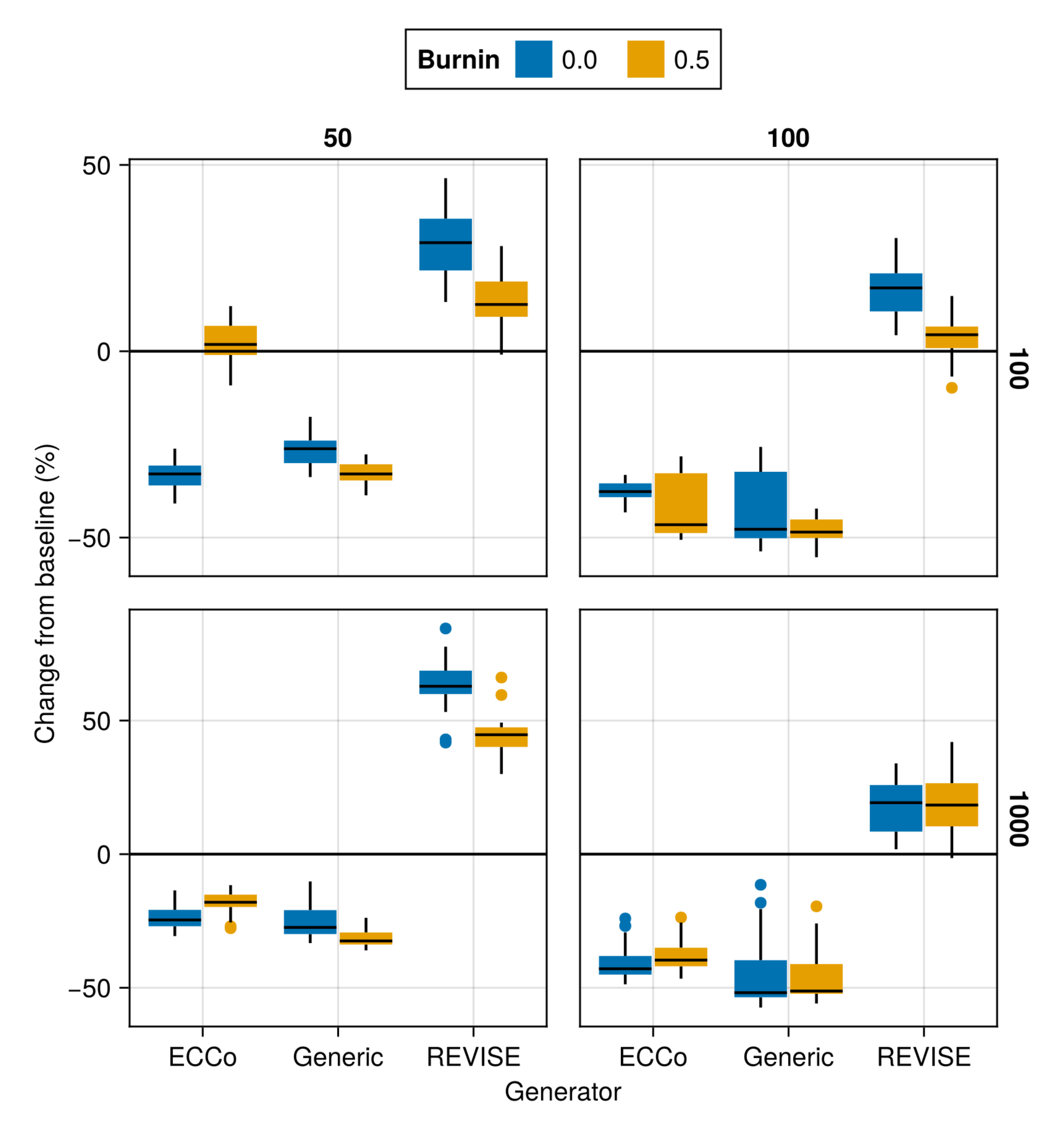}

}

\caption{\label{fig-grid-train-cost-moons}Average outcomes for the cost
measure across hyperparameters. This shows the \% change from the
baseline model for the distance-based cost metric
(\citeproc{ref-wachter2017counterfactual}{Wachter, Mittelstadt, and
Russell 2017}). Boxplots indicate the variation across evaluation runs
and test settings (varying parameters for \emph{ECCCo}). Data: Moons.}

\end{figure}

\begin{figure}

\centering{

\includegraphics[width=0.8\linewidth,height=\textheight,keepaspectratio]{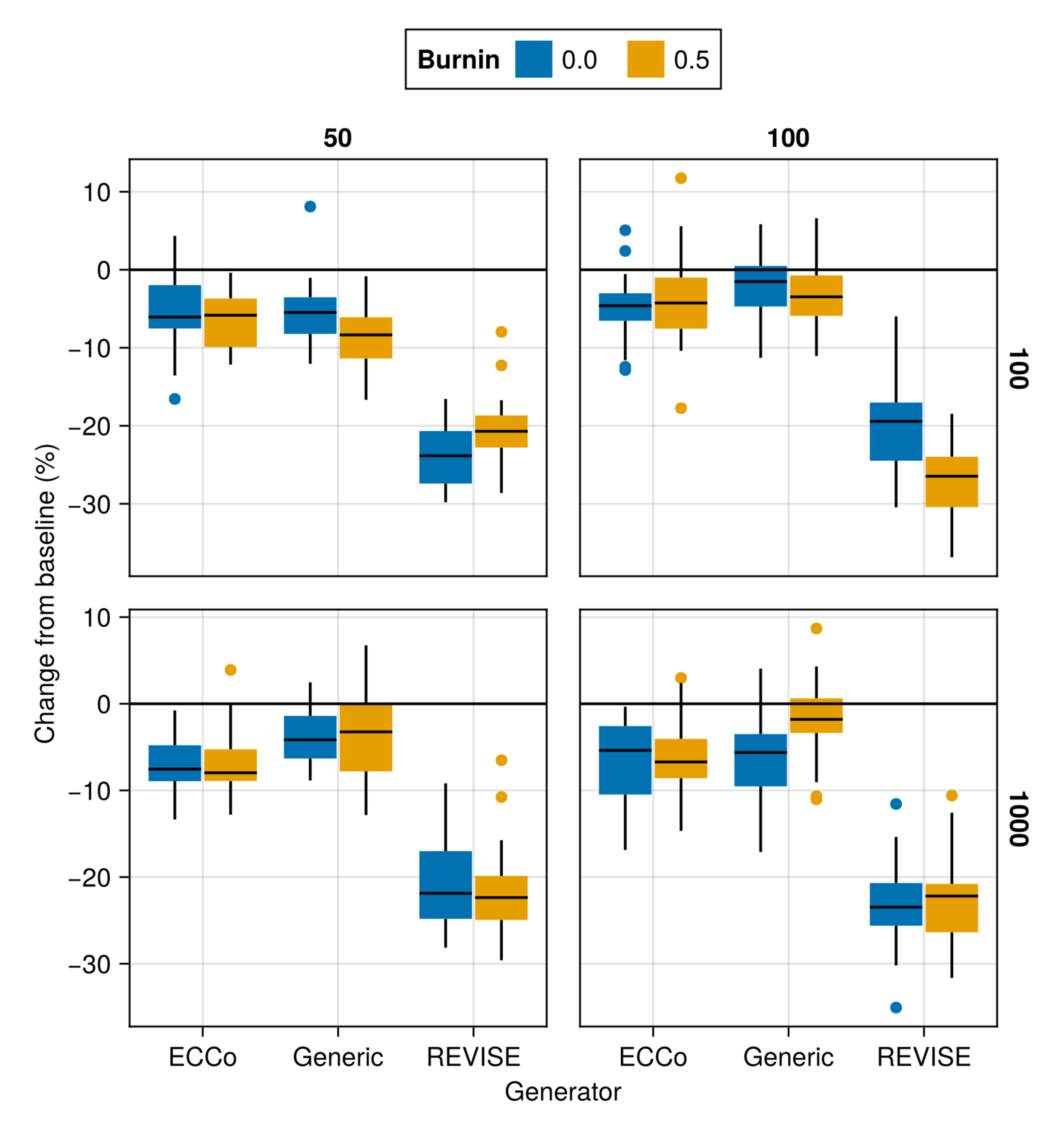}

}

\caption{\label{fig-grid-train-cost-over}Average outcomes for the cost
measure across hyperparameters. This shows the \% change from the
baseline model for the distance-based cost metric
(\citeproc{ref-wachter2017counterfactual}{Wachter, Mittelstadt, and
Russell 2017}). Boxplots indicate the variation across evaluation runs
and test settings (varying parameters for \emph{ECCCo}). Data:
Overlapping.}

\end{figure}

The results with respect to the cost measure are shown in
Figure~\ref{fig-grid-train-cost-circles} to
Figure~\ref{fig-grid-train-cost-over}.

\FloatBarrier

\section{Tuning Key Parameters}\label{sec-app-tune}

Based on the findings from our initial large grid searches
(Section~\ref{sec-app-grid}), we tune selected hyperparameters for all
datasets: namely, the decision threshold \(\tau\) and the strength of
the energy regularization \(\lambda_{\text{reg}}\). The final
hyperparameter choices for each dataset are presented in
Table~\ref{tbl-final-params} in Section~\ref{sec-app-main}. Detailed
results for each data set are shown in Figure~\ref{fig-tune-plaus-adult}
to Figure~\ref{fig-tune-mat-over}. From Table~\ref{tbl-final-params}, we
notice that the same decision threshold of \(\tau=0.5\) is optimal for
all but on dataset. We attribute this to the fact that a low decision
threshold results in a higher share of mature counterfactuals and hence
more opportunities for the model to learn from examples
(Figure~\ref{fig-tune-mat-adult} to Figure~\ref{fig-tune-mat-over}).
This has played a role in particular for our real-world tabular datasets
and MNIST, which suffered from low levels of maturity for higher
decision thresholds. In cases where maturity is not an issue, as for
\emph{Moons}, higher decision thresholds lead to better outcomes, which
may have to do with the fact that the resulting counterfactuals are more
faithful to the model. Concerning the regularization strength, we find
somewhat high variation across datasets. Most notably, we find that
relatively low levels of regularization are optimal for MNIST. We
hypothesize that this finding may be attributed to the uniform scaling
of all input features (digits).

Finally, to increase the proportion of mature counterfactuals for some
datasets, we have also investigated the effect on the learning rate
\(\eta\) for the counterfactual search and even smaller regularization
strengths for a fixed decision threshold of 0.5
(Figure~\ref{fig-tune_lr-plaus-adult} to
Figure~\ref{fig-tune_lr-plaus-over}). For the given low decision
threshold, we find that the learning rate has no discernable impact on
the proportion of mature counterfactuals
(Figure~\ref{fig-tune_lr-mat-adult} to
Figure~\ref{fig-tune_lr-mat-over}). We do notice, however, that the
results for MNIST are much improved when using a low value
\(\lambda_{\text{reg}}\), the strength for the engery regularization:
plausibility is increased by up to \textasciitilde10\%
(Figure~\ref{fig-tune_lr-plaus-mnist}) and the proportion of mature
counterfactuals reaches 100\%.

One consideration worth exploring is to combine high decision thresholds
with high learning rates, which we have not investigated here.

\subsection{Key Parameters}\label{sec-app-tune-key}

The hyperparameter grid for tuning key parameters is shown in
Note~\ref{nte-tune-train}. The corresponding evaluation grid used for
these experiments is shown in Note~\ref{nte-tune-eval}.

\begin{tcolorbox}[enhanced jigsaw, arc=.35mm, colback=white, bottomrule=.15mm, toptitle=1mm, leftrule=.75mm, titlerule=0mm, toprule=.15mm, breakable, colframe=quarto-callout-note-color-frame, left=2mm, opacityback=0, bottomtitle=1mm, opacitybacktitle=0.6, title={Note \ref*{nte-tune-train}space Training Phase}, colbacktitle=quarto-callout-note-color!10!white, rightrule=.15mm, coltitle=black]

\quartocalloutnte{nte-tune-train} 

\begin{itemize}
\tightlist
\item
  Generator Parameters:

  \begin{itemize}
  \tightlist
  \item
    Decision Threshold: \texttt{0.5,\ 0.75,\ 0.9}
  \end{itemize}
\item
  Model: \texttt{mlp}
\item
  Training Parameters:

  \begin{itemize}
  \tightlist
  \item
    \(\lambda_{\text{reg}}\): \texttt{0.1,\ 0.25,\ 0.5}
  \item
    Objective: \texttt{full,\ vanilla}
  \end{itemize}
\end{itemize}

\end{tcolorbox}

\begin{tcolorbox}[enhanced jigsaw, arc=.35mm, colback=white, bottomrule=.15mm, toptitle=1mm, leftrule=.75mm, titlerule=0mm, toprule=.15mm, breakable, colframe=quarto-callout-note-color-frame, left=2mm, opacityback=0, bottomtitle=1mm, opacitybacktitle=0.6, title={Note \ref*{nte-tune-eval}space Evaluation Phase}, colbacktitle=quarto-callout-note-color!10!white, rightrule=.15mm, coltitle=black]

\quartocalloutnte{nte-tune-eval} 

\begin{itemize}
\tightlist
\item
  Generator Parameters:

  \begin{itemize}
  \tightlist
  \item
    \(\lambda_{\text{egy}}\): \texttt{0.1,\ 0.5,\ 1.0,\ 5.0,\ 10.0}
  \end{itemize}
\end{itemize}

\end{tcolorbox}

\subsubsection{Plausibility}\label{plausibility-3}

\begin{figure}

\centering{

\includegraphics[width=1\linewidth,height=\textheight,keepaspectratio]{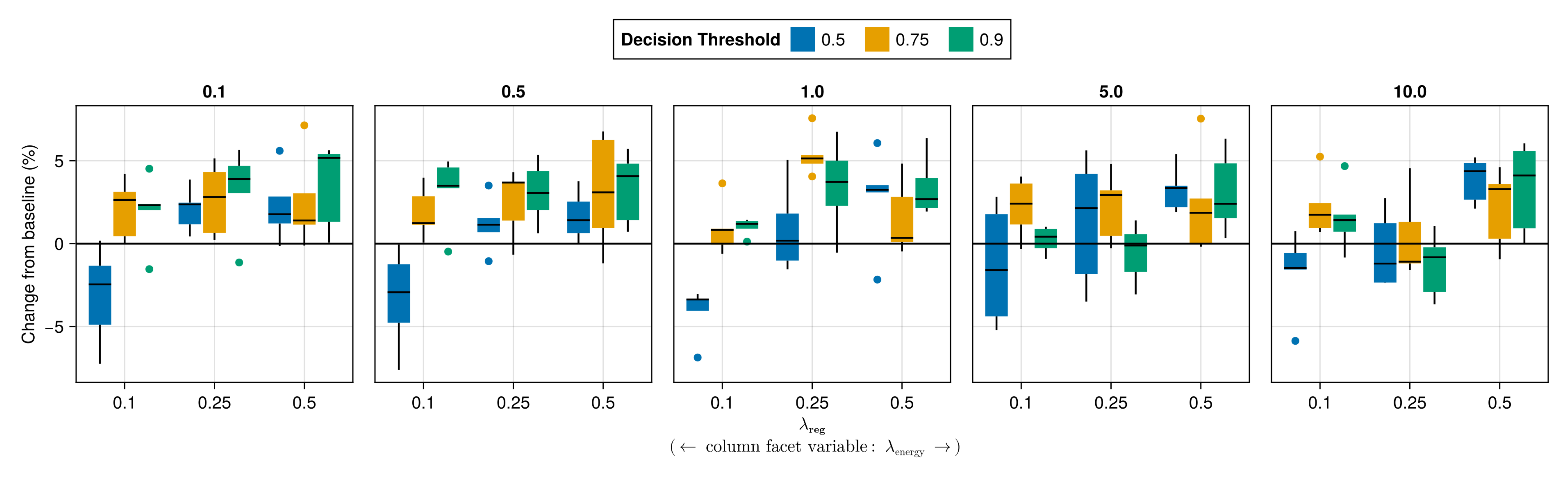}

}

\caption{\label{fig-tune-plaus-adult}Average outcomes for the
plausibility measure across key hyperparameters. This shows the \%
change from the baseline model for the distance-based implausibility
metric (\(\text{IP}\)). Boxplots indicate the variation across
evaluation runs and test settings (varying parameters for \emph{ECCCo}).
Data: Adult.}

\end{figure}

\begin{figure}

\centering{

\includegraphics[width=1\linewidth,height=\textheight,keepaspectratio]{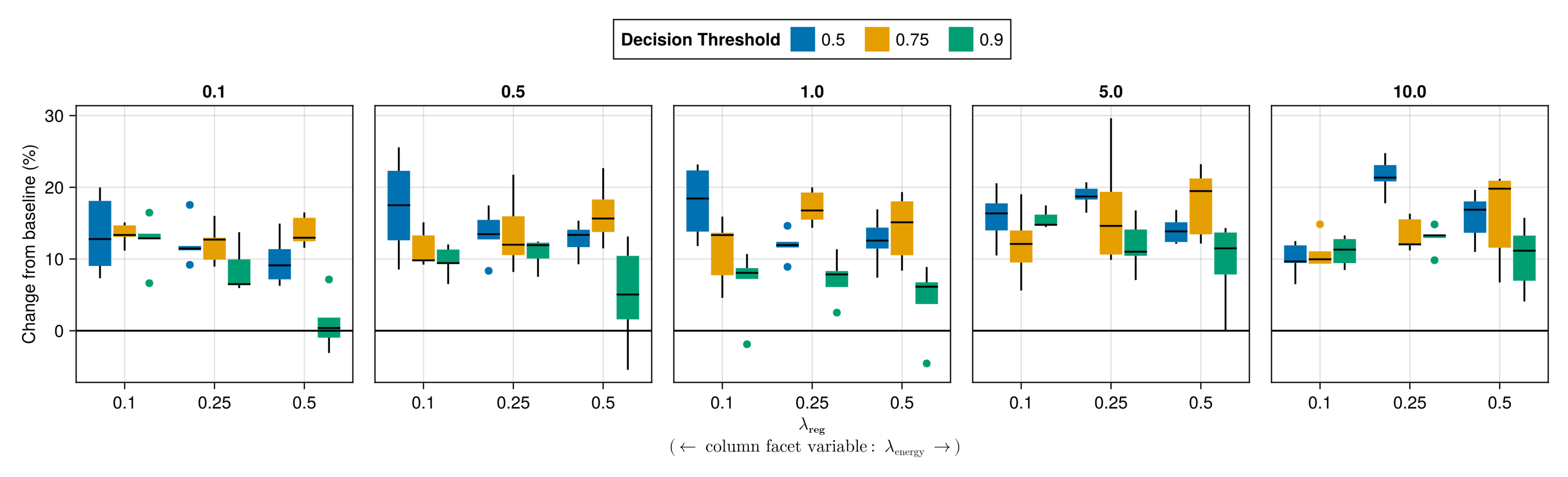}

}

\caption{\label{fig-tune-plaus-cali}Average outcomes for the
plausibility measure across key hyperparameters. This shows the \%
change from the baseline model for the distance-based implausibility
metric (\(\text{IP}\)). Boxplots indicate the variation across
evaluation runs and test settings (varying parameters for \emph{ECCCo}).
Data: California Housing.}

\end{figure}

\begin{figure}

\centering{

\includegraphics[width=1\linewidth,height=\textheight,keepaspectratio]{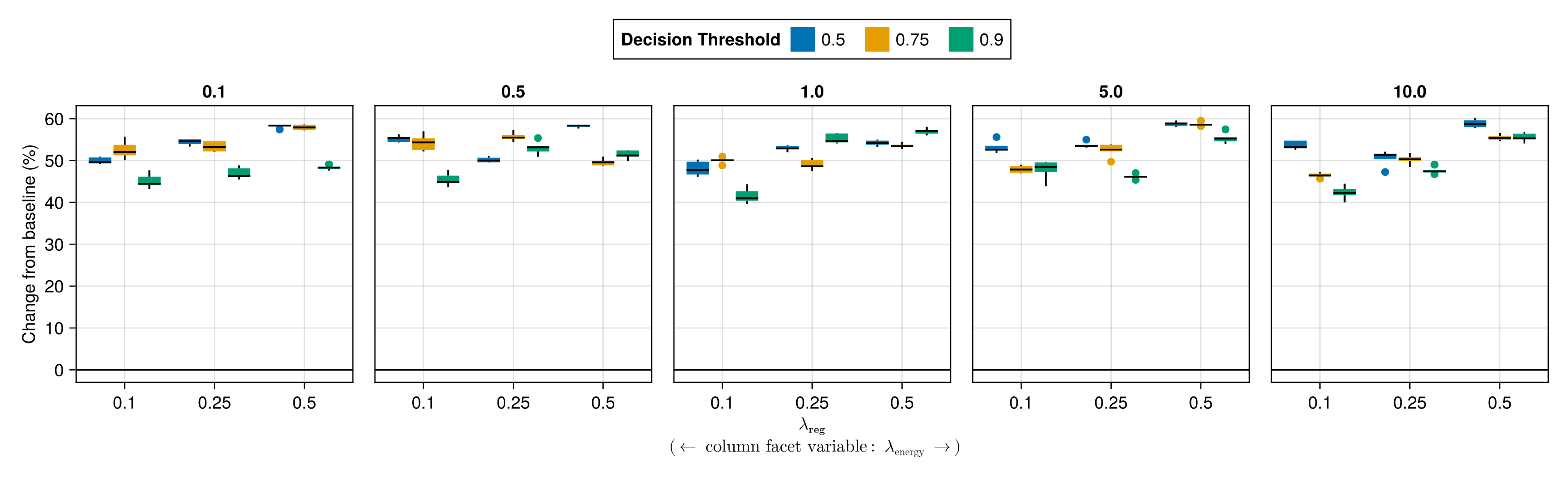}

}

\caption{\label{fig-tune-plaus-circles}Average outcomes for the
plausibility measure across key hyperparameters. This shows the \%
change from the baseline model for the distance-based implausibility
metric (\(\text{IP}\)). Boxplots indicate the variation across
evaluation runs and test settings (varying parameters for \emph{ECCCo}).
Data: Circles.}

\end{figure}

\begin{figure}

\centering{

\includegraphics[width=1\linewidth,height=\textheight,keepaspectratio]{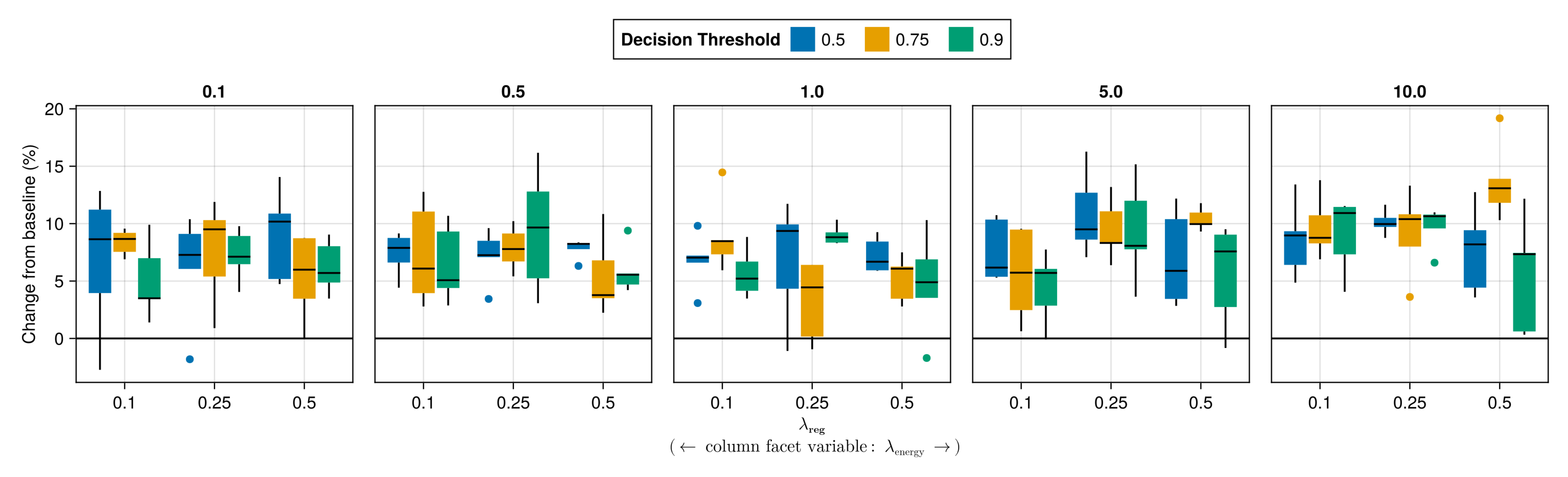}

}

\caption{\label{fig-tune-plaus-credit}Average outcomes for the
plausibility measure across key hyperparameters. This shows the \%
change from the baseline model for the distance-based implausibility
metric (\(\text{IP}\)). Boxplots indicate the variation across
evaluation runs and test settings (varying parameters for \emph{ECCCo}).
Data: Credit.}

\end{figure}

\begin{figure}

\centering{

\includegraphics[width=1\linewidth,height=\textheight,keepaspectratio]{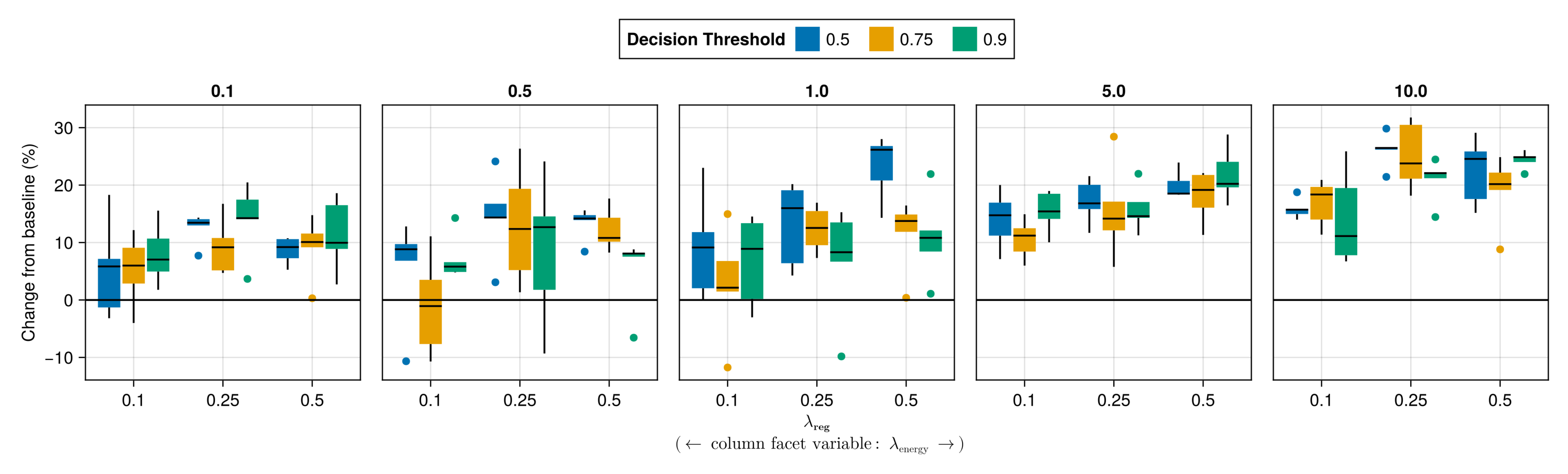}

}

\caption{\label{fig-tune-plaus-gmsc}Average outcomes for the
plausibility measure across key hyperparameters. This shows the \%
change from the baseline model for the distance-based implausibility
metric (\(\text{IP}\)). Boxplots indicate the variation across
evaluation runs and test settings (varying parameters for \emph{ECCCo}).
Data: GMSC.}

\end{figure}

\begin{figure}

\centering{

\includegraphics[width=1\linewidth,height=\textheight,keepaspectratio]{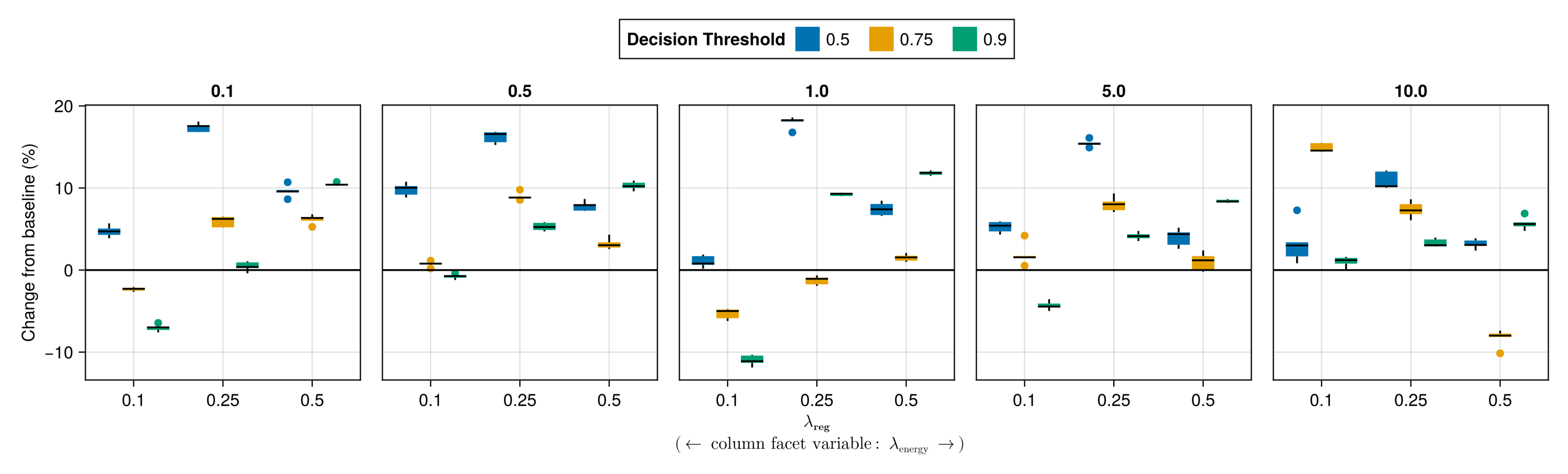}

}

\caption{\label{fig-tune-plaus-lin_sep}Average outcomes for the
plausibility measure across key hyperparameters. This shows the \%
change from the baseline model for the distance-based implausibility
metric (\(\text{IP}\)). Boxplots indicate the variation across
evaluation runs and test settings (varying parameters for \emph{ECCCo}).
Data: Linearly Separable.}

\end{figure}

\begin{figure}

\centering{

\includegraphics[width=1\linewidth,height=\textheight,keepaspectratio]{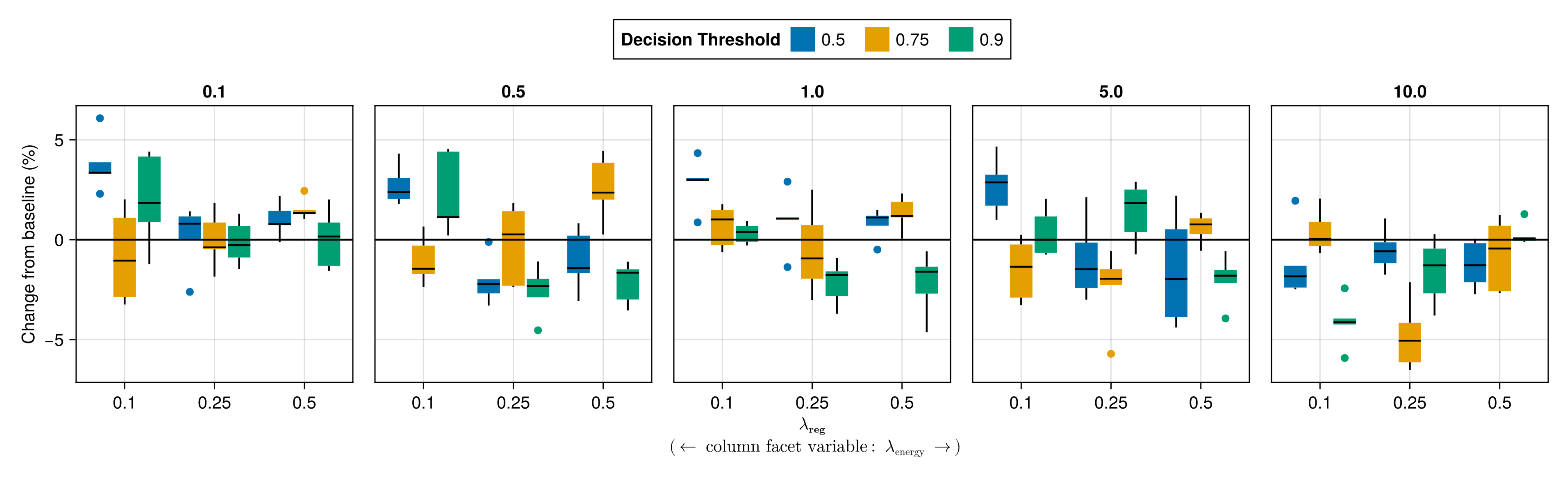}

}

\caption{\label{fig-tune-plaus-mnist}Average outcomes for the
plausibility measure across key hyperparameters. This shows the \%
change from the baseline model for the distance-based implausibility
metric (\(\text{IP}\)). Boxplots indicate the variation across
evaluation runs and test settings (varying parameters for \emph{ECCCo}).
Data: MNIST.}

\end{figure}

\begin{figure}

\centering{

\includegraphics[width=1\linewidth,height=\textheight,keepaspectratio]{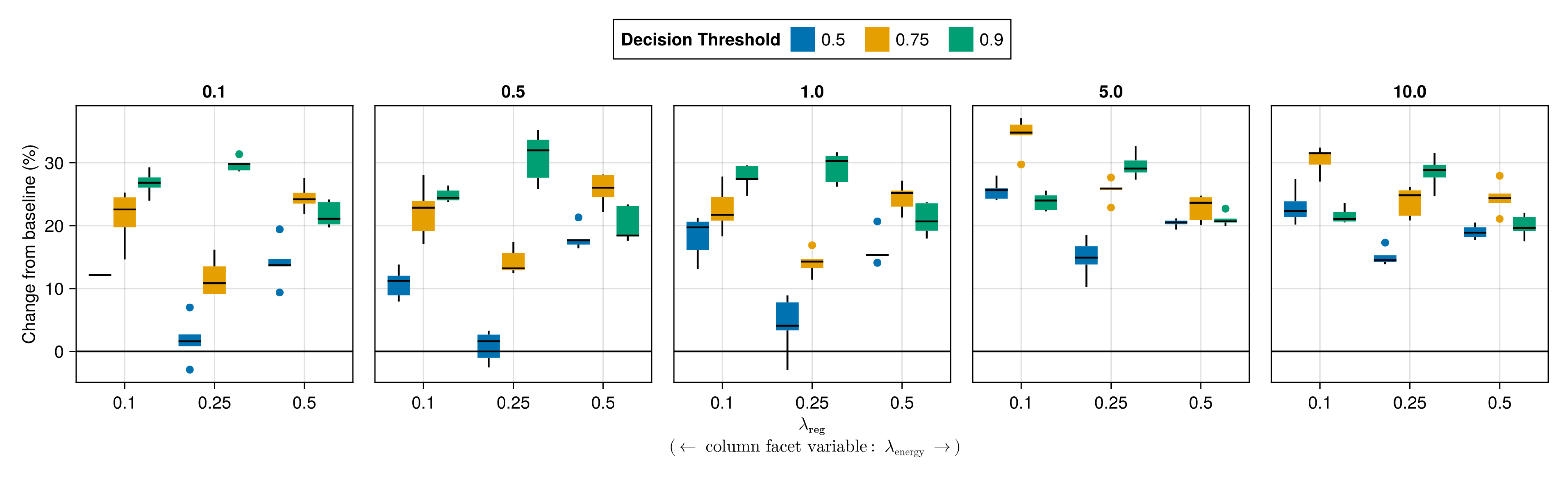}

}

\caption{\label{fig-tune-plaus-moons}Average outcomes for the
plausibility measure across key hyperparameters. This shows the \%
change from the baseline model for the distance-based implausibility
metric (\(\text{IP}\)). Boxplots indicate the variation across
evaluation runs and test settings (varying parameters for \emph{ECCCo}).
Data: Moons.}

\end{figure}

\begin{figure}

\centering{

\includegraphics[width=1\linewidth,height=\textheight,keepaspectratio]{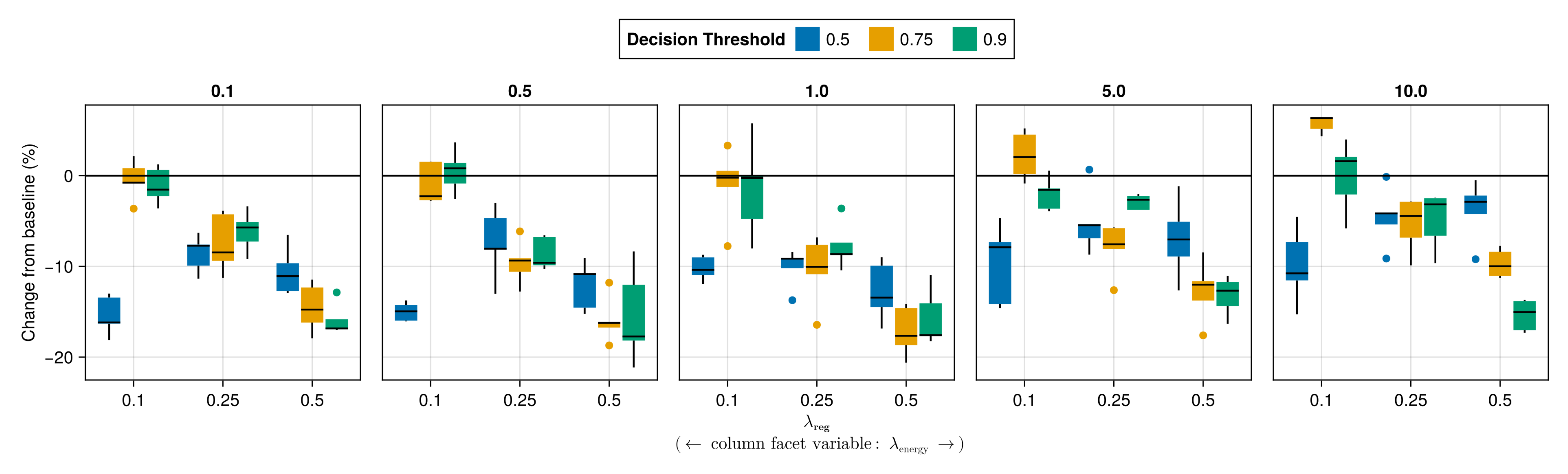}

}

\caption{\label{fig-tune-plaus-over}Average outcomes for the
plausibility measure across key hyperparameters. This shows the \%
change from the baseline model for the distance-based implausibility
metric (\(\text{IP}\)). Boxplots indicate the variation across
evaluation runs and test settings (varying parameters for \emph{ECCCo}).
Data: Overlapping.}

\end{figure}

The results with respect to the plausibility measure are shown in
Figure~\ref{fig-tune-plaus-adult} to Figure~\ref{fig-tune-plaus-over}.

\subsubsection{Proportion of Mature CE}\label{proportion-of-mature-ce}

\begin{figure}

\centering{

\includegraphics[width=1\linewidth,height=\textheight,keepaspectratio]{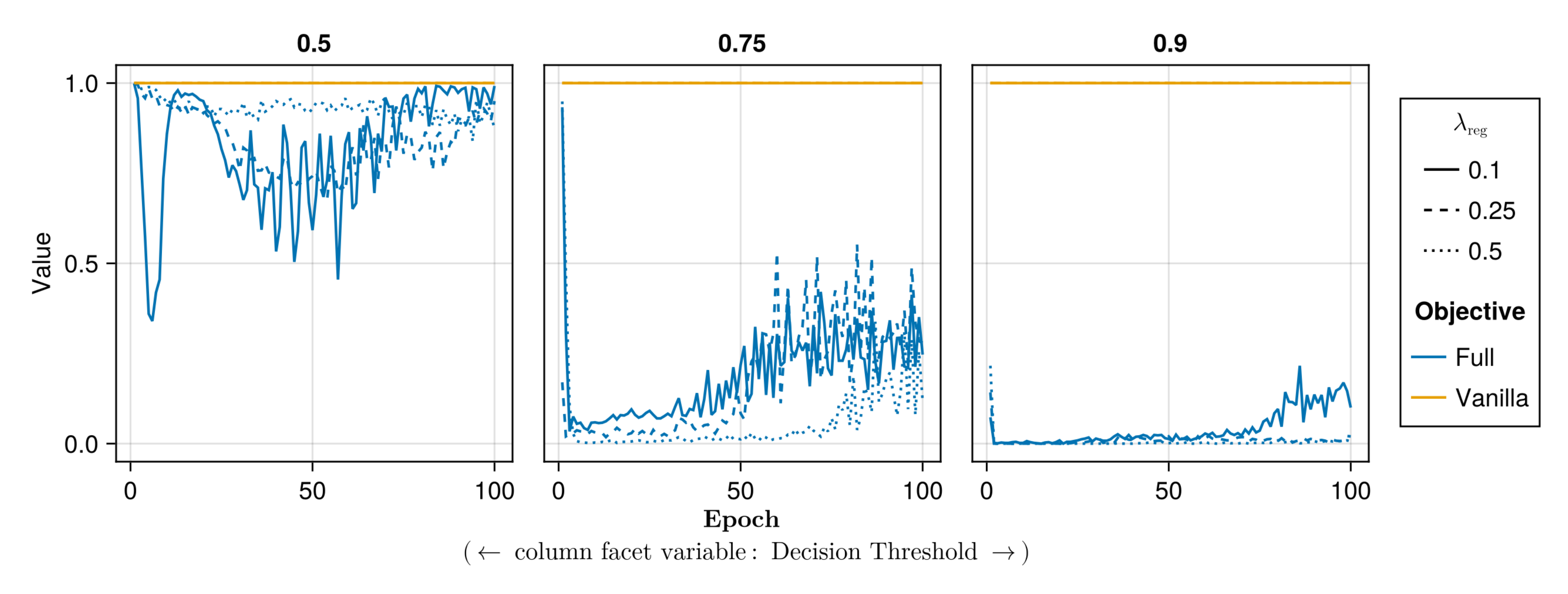}

}

\caption{\label{fig-tune-mat-adult}Proportion of mature counterfactuals
in each epoch. Data: Adult.}

\end{figure}

\begin{figure}

\centering{

\includegraphics[width=1\linewidth,height=\textheight,keepaspectratio]{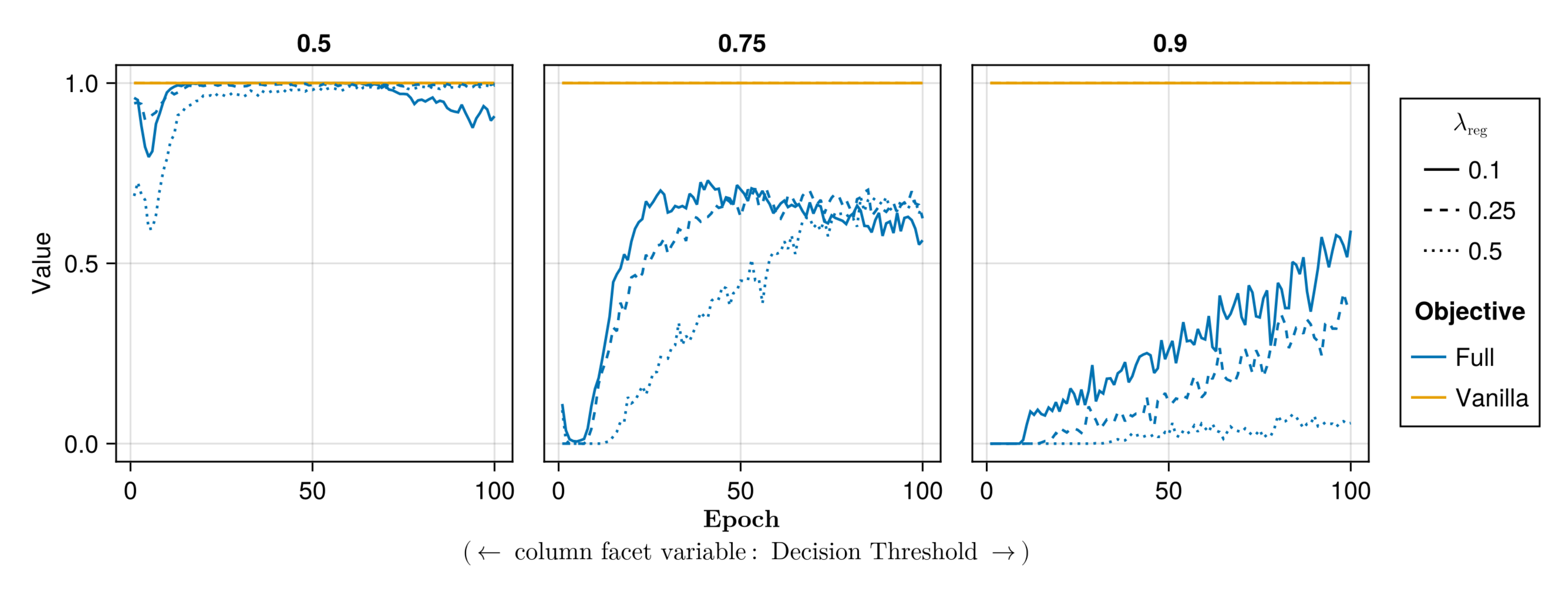}

}

\caption{\label{fig-tune-mat-cali}Proportion of mature counterfactuals
in each epoch. Data: California Housing.}

\end{figure}

\begin{figure}

\centering{

\includegraphics[width=1\linewidth,height=\textheight,keepaspectratio]{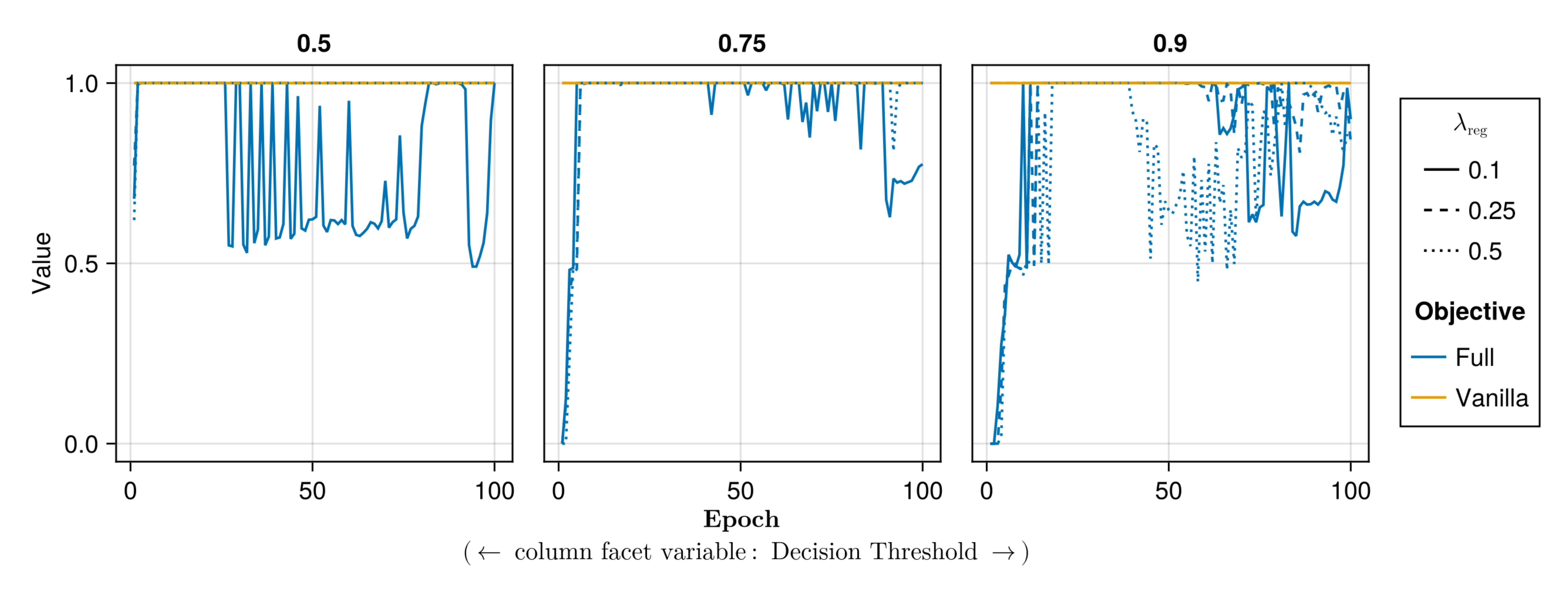}

}

\caption{\label{fig-tune-mat-circles}Proportion of mature
counterfactuals in each epoch. Data: Circles.}

\end{figure}

\begin{figure}

\centering{

\includegraphics[width=1\linewidth,height=\textheight,keepaspectratio]{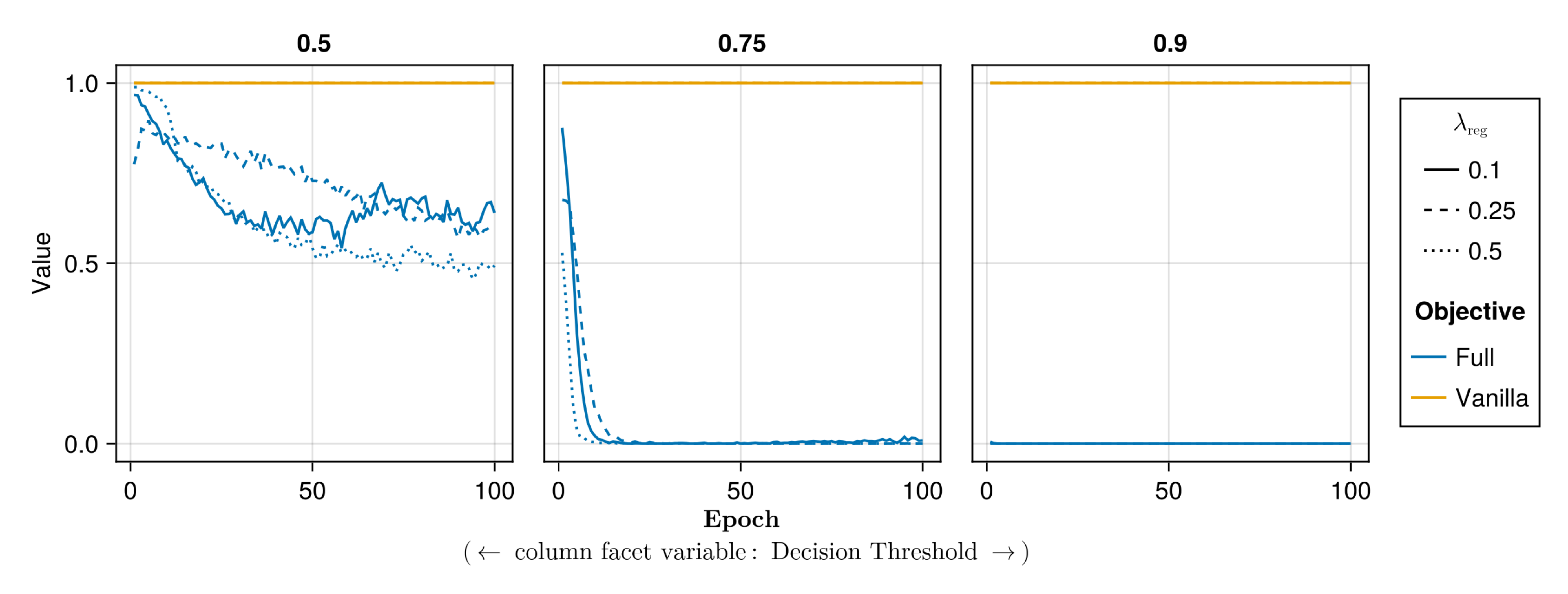}

}

\caption{\label{fig-tune-mat-credit}Proportion of mature counterfactuals
in each epoch. Data: Credit.}

\end{figure}

\begin{figure}

\centering{

\includegraphics[width=1\linewidth,height=\textheight,keepaspectratio]{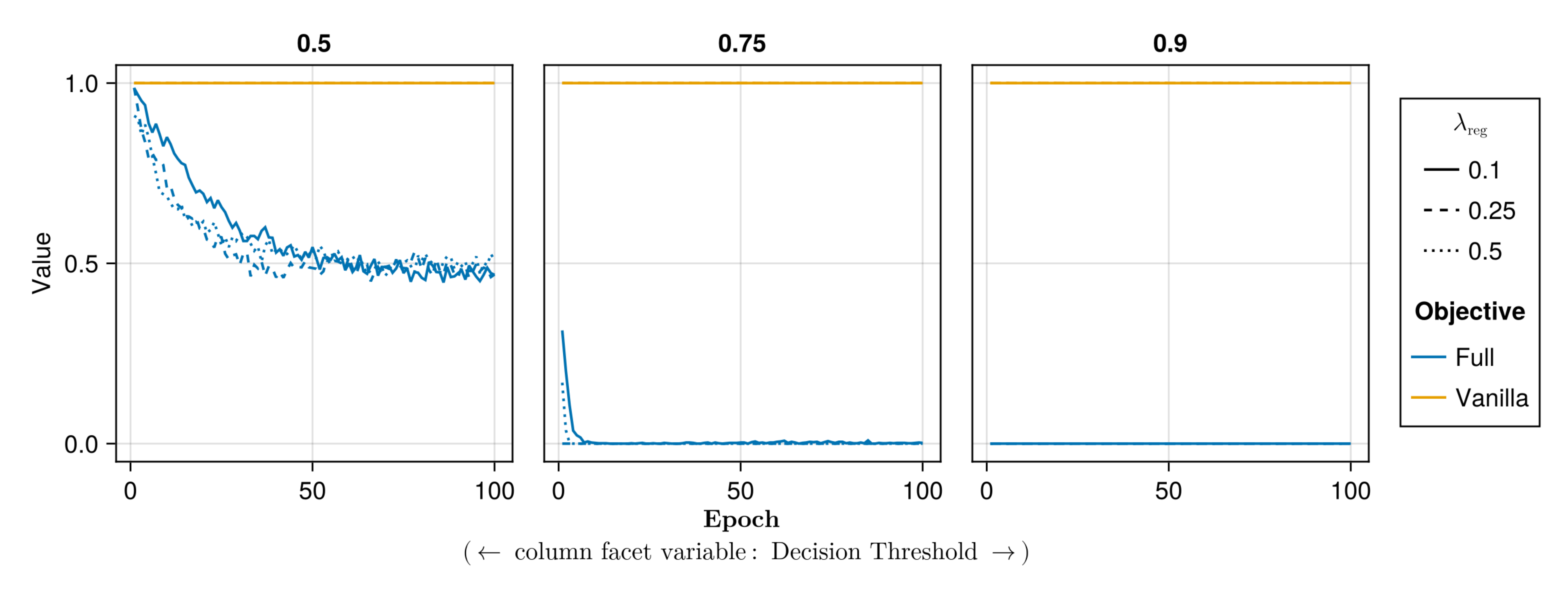}

}

\caption{\label{fig-tune-mat-gmsc}Proportion of mature counterfactuals
in each epoch. Data: GMSC.}

\end{figure}

\begin{figure}

\centering{

\includegraphics[width=1\linewidth,height=\textheight,keepaspectratio]{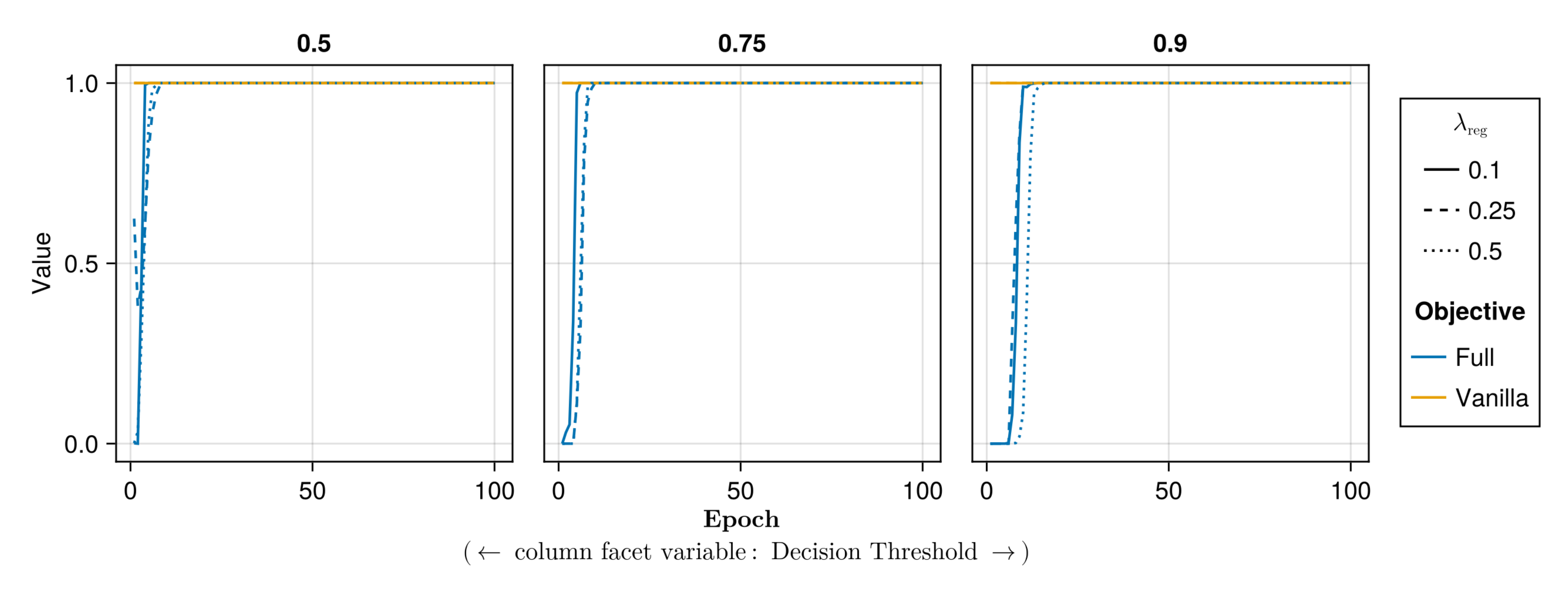}

}

\caption{\label{fig-tune-mat-lin_sep}Proportion of mature
counterfactuals in each epoch. Data: Linearly Separable.}

\end{figure}

\begin{figure}

\centering{

\includegraphics[width=1\linewidth,height=\textheight,keepaspectratio]{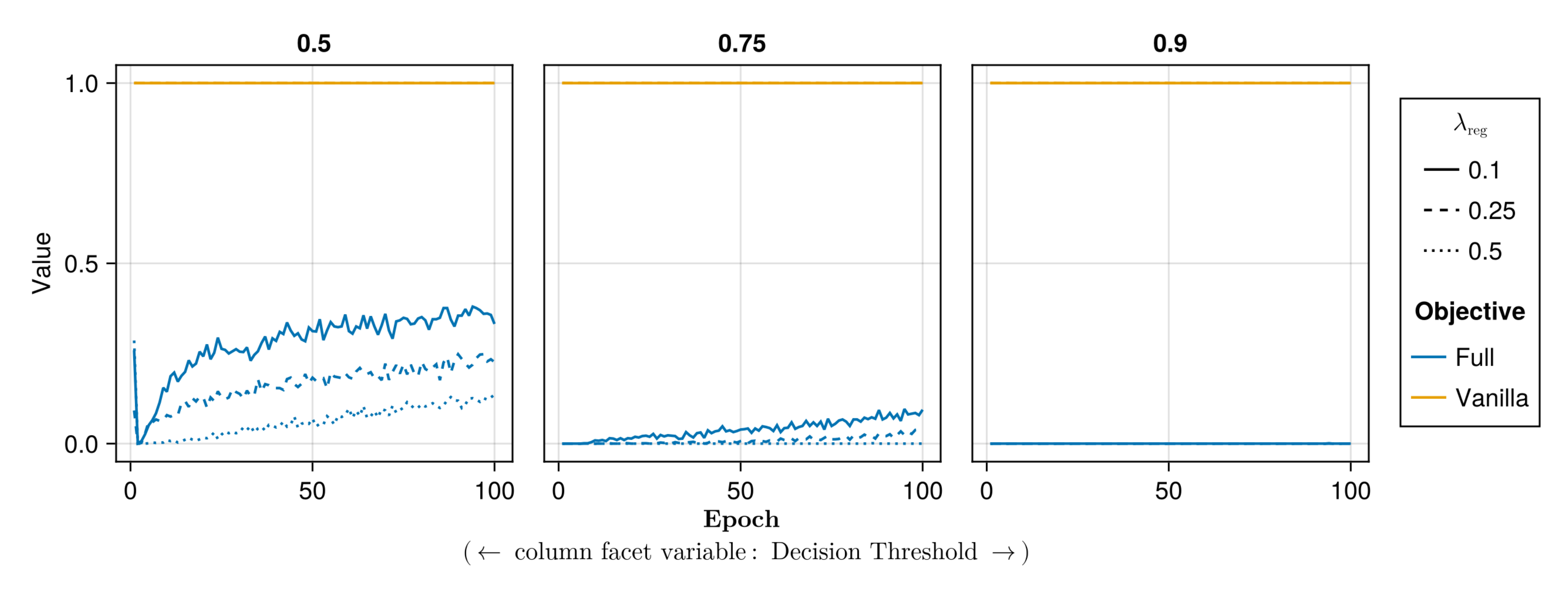}

}

\caption{\label{fig-tune-mat-mnist}Proportion of mature counterfactuals
in each epoch. Data: MNIST.}

\end{figure}

\begin{figure}

\centering{

\includegraphics[width=1\linewidth,height=\textheight,keepaspectratio]{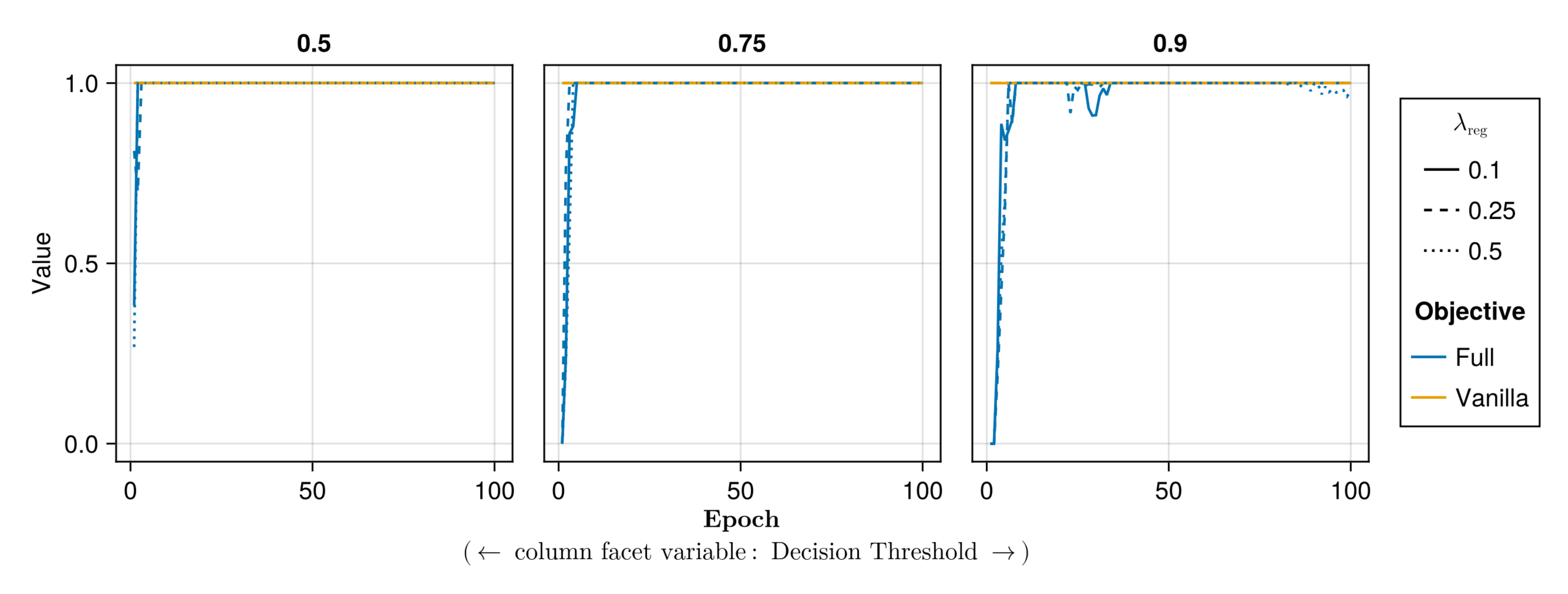}

}

\caption{\label{fig-tune-mat-moons}Proportion of mature counterfactuals
in each epoch. Data: Moons.}

\end{figure}

\begin{figure}

\centering{

\includegraphics[width=1\linewidth,height=\textheight,keepaspectratio]{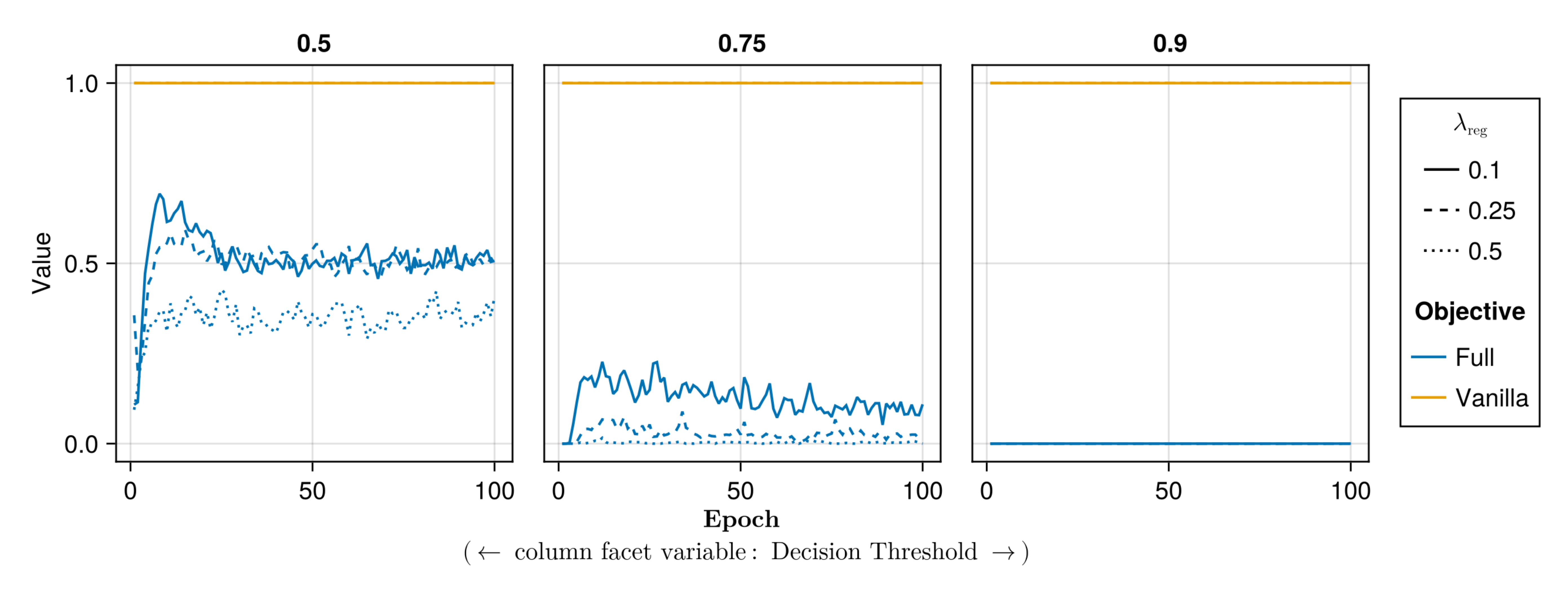}

}

\caption{\label{fig-tune-mat-over}Proportion of mature counterfactuals
in each epoch. Data: Overlapping.}

\end{figure}

The results with respect to the proportion of mature counterfactuals in
each epoch are shown in Figure~\ref{fig-tune-mat-adult} to
Figure~\ref{fig-tune-mat-over}.

\FloatBarrier

\subsection{Learning Rate}\label{sec-app-tune-lr}

The hyperparameter grid for tuning the learning rate is shown in
Note~\ref{nte-tune_lr-train}. The corresponding evaluation grid used for
these experiments is shown in Note~\ref{nte-tune_lr-eval}.

\begin{tcolorbox}[enhanced jigsaw, arc=.35mm, colback=white, bottomrule=.15mm, toptitle=1mm, leftrule=.75mm, titlerule=0mm, toprule=.15mm, breakable, colframe=quarto-callout-note-color-frame, left=2mm, opacityback=0, bottomtitle=1mm, opacitybacktitle=0.6, title={Note \ref*{nte-tune_lr-train}space Training Phase}, colbacktitle=quarto-callout-note-color!10!white, rightrule=.15mm, coltitle=black]

\quartocalloutnte{nte-tune_lr-train} 

\begin{itemize}
\tightlist
\item
  Generator Parameters:

  \begin{itemize}
  \tightlist
  \item
    Learning Rate: \texttt{0.1,\ 0.5,\ 1.0}
  \end{itemize}
\item
  Model: \texttt{mlp}
\item
  Training Parameters:

  \begin{itemize}
  \tightlist
  \item
    \(\lambda_{\text{reg}}\): \texttt{0.01,\ 0.1,\ 0.5}
  \item
    Objective: \texttt{full,\ vanilla}
  \end{itemize}
\end{itemize}

\end{tcolorbox}

\begin{tcolorbox}[enhanced jigsaw, arc=.35mm, colback=white, bottomrule=.15mm, toptitle=1mm, leftrule=.75mm, titlerule=0mm, toprule=.15mm, breakable, colframe=quarto-callout-note-color-frame, left=2mm, opacityback=0, bottomtitle=1mm, opacitybacktitle=0.6, title={Note \ref*{nte-tune_lr-eval}space Evaluation Phase}, colbacktitle=quarto-callout-note-color!10!white, rightrule=.15mm, coltitle=black]

\quartocalloutnte{nte-tune_lr-eval} 

\begin{itemize}
\tightlist
\item
  Generator Parameters:

  \begin{itemize}
  \tightlist
  \item
    \(\lambda_{\text{egy}}\): \texttt{0.1,\ 0.5,\ 1.0,\ 5.0,\ 10.0}
  \end{itemize}
\end{itemize}

\end{tcolorbox}

\subsubsection{Plausibility}\label{plausibility-4}

\begin{figure}

\centering{

\includegraphics[width=1\linewidth,height=\textheight,keepaspectratio]{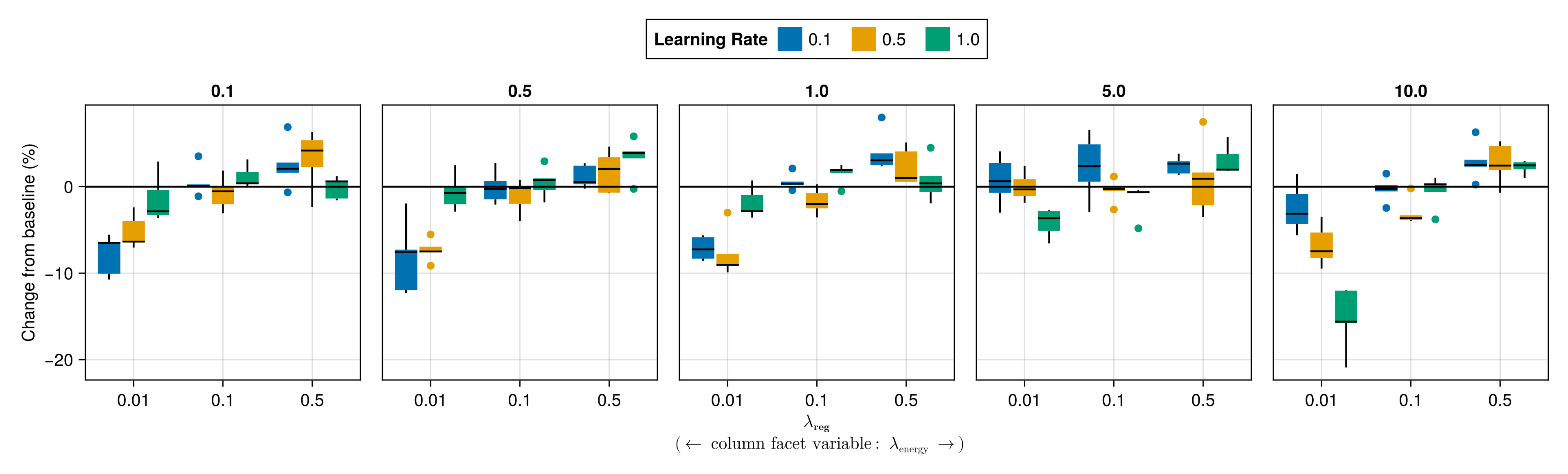}

}

\caption{\label{fig-tune_lr-plaus-adult}Average outcomes for the
plausibility measure across key hyperparameters. This shows the \%
change from the baseline model for the distance-based implausibility
metric (\(\text{IP}\)). Boxplots indicate the variation across
evaluation runs and test settings (varying parameters for \emph{ECCCo}).
Data: Adult.}

\end{figure}

\begin{figure}

\centering{

\includegraphics[width=1\linewidth,height=\textheight,keepaspectratio]{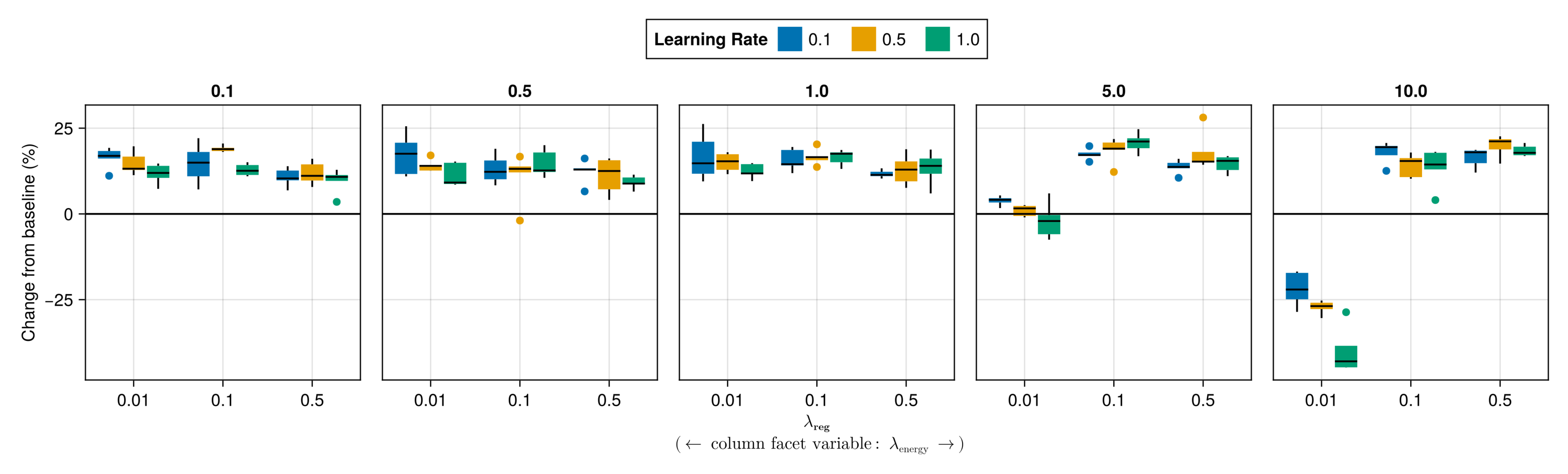}

}

\caption{\label{fig-tune_lr-plaus-cali}Average outcomes for the
plausibility measure across key hyperparameters. This shows the \%
change from the baseline model for the distance-based implausibility
metric (\(\text{IP}\)). Boxplots indicate the variation across
evaluation runs and test settings (varying parameters for \emph{ECCCo}).
Data: California Housing.}

\end{figure}

\begin{figure}

\centering{

\includegraphics[width=1\linewidth,height=\textheight,keepaspectratio]{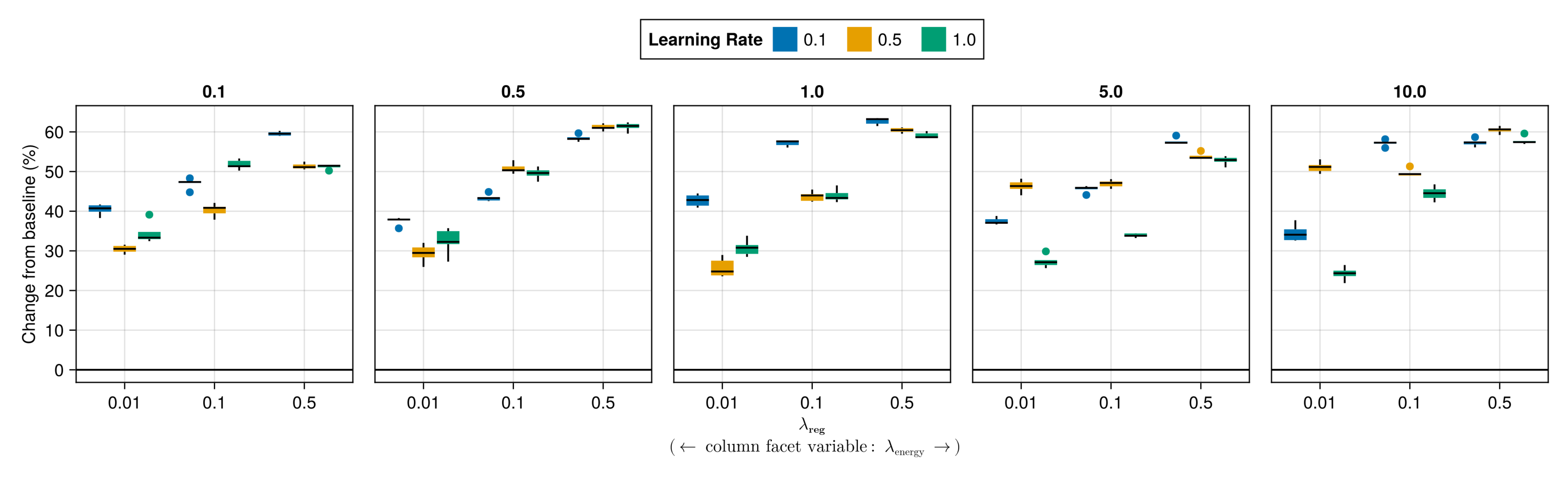}

}

\caption{\label{fig-tune_lr-plaus-circles}Average outcomes for the
plausibility measure across key hyperparameters. This shows the \%
change from the baseline model for the distance-based implausibility
metric (\(\text{IP}\)). Boxplots indicate the variation across
evaluation runs and test settings (varying parameters for \emph{ECCCo}).
Data: Circles.}

\end{figure}

\begin{figure}

\centering{

\includegraphics[width=1\linewidth,height=\textheight,keepaspectratio]{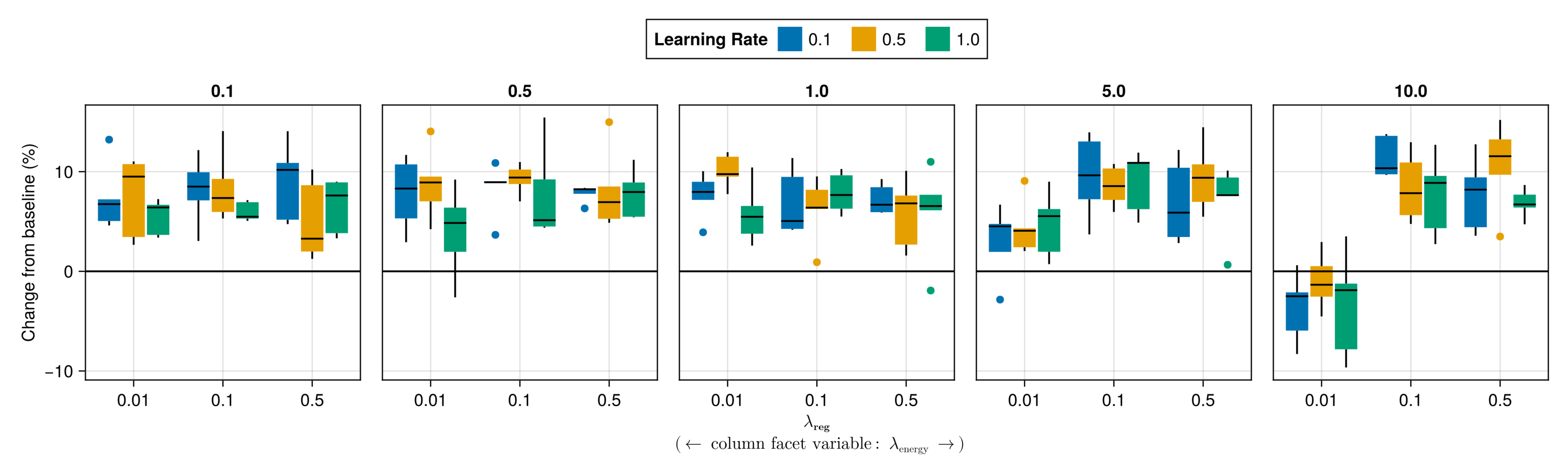}

}

\caption{\label{fig-tune_lr-plaus-credit}Average outcomes for the
plausibility measure across key hyperparameters. This shows the \%
change from the baseline model for the distance-based implausibility
metric (\(\text{IP}\)). Boxplots indicate the variation across
evaluation runs and test settings (varying parameters for \emph{ECCCo}).
Data: Credit.}

\end{figure}

\begin{figure}

\centering{

\includegraphics[width=1\linewidth,height=\textheight,keepaspectratio]{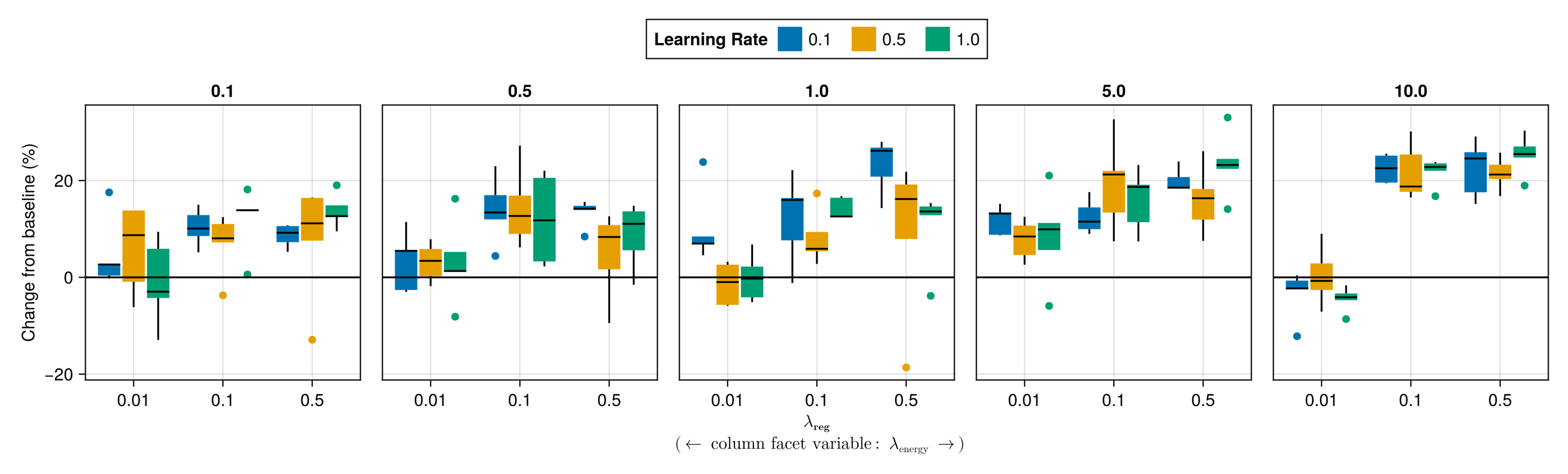}

}

\caption{\label{fig-tune_lr-plaus-gmsc}Average outcomes for the
plausibility measure across key hyperparameters. This shows the \%
change from the baseline model for the distance-based implausibility
metric (\(\text{IP}\)). Boxplots indicate the variation across
evaluation runs and test settings (varying parameters for \emph{ECCCo}).
Data: GMSC.}

\end{figure}

\begin{figure}

\centering{

\includegraphics[width=1\linewidth,height=\textheight,keepaspectratio]{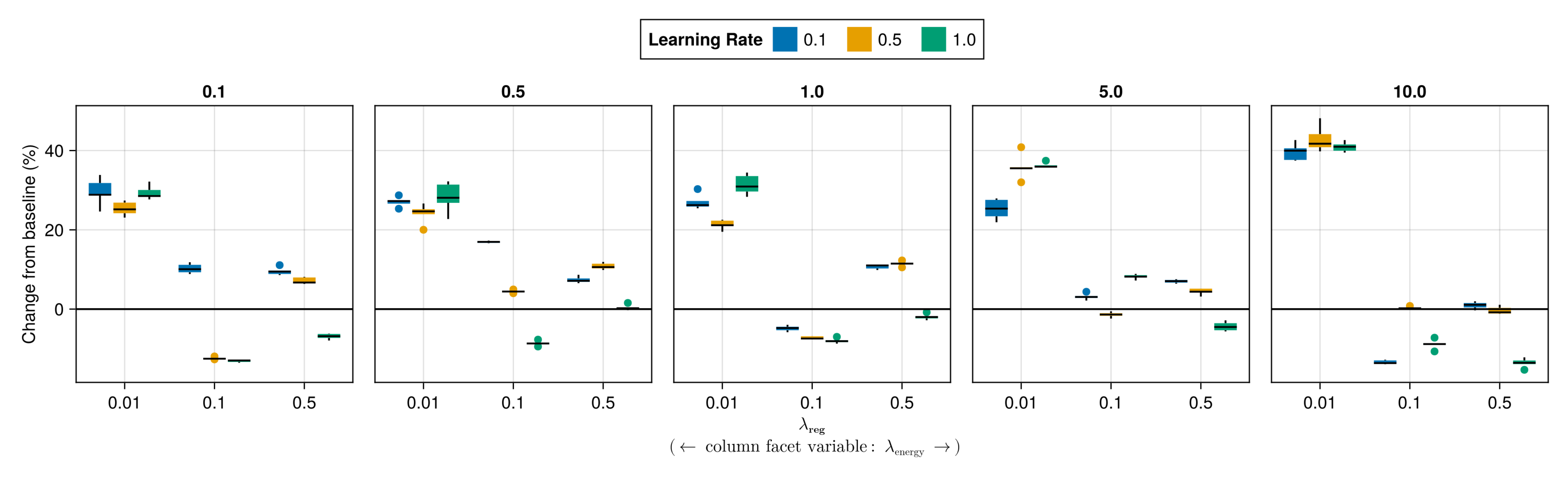}

}

\caption{\label{fig-tune_lr-plaus-lin_sep}Average outcomes for the
plausibility measure across key hyperparameters. This shows the \%
change from the baseline model for the distance-based implausibility
metric (\(\text{IP}\)). Boxplots indicate the variation across
evaluation runs and test settings (varying parameters for \emph{ECCCo}).
Data: Linearly Separable.}

\end{figure}

\begin{figure}

\centering{

\includegraphics[width=1\linewidth,height=\textheight,keepaspectratio]{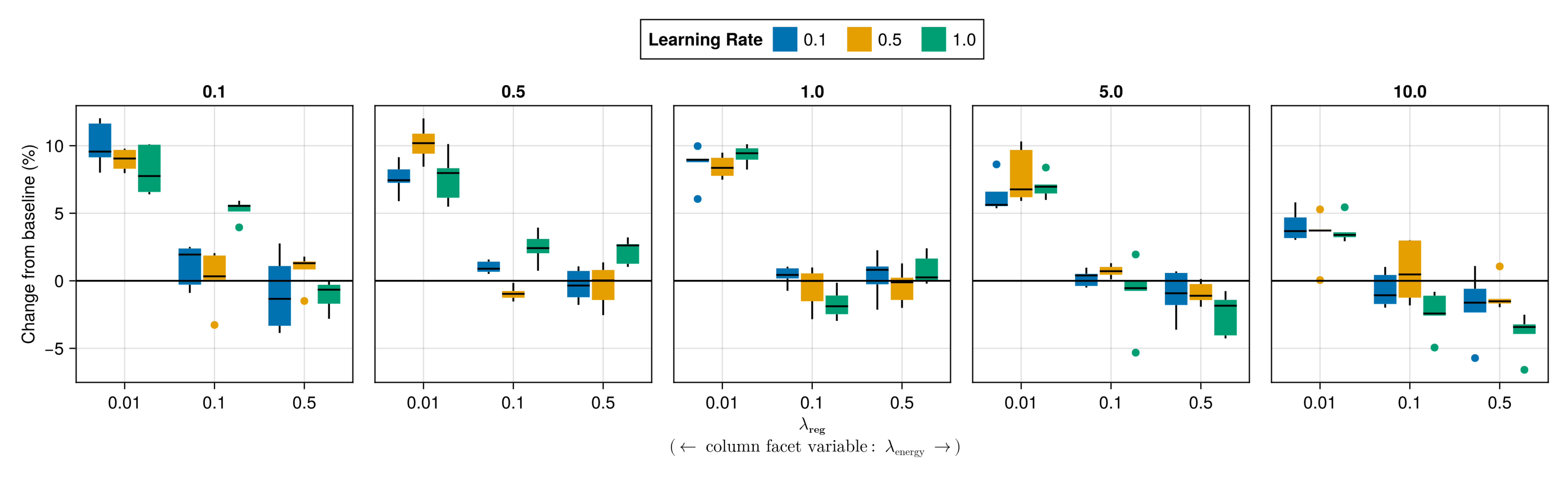}

}

\caption{\label{fig-tune_lr-plaus-mnist}Average outcomes for the
plausibility measure across key hyperparameters. This shows the \%
change from the baseline model for the distance-based implausibility
metric (\(\text{IP}\)). Boxplots indicate the variation across
evaluation runs and test settings (varying parameters for \emph{ECCCo}).
Data: MNIST.}

\end{figure}

\begin{figure}

\centering{

\includegraphics[width=1\linewidth,height=\textheight,keepaspectratio]{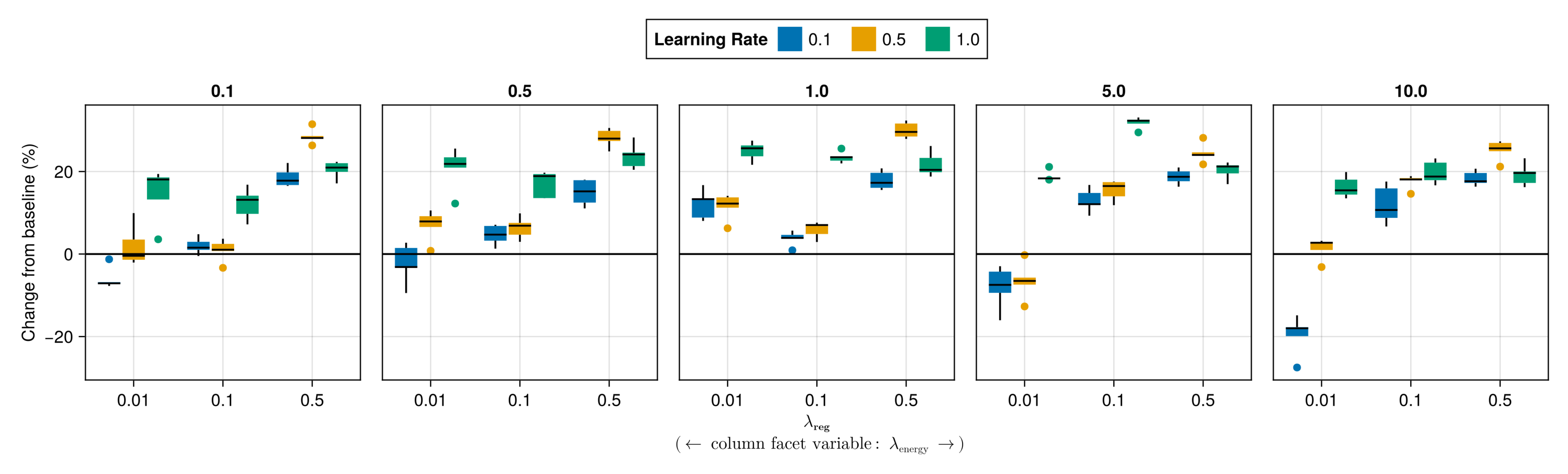}

}

\caption{\label{fig-tune_lr-plaus-moons}Average outcomes for the
plausibility measure across key hyperparameters. This shows the \%
change from the baseline model for the distance-based implausibility
metric (\(\text{IP}\)). Boxplots indicate the variation across
evaluation runs and test settings (varying parameters for \emph{ECCCo}).
Data: Moons.}

\end{figure}

\begin{figure}

\centering{

\includegraphics[width=1\linewidth,height=\textheight,keepaspectratio]{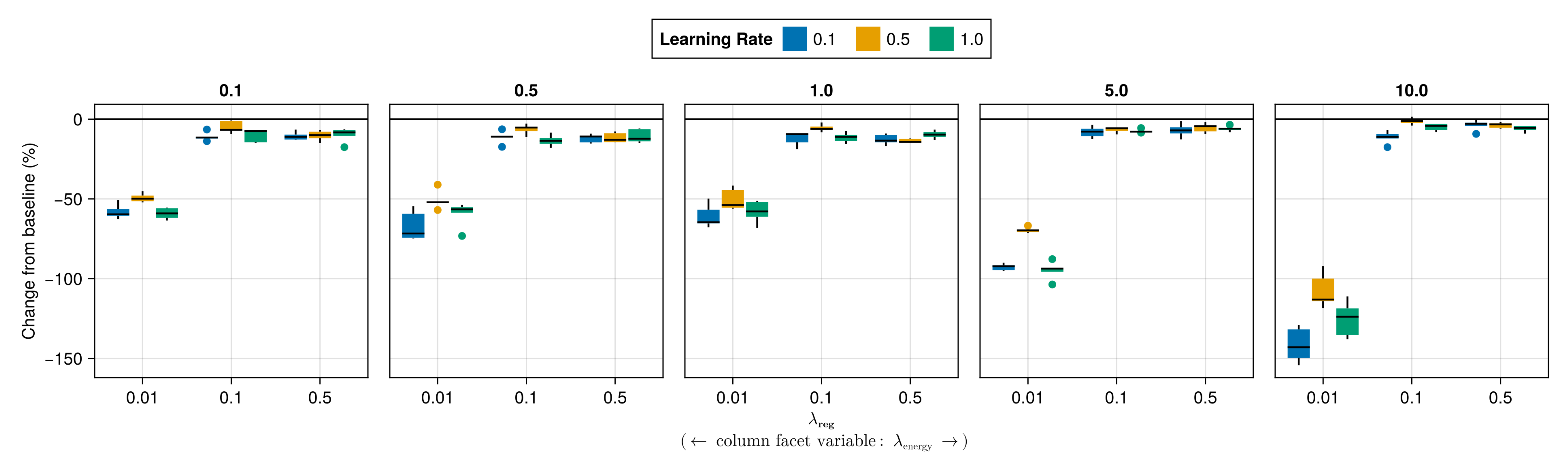}

}

\caption{\label{fig-tune_lr-plaus-over}Average outcomes for the
plausibility measure across key hyperparameters. This shows the \%
change from the baseline model for the distance-based implausibility
metric (\(\text{IP}\)). Boxplots indicate the variation across
evaluation runs and test settings (varying parameters for \emph{ECCCo}).
Data: Overlapping.}

\end{figure}

The results with respect to the plausibility measure are shown in
Figure~\ref{fig-tune_lr-plaus-adult} to
Figure~\ref{fig-tune_lr-plaus-over}.

\subsubsection{Proportion of Mature CE}\label{proportion-of-mature-ce-1}

\begin{figure}

\centering{

\includegraphics[width=1\linewidth,height=\textheight,keepaspectratio]{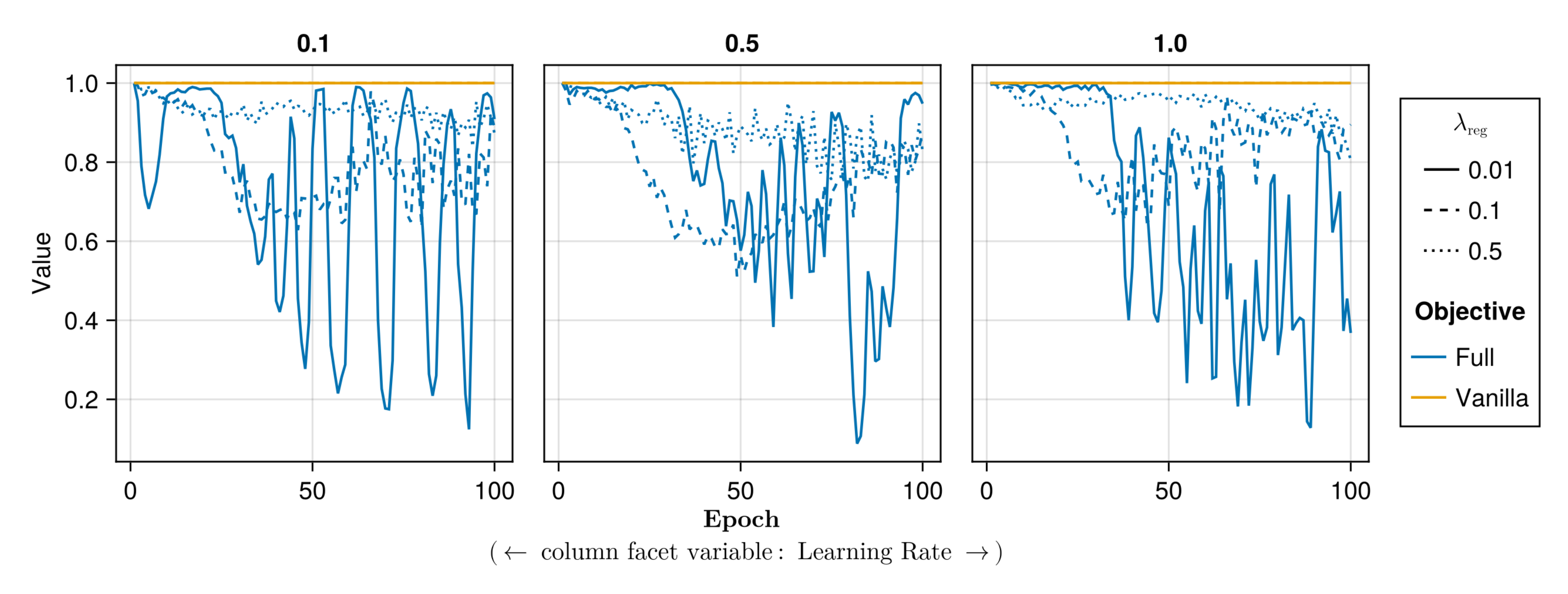}

}

\caption{\label{fig-tune_lr-mat-adult}Proportion of mature
counterfactuals in each epoch. Data: Adult.}

\end{figure}

\begin{figure}

\centering{

\includegraphics[width=1\linewidth,height=\textheight,keepaspectratio]{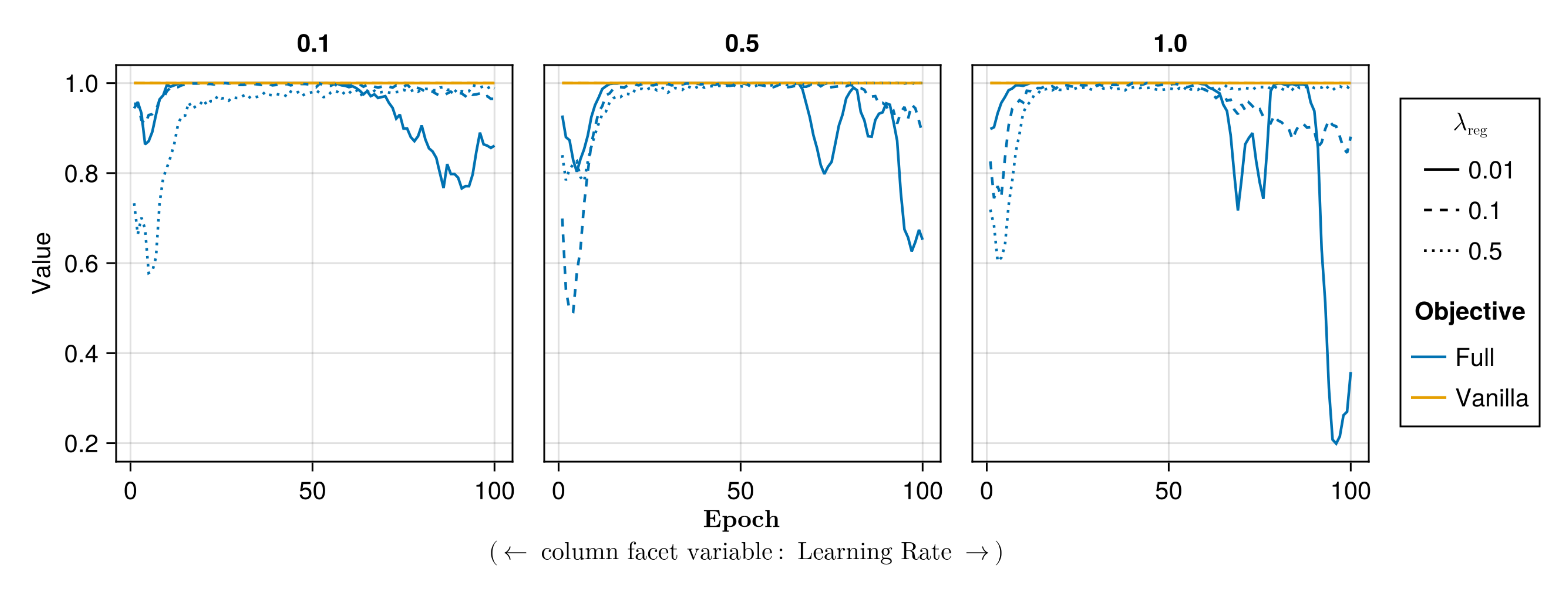}

}

\caption{\label{fig-tune_lr-mat-cali}Proportion of mature
counterfactuals in each epoch. Data: California Housing.}

\end{figure}

\begin{figure}

\centering{

\includegraphics[width=1\linewidth,height=\textheight,keepaspectratio]{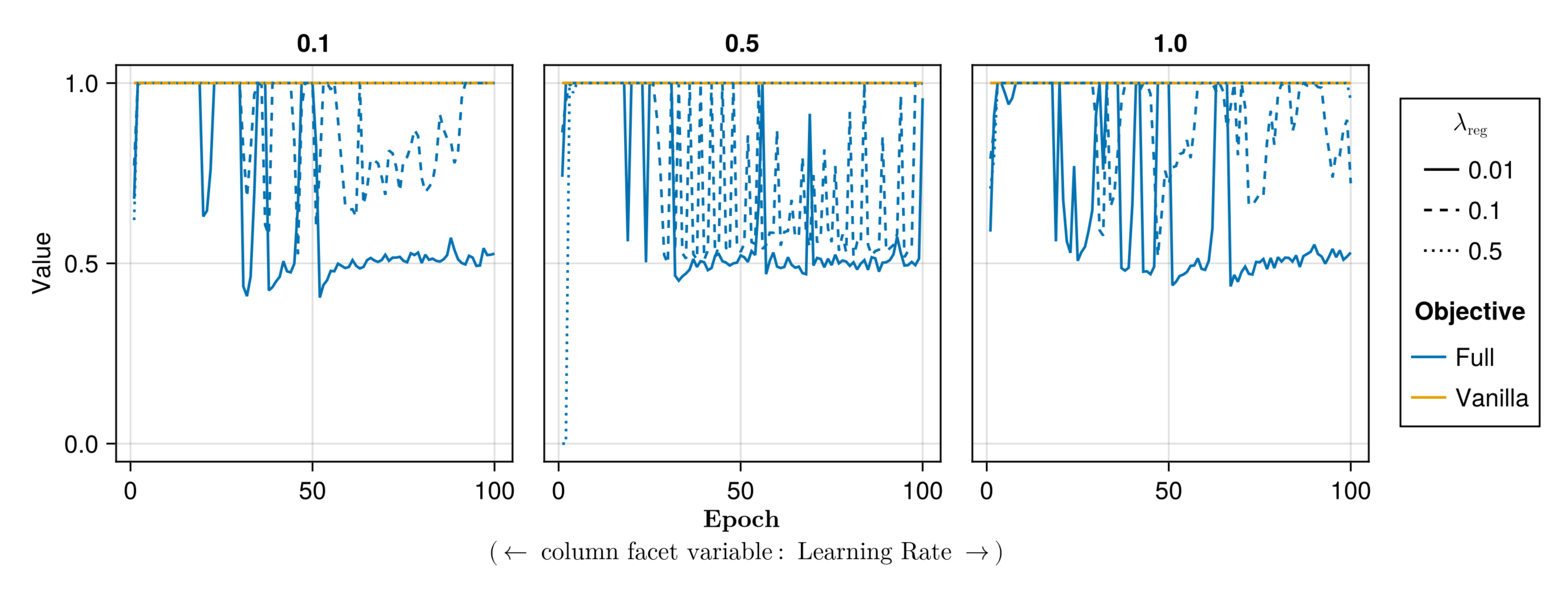}

}

\caption{\label{fig-tune_lr-mat-circles}Proportion of mature
counterfactuals in each epoch. Data: Circles.}

\end{figure}

\begin{figure}

\centering{

\includegraphics[width=1\linewidth,height=\textheight,keepaspectratio]{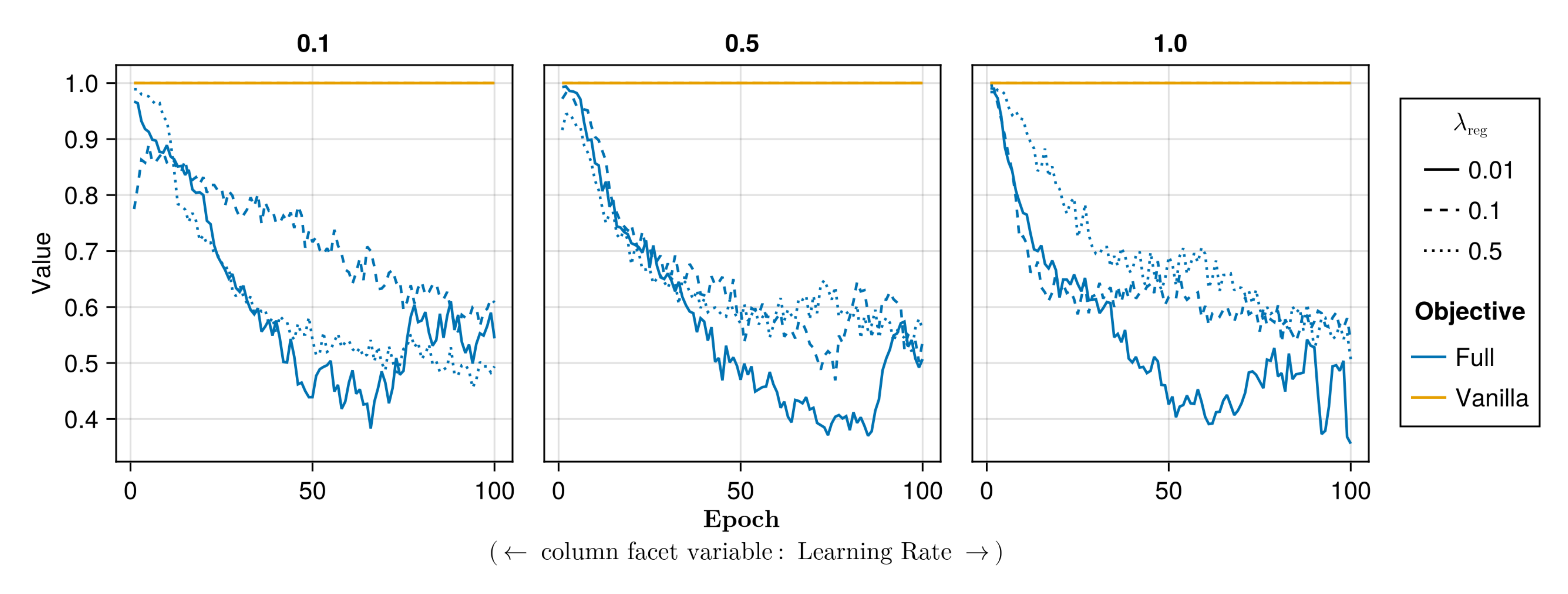}

}

\caption{\label{fig-tune_lr-mat-credit}Proportion of mature
counterfactuals in each epoch. Data: Credit.}

\end{figure}

\begin{figure}

\centering{

\includegraphics[width=1\linewidth,height=\textheight,keepaspectratio]{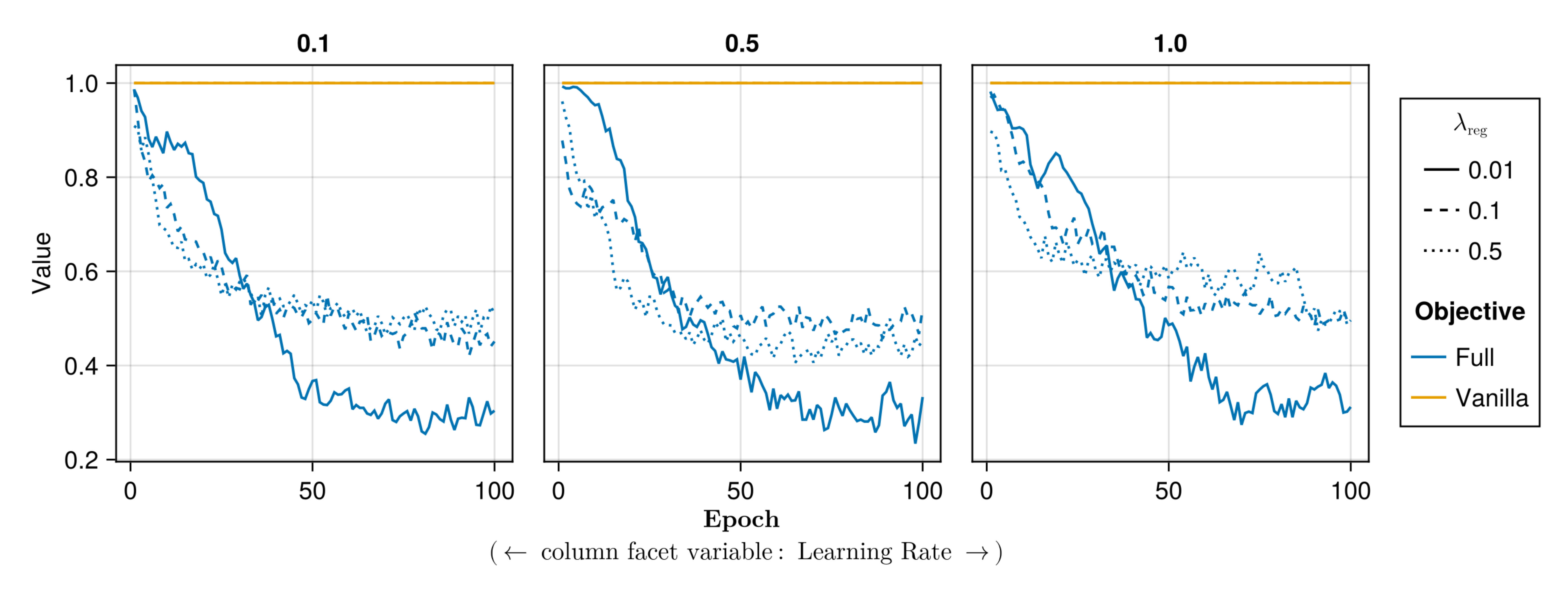}

}

\caption{\label{fig-tune_lr-mat-gmsc}Proportion of mature
counterfactuals in each epoch. Data: GMSC.}

\end{figure}

\begin{figure}

\centering{

\includegraphics[width=1\linewidth,height=\textheight,keepaspectratio]{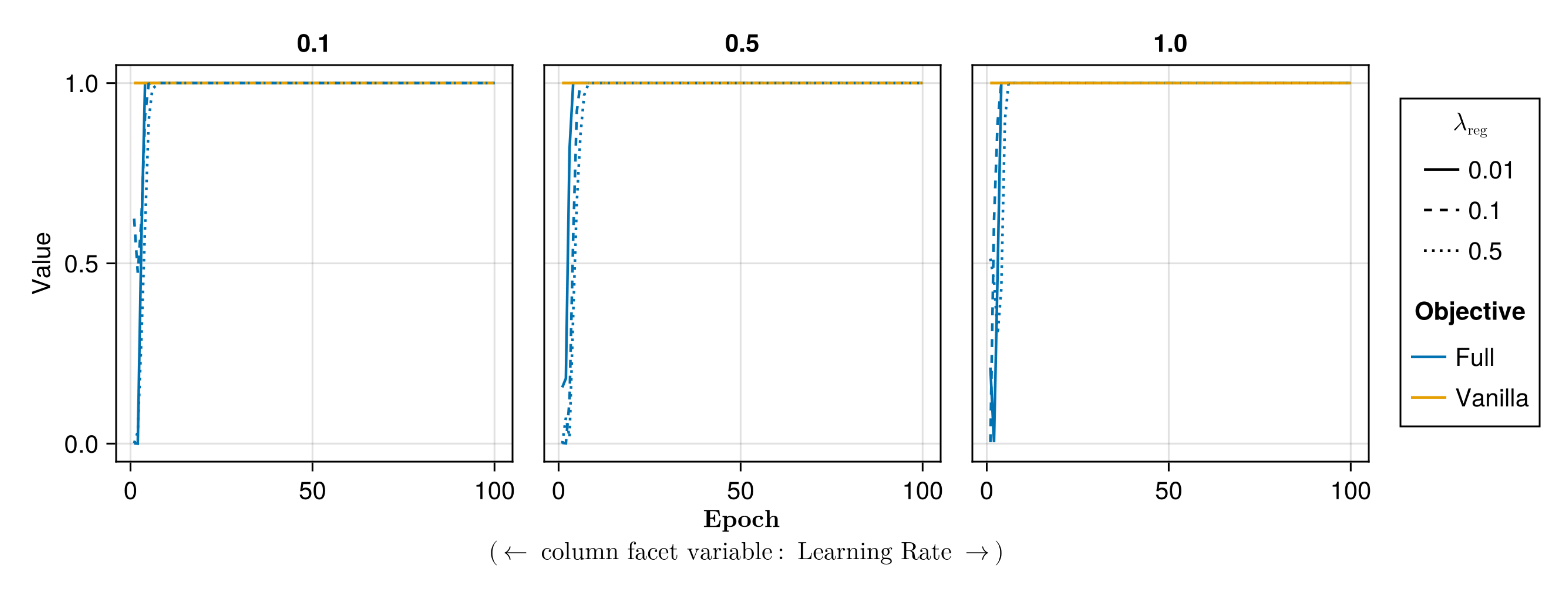}

}

\caption{\label{fig-tune_lr-mat-lin_sep}Proportion of mature
counterfactuals in each epoch. Data: Linearly Separable.}

\end{figure}

\begin{figure}

\centering{

\includegraphics[width=1\linewidth,height=\textheight,keepaspectratio]{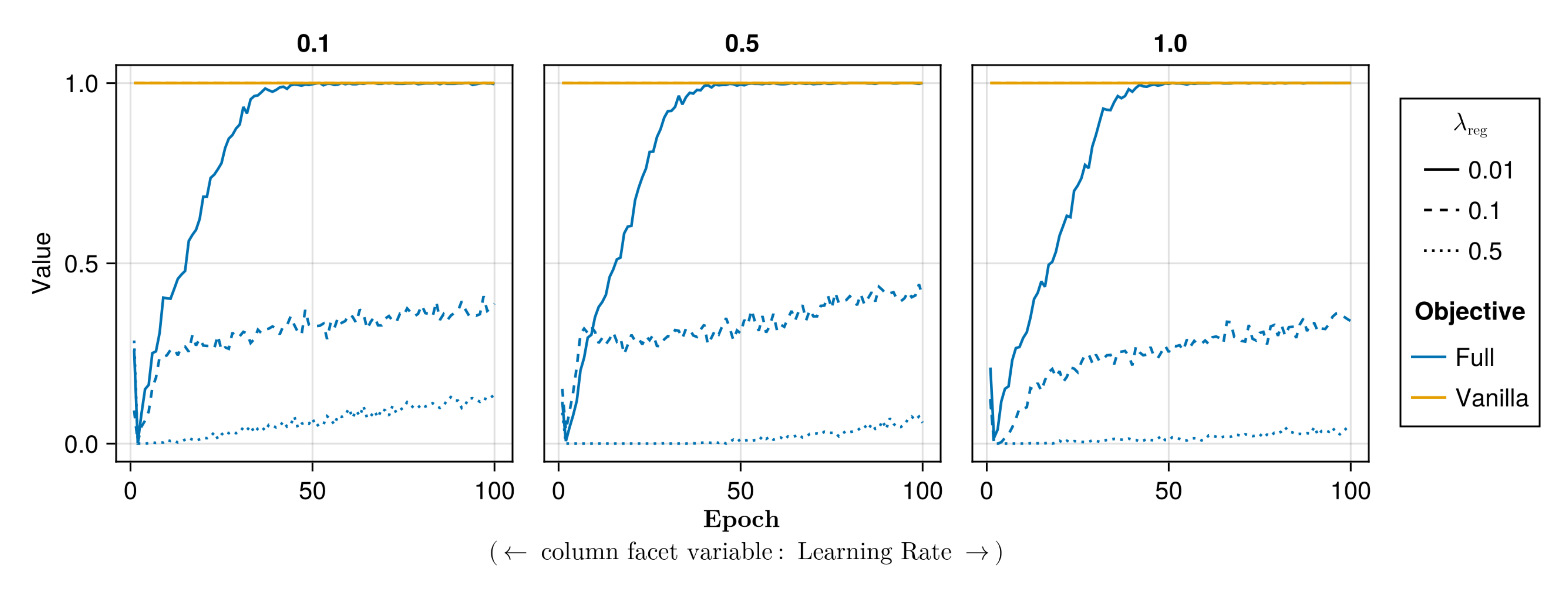}

}

\caption{\label{fig-tune_lr-mat-mnist}Proportion of mature
counterfactuals in each epoch. Data: MNIST.}

\end{figure}

\begin{figure}

\centering{

\includegraphics[width=1\linewidth,height=\textheight,keepaspectratio]{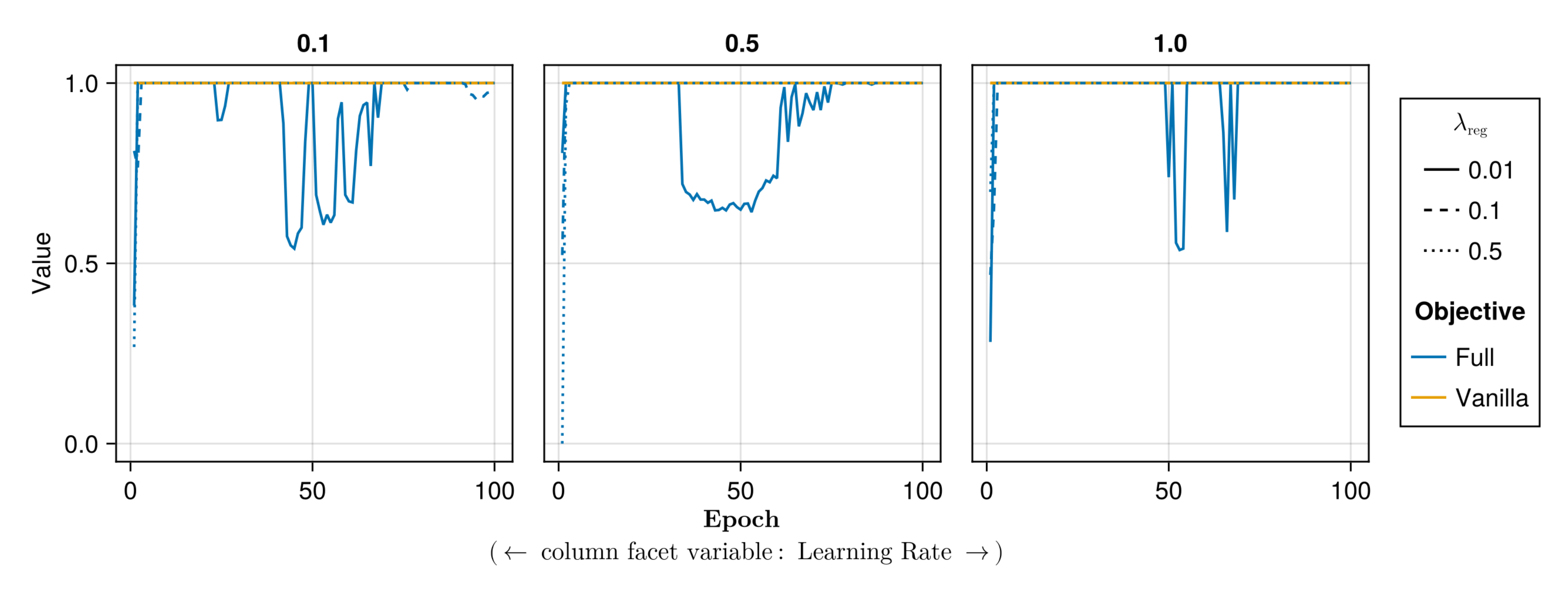}

}

\caption{\label{fig-tune_lr-mat-moons}Proportion of mature
counterfactuals in each epoch. Data: Moons.}

\end{figure}

\begin{figure}

\centering{

\includegraphics[width=1\linewidth,height=\textheight,keepaspectratio]{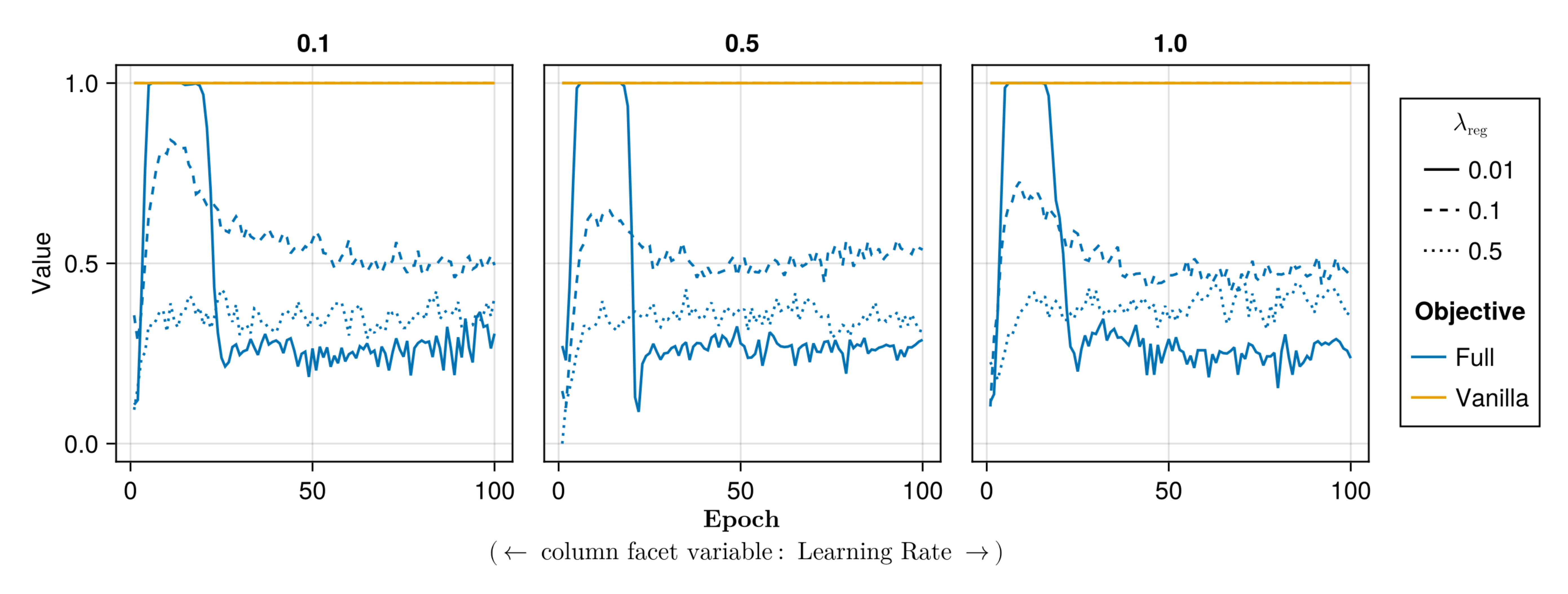}

}

\caption{\label{fig-tune_lr-mat-over}Proportion of mature
counterfactuals in each epoch. Data: Overlapping.}

\end{figure}

The results with respect to the proportion of mature counterfactuals in
each epoch are shown in Figure~\ref{fig-tune_lr-mat-adult} to
Figure~\ref{fig-tune_lr-mat-over}.

\FloatBarrier

\section{Computation Details}\label{computation-details}

\subsection{Hardware}\label{sec-app-hardware}

We performed our experiments on a high-performance cluster
(\citeproc{ref-DHPC2022}{(DHPC) 2022}). Since our experiments involve
highly parallel tasks and rather small models by today's standard, we
have relied on distributed computing across multiple central processing
units (CPU). Graphical processing units (GPU) were \emph{not} used.

\subsubsection{Grid Searches}\label{grid-searches}

Model training for the largest grid searches with 270 unique parameter
combinations was parallelized across 34 CPUs with 2GB memory each. The
time to completion varied by dataset: 0h49m (\emph{Moons}), 1h4m
(\emph{Linearly Separable}), 1h49m (\emph{Circles}), 3h52m
(\emph{Overlapping}). Model evaluations for large grid searches were
parallelized across 20 CPUs with 3GB memory each. Evaluations for all
data sets took less than one hour (\textless1h) to complete but were
generally more memory-intensive (see Section~\ref{sec-app-software} for
additional details)

\subsubsection{Tuning}\label{tuning}

For tuning of selected hyperparameters, we distributed the task of
generating counterfactuals during training across 40 CPUs with 2GB
memory each for all tabular datasets. Except for the \emph{Adult}
dataset, all training runs were completed in less that half an hour
(\textless0h30m). The \emph{Adult} dataset took around 0h35m to
complete. Evaluations across 20 CPUs with 3GB memory each generally took
less than 0h30m to complete. For \emph{MNIST}, we relied on 100 CPUs
with 2GB memory each. For the \emph{MLP}, training of all models could
be completed in 1h30m, while the evaluation across 20 CPUs (6GB memory)
took 4h12m. For the \emph{CNN}, training of all models took
\textasciitilde8h, with conventionally trained models taking
\textasciitilde0h15m each and model with CT taking
\textasciitilde0h30m-0h45m each.

\subsection{Software}\label{sec-app-software}

Our code has been open-sourced on GitHub as Julia package:
\href{https://github.com/JuliaTrustworthyAI/CounterfactualTraining.jl}{CounterfactualTraining.jl}.
All computations were performed in the Julia Programming Language
(\citeproc{ref-bezanson2017julia}{Bezanson et al. 2017}). We have
developed a package for counterfactual training that leverages and
extends the functionality provided by several existing packages, most
notably
\href{https://github.com/JuliaTrustworthyAI/CounterfactualExplanations.jl}{CounterfactualExplanations.jl}
(\citeproc{ref-altmeyer2023explaining}{Altmeyer, Deursen, and Liem
2023}) and the \href{https://fluxml.ai/Flux.jl/v0.16/}{Flux.jl} library
for deep learning (\citeproc{ref-innes2018fashionable}{Michael Innes et
al. 2018}; \citeproc{ref-innes2018flux}{Mike Innes 2018}). We chose to
work with
\href{https://github.com/JuliaTrustworthyAI/CounterfactualExplanations.jl}{CounterfactualExplanations.jl}
because it currently appears to be the most comprehensive and extensible
package for counterfactual explanations. Despite its good interplay with
Flux.jl, the package is not, however, optimized to be used in training.
This has caused some issues with memory management and bottlenecked
performance. The code is commented with clearly marked references to the
paper (look for \texttt{\#\ -\/-\/-\/-\/-\ PAPER\ REF\ -\/-\/-\/-\/-}).

For data-wrangling and presentation-ready tables we relied on
\href{https://dataframes.juliadata.org/v1.7/}{DataFrames.jl}
(\citeproc{ref-milan2023dataframes}{Bouchet-Valat and Kamiński 2023})
and
\href{https://ronisbr.github.io/PrettyTables.jl/v2.4/}{PrettyTables.jl}
(\citeproc{ref-chagas2024pretty}{Chagas et al. 2024}), respectively. For
plots and visualizations we used both
\href{https://docs.juliaplots.org/v1.40/}{Plots.jl}
(\citeproc{ref-PlotsJL}{Christ et al. 2023}) and
\href{https://docs.makie.org/v0.22/}{Makie.jl}
(\citeproc{ref-danisch2021makie}{Danisch and Krumbiegel 2021}), in
particular \href{https://aog.makie.org/v0.9.3/}{AlgebraOfGraphics.jl}.
To distribute computational tasks across multiple processors, we have
relied on \href{https://juliaparallel.org/MPI.jl/v0.20/}{MPI.jl}
(\citeproc{ref-byrne2021mpi}{Byrne, Wilcox, and Churavy 2021}).

\subsection{Reproducibility}\label{reproducibility}

We have taken care to set random seeds for reproducibility using Julia's
Random.jl package from the standard library. A global seed and (if
applicable or wanted) dataset-specific seeds can be specified in TOML
configuration files, environment variables or in interactive Julia
sessions. Additional details can be found in the code base.

\end{appendices}

\end{document}